\documentclass[11pt,table]{article}
\pdfoutput=1

\usepackage[utf8]{inputenc} 
\usepackage[T1]{fontenc}    
\usepackage[in]{fullpage}
\usepackage{microtype}
\usepackage{graphicx}
\usepackage{amsmath}
\usepackage{amsthm}
\usepackage{enumitem}
\usepackage{xfrac}
\usepackage{parskip}
\usepackage{numprint}
\usepackage{booktabs} 
\usepackage[svgnames,dvipsnames,color,table]{xcolor}
\usepackage[textsize=scriptsize,textwidth=1.9cm]{todonotes}
\usepackage{xspace}
\usepackage{subcaption}
\usepackage{etoolbox}
\usepackage{comment}
\usepackage{etoc}
\usepackage{wrapfig}
\usepackage{placeins}
\usepackage{makecell}

\newtheorem{question}{Question}

\newtoggle{showtodos}
\toggletrue{showtodos}

\newtoggle{isarxiv}
\toggletrue{isarxiv}

\definecolor{DarkRed}{rgb}{0.5,0.1,0.1}
\definecolor{DarkBlue}{rgb}{0.1,0.1,0.5}

\usepackage{hyperref}
\hypersetup{
		colorlinks=true,
		linkcolor=blue!70!black,
		citecolor=blue!70!black,
		urlcolor=blue!70!black}

\usepackage[numbers,sort]{natbib}

\usepackage{enumitem}
\usepackage{amsthm}
\usepackage{amsmath}
\usepackage{amsfonts}
\usepackage{mathtools}
\usepackage{xparse}
\usepackage{thmtools}
\usepackage{thm-restate}
\usepackage{natbib}
\usepackage{mathabx}

\iftoggle{showtodos}{
\newcommand{\john}[1]{{\todo[color=Orange!10]{John: #1}}}
\newcommand{\rohan}[1]{{\todo[color=Red!10]{Rohan: #1}}}
\newcommand{\yairside}[1]{{\todo[color=Blue!10]{Yair: #1}}}
\newcommand{\yair}[1]{{\todo[color=Blue!10]{Yair: #1}}}
\newcommand{\ludwig}[1]{\todo[color=Green!10]{Ludwig: #1}}
\newcommand{\vaishaal}[1]{{\todo[color=White!10]{Vaishaal: #1}}}
\newcommand{\pwside}[1]{{\todo[color=Purple!10]{PW: #1}}}
\newcommand{\pw}[1]{{\todo[color=Purple!10]{PW: #1}}}
\newcommand{\ar}[1]{{\todo[color=Sienna!10]{AR: #1}}}
\newcommand{\pl}[1]{{\todo[color=Yellow!10]{PL: #1}}}
\newcommand{\shiori}[1]{{\todo[color=Teal!10]{Shiori: #1}}}
\newcommand{\todogeneric}[1]{{\todo[color=Pink!10]{#1}}}
}{
\newcommand{\john}[1]{\ignorespaces}
\newcommand{\rohan}[1]{\ignorespaces}
\newcommand{\yair}[1]{\ignorespaces}
\newcommand{\yairside}[1]{\ignorespaces}
\newcommand{\ludwig}[1]{\ignorespaces}
\newcommand{\vaishaal}[1]{\ignorespaces}
\newcommand{\pw}[1]{\ignorespaces}
\newcommand{\pwside}[1]{\ignorespaces}
\newcommand{\ar}[1]{\ignorespaces}
\newcommand{\pl}[1]{\ignorespaces}
\newcommand{\shiori}[1]{\ignorespaces}
\newcommand{\todogeneric}[1]{\ignorespaces}
}

\theoremstyle{plain}
\newtheorem{conjecture}{Conjecture}

\newtheorem{definition}{Definition}

\theoremstyle{definition}

\newtheorem*{example*}{Example}

\DeclarePairedDelimiter{\abs}{\lvert}{\rvert} %

\DeclarePairedDelimiter{\prn}{(}{)}

\DeclarePairedDelimiter{\norm}{\|}{\|}

\newcommand{\mc}[1]{\mathcal{#1}}
\newcommand{\R}{\mathbb{R}}
\DeclareMathOperator{\sign}{\mathrm{sign}}
\DeclareMathOperator{\tr}{\mathrm{tr}}

\newcommand{\cifarten}{CIFAR-10\xspace}
\newcommand{\cifartenone}{CIFAR-10.1\xspace}
\newcommand{\cifartentwo}{CIFAR-10.2\xspace}
\newcommand{\cifartenc}{CIFAR-10-C\xspace}
\newcommand{\cifartencorrupted}{CIFAR-10-Corrupted\xspace}
\newcommand{\fmow}{FMoW-WILDS\xspace}
\newcommand{\iwildcam}{iWildCam-WILDS\xspace}
\newcommand{\camelyon}{Camelyon17-WILDS\xspace}
\newcommand{\imagenet}{ImageNet\xspace}
\newcommand{\imagenettwo}{ImageNet-V2\xspace}
\newcommand{\imagenetc}{ImageNet-C\xspace}
\newcommand{\stlten}{STL-10\xspace}
\newcommand{\cinicten}{CINIC-10\xspace}
\newcommand{\ycb}{YCB-Objects\xspace}

\begin{document}

\title{Accuracy on the Line: On the Strong Correlation \\ Between Out-of-Distribution and In-Distribution Generalization}
\date{}
\author{
    \hspace{1.7cm}
    John Miller\thanks{University of California, Berkeley, \texttt{\{miller\_john, vaishaal\}@berkeley.edu}}
    \and
    \hspace{0.4cm}
    Rohan Taori\thanks{Stanford University, \texttt{\{rtaori, aditir, ssagawa, pangwei, pliang\}@cs.stanford.edu}}
    \and
    \hspace{0.4cm}
    Aditi Raghunathan\footnotemark[2]
    \hspace{0.4cm}
    \and
    Shiori Sagawa\footnotemark[2]
    \and
    Pang Wei Koh\footnotemark[2]
    \and
    Vaishaal Shankar\footnotemark[1]
    \and
    Percy Liang\footnotemark[2]
    \and
    Yair Carmon\thanks{Tel Aviv University, \texttt{ycarmon@gmail.com}}
    \and
    Ludwig Schmidt\thanks{Toyota Research Institute, \texttt{ludwigschmidt2@gmail.com}}
}
\maketitle
\vspace{-.2cm}

\begin{abstract}
For machine learning systems to be reliable, we must understand their performance in unseen, out-of-distribution environments.
In this paper, we empirically show that out-of-distribution performance is strongly correlated with in-distribution performance for a wide range of models and distribution shifts. 
Specifically, we demonstrate strong correlations between in-distribution and out-of-distribution performance on variants of CIFAR-10 \& ImageNet, a synthetic pose estimation task derived from YCB objects, satellite imagery classification in FMoW-WILDS, and wildlife classification in iWildCam-WILDS.
The strong correlations hold across model architectures, hyperparameters, training set size, and training duration, and are more precise than what is expected from existing domain adaptation theory.
To complete the picture, we also investigate cases where the correlation is weaker, for instance some synthetic distribution shifts from CIFAR-10-C and the tissue classification dataset Camelyon17-WILDS. 
Finally, we provide a candidate theory based on a Gaussian data model that shows how changes in the data covariance arising from distribution shift can affect the observed correlations. 
\end{abstract}


\etocdepthtag.toc{mtsection}

\iftoggle{isarxiv}{
    \begin{figure*}[htp!]
}{
    \begin{figure*}[ht!]
}
    \centering
    \includegraphics[width=\linewidth]{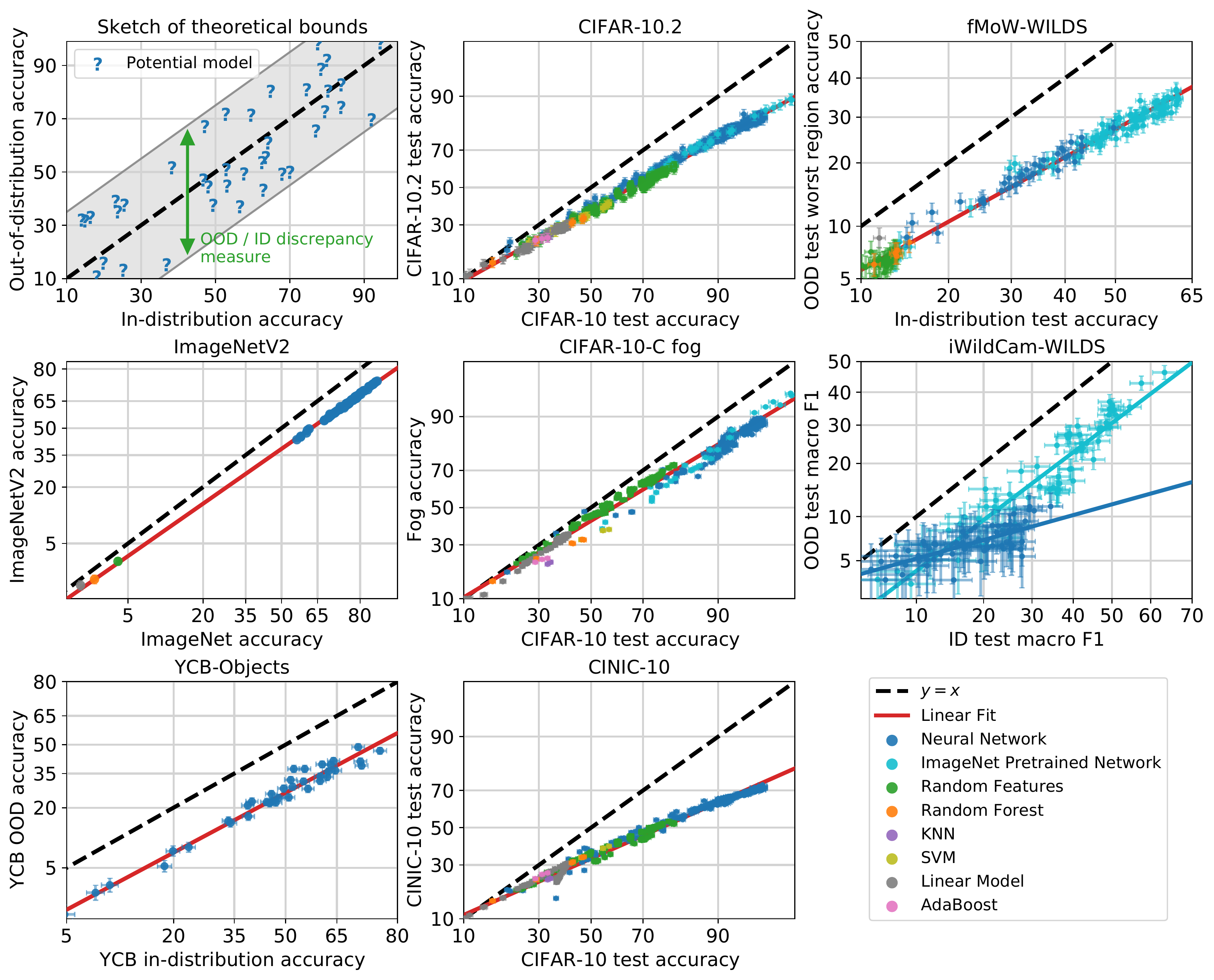}
    \vspace{-.3cm}
    \caption{
      Out-of-distribution accuracies vs. in-distribution accuracies for a wide range of models, datasets, and distribution shifts.
      \textbf{Top left:} A sketch of the current bounds from domain adaptation theory. These bounds depend on distributional distances between in-distribution and out-of-distribution data, and they are loose in that they limit the deviation away from the y = x diagonal but do not prescribe a specific trend within these wide bounds (see Section \ref{sec:related_work}).
      \textbf{Remaining panels:} In contrast, we show that for a wide range of
      models and datasets, there is a precise linear trend between
      out-of-distribution accuracy and in-distribution accuracy. Unlike what we
      might expect from theory, the linear trend does not follow the $y = x$
      diagonal. The different panels represent different pairs of
      in-distribution and out-of-distribution datasets. Within each panel, we
      plot the performances of many different models, with different model
      architectures and hyperparameters.
      These datasets capture a variety of distribution shifts from dataset
      reproduction (\cifartentwo, \imagenettwo); a real-world spatiotemporal distribution shift
      on satellite imagery (\fmow); using a different benchmark test dataset
      (\cinicten); synthetic perturbations (\cifartenc and \ycb); and a real-world
      geographic shift in wildlife monitoring (\iwildcam). Interestingly, for
      \iwildcam, models pretrained on \imagenet follow a different linear trend
      than models trained from scratch in-distribution, and we plot a separate
      trend line for \imagenet pretrained models in the \iwildcam panel.  We
      explore this phenomenon more in Section~\ref{sec:training_data}.
    }
    \label{fig:main_figure}
    \vspace{-.3cm}
\end{figure*}

\section{Introduction}
\label{sec:intro}
Machine learning models often need to generalize from training data to new environments.
A kitchen robot should work reliably in different homes, autonomous vehicles should drive reliably in different cities, and analysis software for satellite imagery should still perform well next year.
The standard paradigm to measure generalization is to evaluate a model on a single test set drawn from the same distribution as the training set.
But this paradigm provides only a narrow \emph{in-distribution} performance guarantee: a small test error certifies future performance on new samples from exactly the same distribution as the training set. 
In many scenarios, it is hard or impossible to train a model on precisely the distribution it will be applied to.
Hence a model will inevitably encounter \emph{out-of-distribution} data on which its performance could vary widely compared to in-distribution performance.
Understanding the performance of models beyond the training distribution therefore raises the following fundamental question: how does out-of-distribution performance relate to in-distribution performance? 


Classical theory for generalization across different distributions provides a partial answer~\cite{mansour2009domain,bendavid2010theory}.
For a model $f$ trained on a distribution $D$, known guarantees typically relate the in-distribution test accuracy on $D$ to the out-of-distribution test accuracy on a new distribution $D'$ via inequalities of the form
\[
    | \text{acc}_D(f) - \text{acc}_{D'}(f) | \; \leq \;  d(D, D')
\]
where $d$ is a distance between the distributions $D$ and $D'$ such as the total variation distance.
Qualitatively, these bounds suggest that out-of-distribution accuracy may vary widely as a function of in-distribution accuracy unless the distribution distance $d$ is small and the accuracies are therefore close (see Figure 1 (top-left) for an illustration).
More recently, empirical studies have shown that in some settings, models with similar in-distribution performance can indeed have different out-of-distribution performance~\citep{mccoy2019berts,zhou2020curse,damor2020underspecification}.

In contrast to the aforementioned results, recent dataset reconstructions of the popular CIFAR-10, ImageNet, MNIST, and SQuAD benchmarks showed a much more regular pattern \cite{recht2019imagenet,miller2020effect,qmnist,lu2020harder}.
The reconstructions closely followed the original dataset creation processes to assemble new test sets, but small differences were still enough to cause substantial changes in the resulting model accuracies.
Nevertheless, the new out-of-distribution accuracies are almost perfectly linearly correlated with the original in-distribution accuracies for a range of deep neural networks.
Importantly, this correlation holds \emph{despite the substantial gap between in-distribution and out-of-distribution accuracies} (see Figure 1 (top-middle) for an example).
However, it is currently unclear how widely these linear trends apply since they have been mainly observed for dataset reproductions and common variations of convolutional neural networks.

In this paper, we conduct a broad empirical investigation to characterize when precise linear trends such as in Figure 1 (top-middle) may be expected, and when out-of-distribution performance is less predictable as in Figure 1 (top-left).
Concretely, we make the following contributions:
\begin{itemize}
  \item We show that precise linear trends occur on several datasets and associated distribution shifts (see Figure 1).
  Going beyond the dataset reproductions in earlier work, we find linear trends on 
  \begin{itemize}
    \item popular image classification benchmarks (\cifarten \cite{krizhevsky2009learning}, \cifartenone \cite{recht2019imagenet}, \cifartentwo \cite{lu2020harder}, \cifartenc \cite{hendrycks2018benchmarking}, \cinicten \cite{darlow2018cinic}, \stlten \cite{coates2011analysis}, \imagenet \cite{deng2009imagenet}, \imagenettwo \cite{recht2019imagenet}),
    \item a pose estimation testbed based on \ycb \cite{calli2015benchmarking},
    \item and two distribution shifts derived from concrete applications of image classification: satellite imagery and wildlife photos via the \fmow and \iwildcam variants from WILDS~\citep{christie2018functional,beery2020iwildcam,koh2020wilds}.
  \end{itemize}

\item We show that the linear trends hold for many models ranging from state-of-the-art methods such as convolutional neural networks, visual transformers, and self-supervised models, to classical methods like logistic regression, nearest neighbors, and kernel machines.
  Importantly, we find that classical methods follow the same linear trend as more recent deep learning architectures.
  Moreover, we demonstrate that varying model or training hyperparameters, training set size, and training duration all result in models that follow the same linear trend.

\item We also identify three settings in which the linear trends do \emph{not} occur or are less regular: some of the synthetic distribution shifts in \cifartenc (e.g., Gaussian noise), the \camelyon shift of tissue slides from different hospitals, and a version of the aforementioned \iwildcam wildlife classification problem with a different in-distribution train-test split \cite{beery2020iwildcam}. 
We analyze these cases in detail via additional experiments to pinpoint possible causes of the linear trends.

\item Pre-training a model on a larger and more diverse dataset offers a possibility to increase robustness.
Hence we evaluate a range of models pre-trained on other datasets to study the impact of pre-training on the linear trends. 
Interestingly, even pre-trained models sometimes follow the same linear trends as models trained only on the in-distribution training set.
Two examples are ImageNet pre-trained models evaluated on \cifarten and \fmow.
In other cases (e.g., \iwildcam), pre-training yields clearly different relationships between in-distribution and out-of-distribution accuracies.

\item As a starting point for theory development, we provide a candidate theory based on a simple Gaussian data model.
Despite its simplicity, this data model correctly identifies the covariance structure of the distribution shift as one property affecting the performance correlation on the Gaussian noise corruption from \cifartenc. 
\end{itemize}

Overall, our results show a striking linear correlation between the in-distribution and out-of-distribution performance of many models on multiple distribution shifts.
This raises the intriguing possibility that, despite their different creation mechanisms, a diverse range of distribution shifts may share common phenomena.
In particular, improving in-distribution performance reliably improves out-of-distribution performance as well. 
However, it is currently unclear whether improving in-distribution performance is the only way, or even the best way, to improve out-of-distribution performance.
More research is needed to understand the extent of the linear trends observed in this work and whether robustness interventions can improve over the baseline given by empirical risk minimization.
We hope that our work serves as a step towards a better understanding of distribution shift and how we can train models that perform robustly out-of-distribution.

\iftoggle{isarxiv}{\paragraph{Paper outline.} Next, we introduce our main experimental framework that forms the backbone of our investigation.
The following sections instantiate this framework for multiple distribution shifts.

Section \ref{sec:linear_trends} shows results on a wide range of distribution shifts where precise linear trends do occur.
Section \ref{sec:fit_failure} then turns to distribution shifts where the linear trends are less regular or do not exist.
Section \ref{sec:training_data} investigates the role of pretraining in more detail since models pre-trained on a different dataset sometimes -- but not always -- deviate from the linear trends.

After our experiments, we briefly summarize the empirical phenomena in Section \ref{sec:empirical-summary} and then present our theoretical model in Section \ref{sec:theory}.
Section \ref{sec:related_work} describes related work and Section \ref{sec:discussion} concludes with a discussion of our results, possible implications for research on reliable machine learning, and directions for future work.
}{}

\section{Experimental setup}
In each of our main experiments, we compare performance on two data distributions.
The first is the training distribution $D$, which  we refer to as ``in-distribution'' (ID).
Unless noted otherwise, all models are trained only on samples from $D$ (the main exception is pre-training on a different distribution).
We compute ID performance via a held-out test set sampled from $D$.
The second distribution is the ``out-of-distribution'' (OOD) distribution $D'$ that we also evaluate the models on.
    For a loss function $\ell$ (e.g., error or accuracy), we denote the loss of model $f$ on distribution $D$ with $\ell_{D}(f) = \mathbb{E}_{x,y\sim D} \left[ \ell(f(x), y) \right]$.

\paragraph{Experimental procedure.} The goal of our paper is to understand the relationship between $\ell_{D}(f)$ and $\ell_{D'}(f)$ for a wide range of models $f$ (convolutional neural networks, kernel machines, etc.) and pairs of distributions $D, D'$ (e.g., CIFAR-10 and the CIFAR-10.2 reproduction).
Hence for each pair $D, D'$, our core experiment follows three steps:
\begin{enumerate}
  \item Train a set of models $\{ f_1, f_2, \ldots \}$ on samples drawn from $D$.
  Apart from the shared training distribution, the models are trained independently with different training set sizes, model architectures, random seeds, optimization algorithms, etc.
\item Evaluate the trained models $f_i$ on two test sets drawn from $D$ and $D'$, respectively.
\item Display the models $f_i$ in a scatter plot with each model's two test accuracies on the two axes to inspect the resulting correlation.
\end{enumerate}
An important aspect of our scatter plots is that we apply a non-linear transformation to each axis.
Since we work with loss functions bounded in $[0, 1]$, we apply an axis scaling that maps $[0, 1]$ to $[-\infty, +\infty]$ via the probit transform.
The probit transform is the inverse of the cumulative density function (CDF) of the standard Gaussian distribution, i.e., $l_\text{transformed} = \Phi^{-1}(l)$.
Transformations like the probit or closely related logit transform are often used in statistics since a quantity bounded in $[0, 1]$ can only show linear trends for a bounded range.
The linear trends we observe in our correlation plots are substantially more precise with the probit (or logit) axis scaling.
Unless noted otherwise, each point in a scatter plot is a single model (not averaged over random seeds) and we show each point with 95\% Clopper-Pearson confidence intervals for the accuracies.

We assembled a unified testbed that is shared across experiments and includes a multitude of models ranging from classical methods like nearest neighbors, kernel machines, and random forests to a variety of high-performance convolutional neural networks.
Our experiments involved more than 3,000 trained models and 100,000 test set evaluations of these models and their training checkpoints.
Due to the size of these experiments, we defer a detailed description of the
testbed used to Appendix~\ref{app:testbed}. 

\section{The linear trend phenomenon}
\label{sec:linear_trends}
In this section, we show precise linear trends between in-distribution and
out-of-distribution performance occur across a diverse set of models, data
domains, and distribution shifts. Moreover, the linear trends holds not just
across variations in models and model architectures, but also across variation in
model or training hyperparameters, training dataset size, and training duration.

\renewcommand{\floatpagefraction}{.75}%
\iftoggle{isarxiv}{
    \begin{table}
        \centering
        \rowcolors{2}{gray!15}{white}
\renewcommand{\arraystretch}{1.3}
\begin{tabular}{llcc}
\toprule
    ID Dataset & OOD Dataset & \makecell{$R^2$ of linear fit \\ (probit domain)} & \makecell{Number of models \\ evaluated}\\[.1cm]
\midrule
    \cifarten & \cifartenone & 0.995 & \numprint{1,060} \\
              & \cifartentwo & 0.997 & \numprint{1,060} \\
              & \cinicten & 0.991 & \numprint{949} \\
              & \stlten & 0.995 & \numprint{456} \\
              & \cifartenc{} {\tt Fog} & 0.990 & \numprint{790} \\
              & \cifartenc{} {\tt Brightness} & 0.940 & \numprint{519} \\
    \imagenet & \imagenettwo & 0.996 & \numprint{219} \\
    \ycb & \ycb OOD & 0.975 & \numprint{39}\\
    \iwildcam ID & \iwildcam OOD & 0.881 (0.536) & \numprint{66} (\numprint{63})\\
    \fmow ID  & \fmow OOD & 0.984 & \numprint{162} \\
\bottomrule
\end{tabular}
\caption{Summary of ID and OOD pairs where we observe precise linear trends in
our experiments. Number of models evaluated is the number of models trained on
the ID training set, evaluated on the ID and OOD test sets, and used to compute
the linear fits. For brevity, we list only two \cifartenc shifts;
see Appendix~\ref{app:corruptions} for a complete list.  As
discussed in Section~\ref{sec:training_data}, on \iwildcam, \imagenet pretrained
models exhibit a different linear trend than models trained from scratch.
Therefore, for \iwildcam, we report the number of models and $R^2$
for pretrained models and models trained from scratch (given in parentheses)
separately.}

        \label{table:linear_trends}
    \end{table}
}{}
\renewcommand{\floatpagefraction}{.6}%

\subsection{Distribution shifts with linear trends}
We find linear trends for models in our testbed trained on five different
datasets---\cifarten, \imagenet, \fmow, \iwildcam, and \ycb---and evaluated on
distribution shifts that fall into four broad categories. 
\iftoggle{isarxiv}{
These trends are summarized in Table~\ref{table:linear_trends}.
}{}

\textbf{Dataset reproduction shifts.}
Dataset reproductions involve collecting a new test set by closely matching the
creation process of the original. Distribution shift arises as a result of
subtle differences in the dataset construction pipelines.  Recent examples of
dataset reproductions are the \cifartenone and \imagenettwo test sets
from~\citet{recht2019imagenet}, who observed linear trends for deep models on
these shifts. In Figure~\ref{fig:main_figure}, we extend this result and show
both deep \emph{and classical} models trained on \cifarten and evaluated on
\cifartentwo~\citep{lu2020harder} follow a linear trend. In
Appendix~\ref{app:more_linear_trends}, we further show linear trends occur for
deep and classical \cifarten models evaluated on \cifartenone and for~\imagenet
models evaluated on~\imagenettwo.

\textbf{Distribution shifts between machine learning benchmarks.}
We also consider distribution shifts between distinct benchmarks which are drawn
from different data sources, but which use a compatible set of labels.  For
instance, both~\cifarten and~\cinicten~\citep{darlow2018cinic} use the same set
of labels, but \cifarten is drawn from TinyImages~\citep{tinyimages}
and~\cinicten is drawn from~\imagenet~\cite{deng2009imagenet} images.
We show \cifarten models exhibit linear trends
when evaluated  on~\cinicten (Figure~\ref{fig:main_figure}) or
on~\stlten~\citep{coates2011analysis} (Appendix~\ref{app:more_linear_trends}).

\textbf{Synthetic perturbations.}
Synthetic distribution shifts arise from applying a perturbation, such as adding
Gaussian noise, to existing test examples.
\cifartenc~\cite{hendrycks2018benchmarking} applies 19 different synthetic
perturbations to the \cifarten test set.  For many of these perturbations, we
observe linear trends for~\cifarten trained models, e.g. the {\tt Fog} shift in
Figure~\ref{fig:main_figure}.  However, there are several exceptions, most
notably adding isotropic Gaussian noise. We give further examples of linear
trends on synthetic \cifartenc shifts in Appendix~\ref{app:more_linear_trends},
and we more thoroughly discuss non-examples of linear trends in
Section~\ref{sec:fit_failure}. In Figure~\ref{fig:main_figure}, we also show
that pose-estimation models trained on rendered images of
\ycb~\citep{calli2015benchmarking} follow a linear trend when evaluated on a
images rendered with perturbed lighting and texture conditions.

\textbf{Distribution shifts in the wild.}
We also find linear trends on two of the real-world distribution shifts from the
WILDS benchmark~\citep{koh2020wilds}:~\fmow~and~\iwildcam. \fmow is a satellite
image classification task derived from~\citet{christie2018functional} where
in-distribution data is taken from regions (e.g., the Americas, Africa, Europe)
across the Earth between 2002 and 2013, the out-of-distribution test-set is
sampled from each region during 2016 to 2018, and models are evaluated by their
accuracy on the worst-performing region.  In Figure~\ref{fig:main_figure}, we
show models trained on \fmow exhibit linear trends when evaluated
out-of-distribution under both of these temporal and subpopulation distribution
shifts.

\iwildcam is an image dataset of animal photos taken by camera traps
deployed in multiple locations around the world~\citep{koh2020wilds, beery2020iwildcam}.
It is a multi-class classification task, where the goal is to identify the animal species (if any)
within each photo.  The held-out test set comprises photos taken by camera
traps that are not seen in the training set, and the distribution shift arises
because different camera traps vary markedly in terms of angle, lighting,
and background. In Figure~\ref{fig:main_figure}, we
show models trained on \iwildcam also exhibit linear trends when evaluated
OOD across different camera traps.

\subsection{Variations in model hyperparameters, training duration, and training
set size}
\label{sec:linear_trends-results}
\begin{figure*}
    \centering
    \iftoggle{isarxiv}{
        \includegraphics[width=\linewidth]{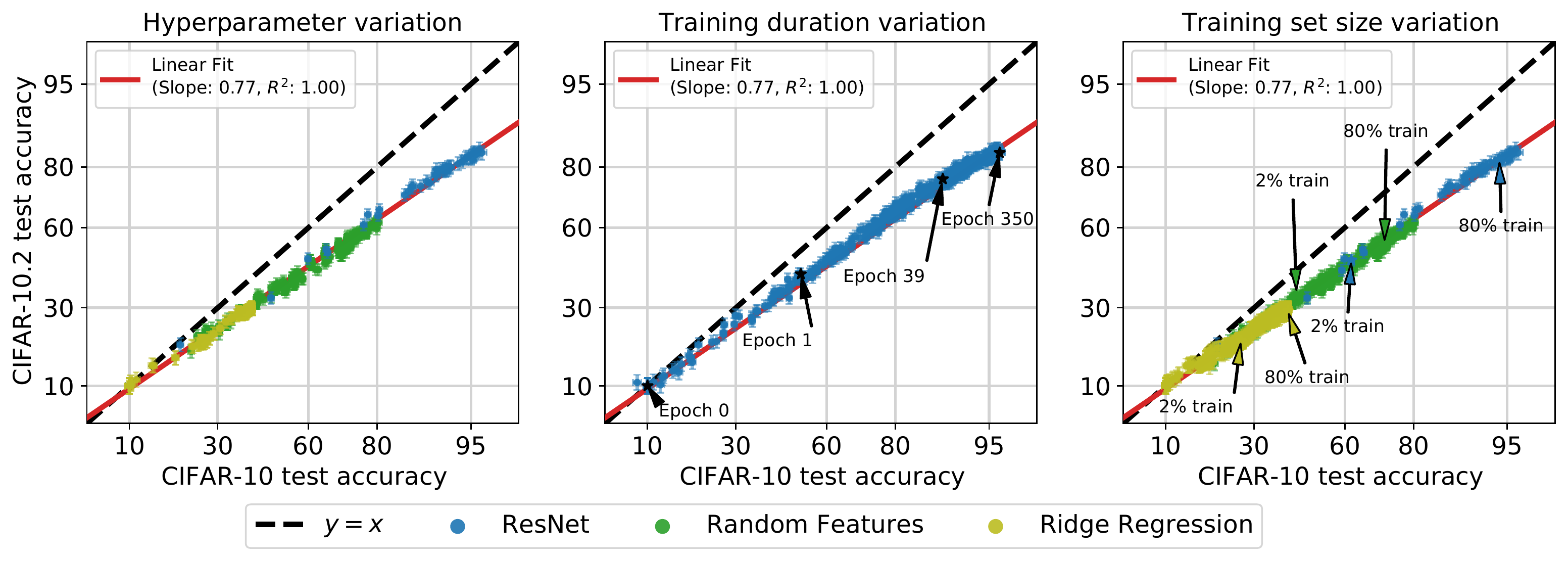}
    }{
        \includegraphics[height=5.3cm]{figures/linear_fits_cifar}
    }
    \vspace{-0.3cm}
    \caption{
        The linear trend between ID and OOD accuracy is invariant to changes in
        model hyperparameters, the number of training steps, and training set
        size.  In each panel, we compare models with the linear fit
        from Figure~\ref{fig:main_figure}. \textbf{Left:} For each model family,
        we vary model-size, regularization, and optimization hyperparameters.
        \textbf{Middle:} We evaluate each network after every epoch of training.
        \textbf{Right:} We train models on randomly sampled subsets of the
        training data, ranging from 1\% to 80\% of the \cifarten training set
        size. In each setting, variation in hyperparameters, training duration,
        or training set size moves models along the trend line, but does not
        affect the linear fit.
    }
    \label{fig:exploring_linear_fits}
\end{figure*}

The linear trends we observe hold not just across different models, but also
across variation in model and optimization hyperparameters, training dataset
size, and training duration. 

In Figure~\ref{fig:exploring_linear_fits}, we train and evaluate
both classical and neural models on \cifarten and
\cifartentwo while systematically varying (1) model hyperparameters, (2)
training duration, and (3) training dataset size. When varying hyperparameters
controlling the model size, regularization, and the optimization algorithm, the
model families continue to follow the same trend line ($R^2=0.99$).
We also find models lie on the same linear trend line
\emph{throughout training} ($R^2=0.99$).  Finally, we observe models on trained
on random subsets of \cifarten lie on the same linear trend line as
models trained on the full \cifarten training set, despite their corresponding
drop in in-distribution accuracy ($R^2=0.99$).  In each case, hyperparameter
tuning, early stopping, or changing the amount of i.i.d. training data moves
models along the trend line, but does not alter the linear fit.

While we focus here on \cifarten models evaluated on \cifartentwo, 
in Appendix~\ref{app:more_linear_trends}, we conduct an identical set of
experiments for \cinicten, \cifartenc{} {\tt Fog}, \ycb, and \fmow.
We find the same invariance to hyperparameter, dataset size, and training
duration shown in Figure~\ref{fig:exploring_linear_fits} also holds for these
diverse collection of datasets.

\section{Distribution shifts with weaker correlations}
\label{sec:fit_failure}

We now investigate distribution shifts with a weaker correlation between ID and OOD performance than the examples presented in the previous section.
We will discuss the \camelyon tissue classification dataset and specific image corruptions from \cifartenc.
Further discussion of a version of the \iwildcam wildlife classification dataset with a different in-distribution train-test split can be found in Appendix \ref{app:linear-trends-iwc}.

\subsection{\camelyon}

\iftoggle{isarxiv}{
\begin{figure*}[t!]
    \centering
    \includegraphics[width=0.38\textwidth]{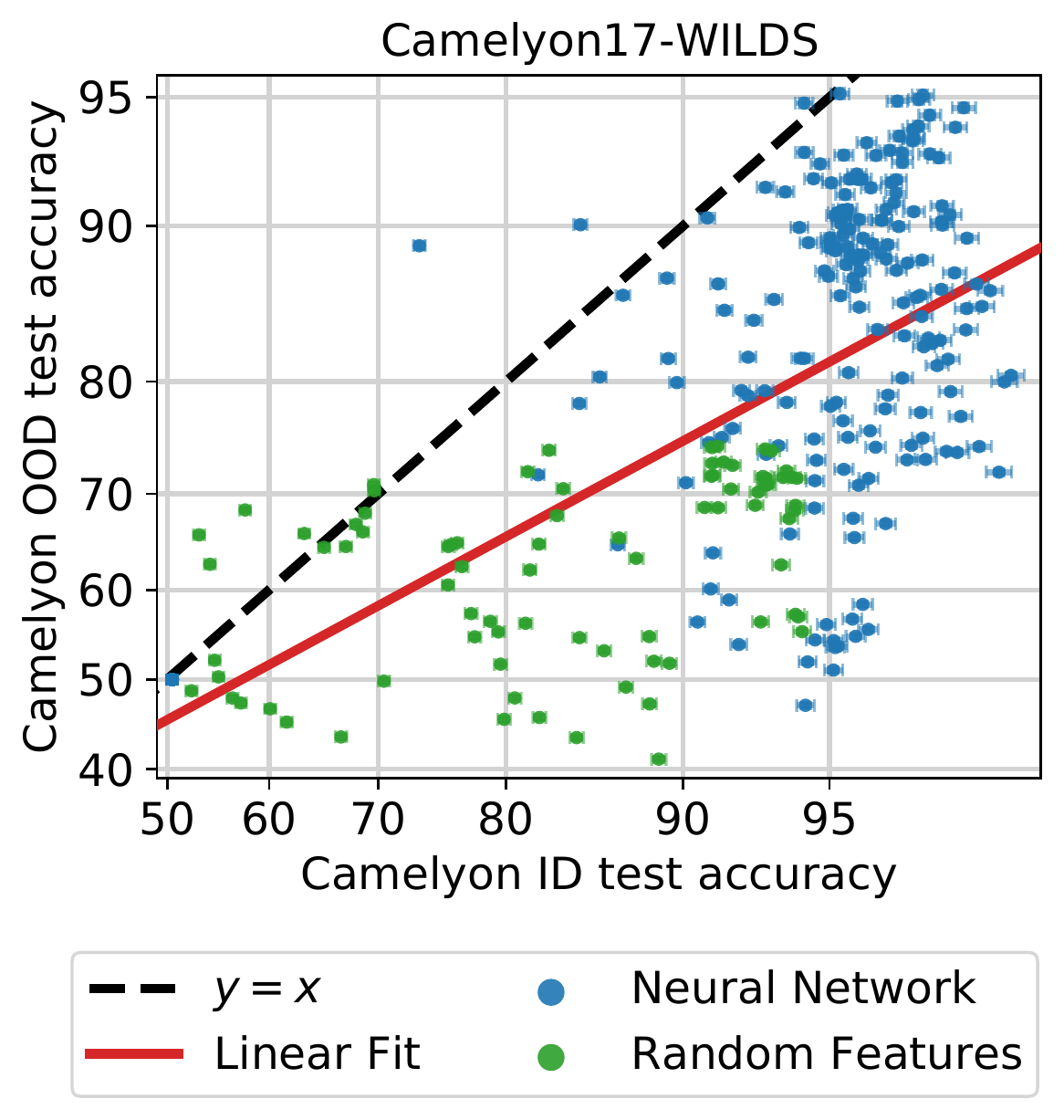}
    \caption{
        A range of neural network and random feature models trained on \camelyon{} and evaluated on the ID and OOD test sets.
        OOD accuracy is highly variable across the spectrum of ID accuracies,
        and there is no precise linear trend.
    } \label{fig:camelyon}
\end{figure*}
}{
\begin{figure}[ht!]
    \centering
        \includegraphics[height=7cm]{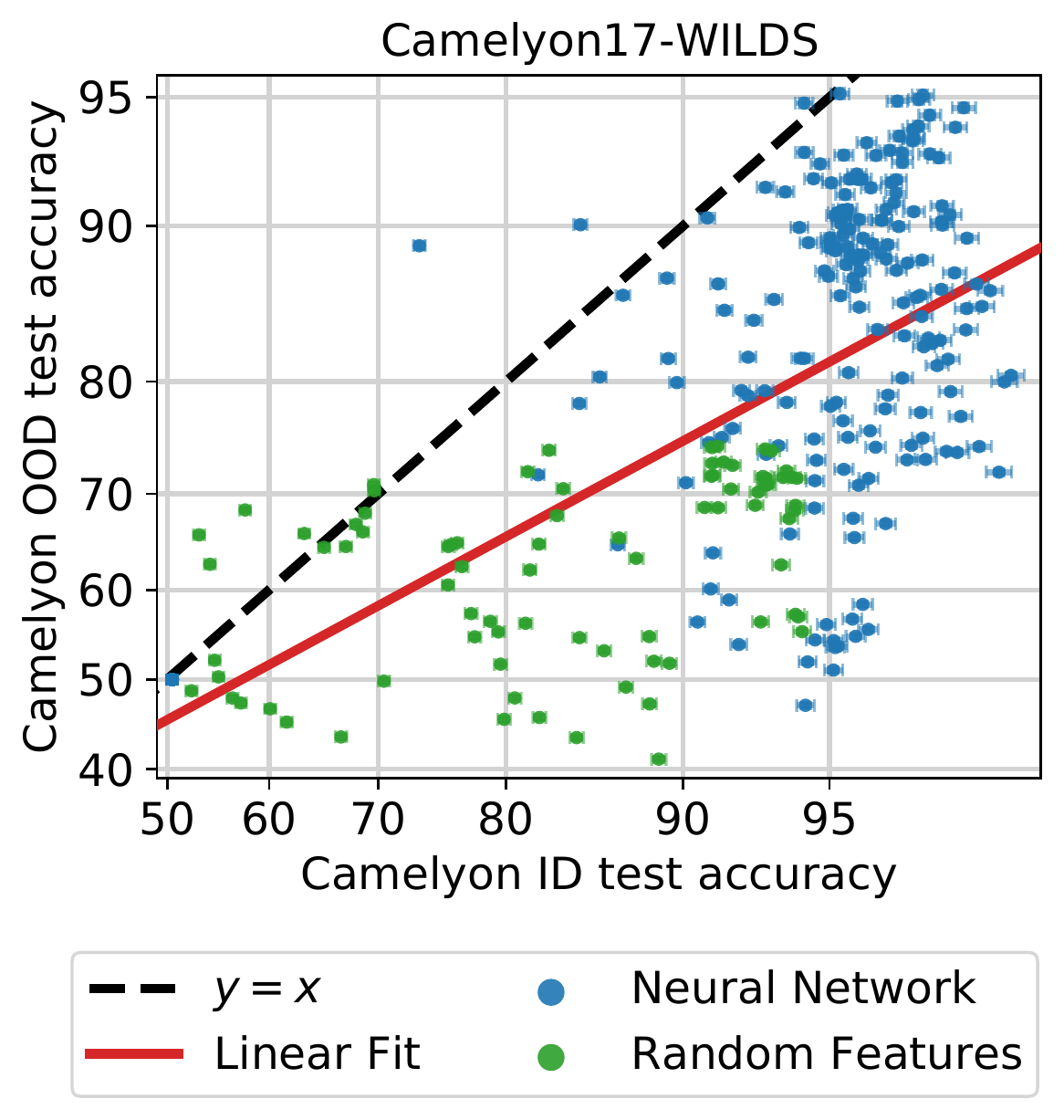}
    \vspace{-0.3cm}
    \caption{
        A range of neural network and random feature models trained on \camelyon{} and evaluated on the ID and OOD test sets.
        OOD accuracy is highly variable across the spectrum of ID accuracies,
        and there is no precise linear trend.
    } \label{fig:camelyon}
\end{figure}
}

\camelyon \cite{bandi2018camelyon,koh2020wilds} is an image dataset of metastasized breast cancer tissue samples collected from different hospitals.
It is a binary image classification task where each example is a tissue patch.
The corresponding label is whether the patch contains any tumor tissue.
The held-out OOD test set contains tissue samples from a hospital not seen in the training set.
The distribution shift largely arises from differences in staining and imaging protocols across hospitals.

In Figure \ref{fig:camelyon}, we plot the results of training different \imagenet{} models
and random features models from scratch across a variety of random seeds.
There is significant variation in OOD performance.
For example, the models with 95\% ID accuracy have OOD accuracies that range from about 50\% (random chance) to 95\%.
This high degree of variability holds even after averaging each model over ten independent training runs (see Appendix \ref{app:camelyon}).

Appendix \ref{app:camelyon} also contains additional analyses exploring the potential sources of OOD performance variation, including \imagenet{} pretraining, data augmentation, and similarity between test examples.
Specifically, we observe that \imagenet{} pretraining does not increase the ID-OOD correlation, while strong data augmentation significantly reduces, but does not eliminate, the OOD variation.
Another potential reason for the variation is the similarity between images from the same slide / hospital, as similar examples have been shown to result in analogous phenomena in natural language processing \citep{zhou2020curse}.
We explore this hypothesis in a synthetic CIFAR-10 setting, where we simulate increasing the similarity between examples by taking a small seed set of examples and then using data augmentations to create multiple similar versions. 
We find that in this CIFAR-10 setting, shrinking the effective test set size in this way increases OOD variation to a substantially greater extent than shrinking the effective training set size.

\subsection{\cifartencorrupted}\label{ssec:cifarten-corrupted}

\iftoggle{isarxiv}{
\cifartenc{} \cite{hendrycks2018benchmarking} corrupts the \cifarten{} test set with various image perturbations.
The choice of corruption can have a significant impact on the correlation
between ID and OOD accuracy.  Appendix \ref{app:corruptions} provides plots and
$R^2$ values for each corruption.  We already showed an example of one of the
more precise fits, fog corruption, in Figure \ref{fig:main_figure} (bottom
middle).  Interestingly, the mathematically easy to describe corruption with
Gaussian noise is one of the corruptions with worst ID-OOD correlation (see
Figure \ref{fig:cifar10_gaussian_noise_comparison} left).

\begin{figure*}
    \centering
    \includegraphics[height=6.5cm]{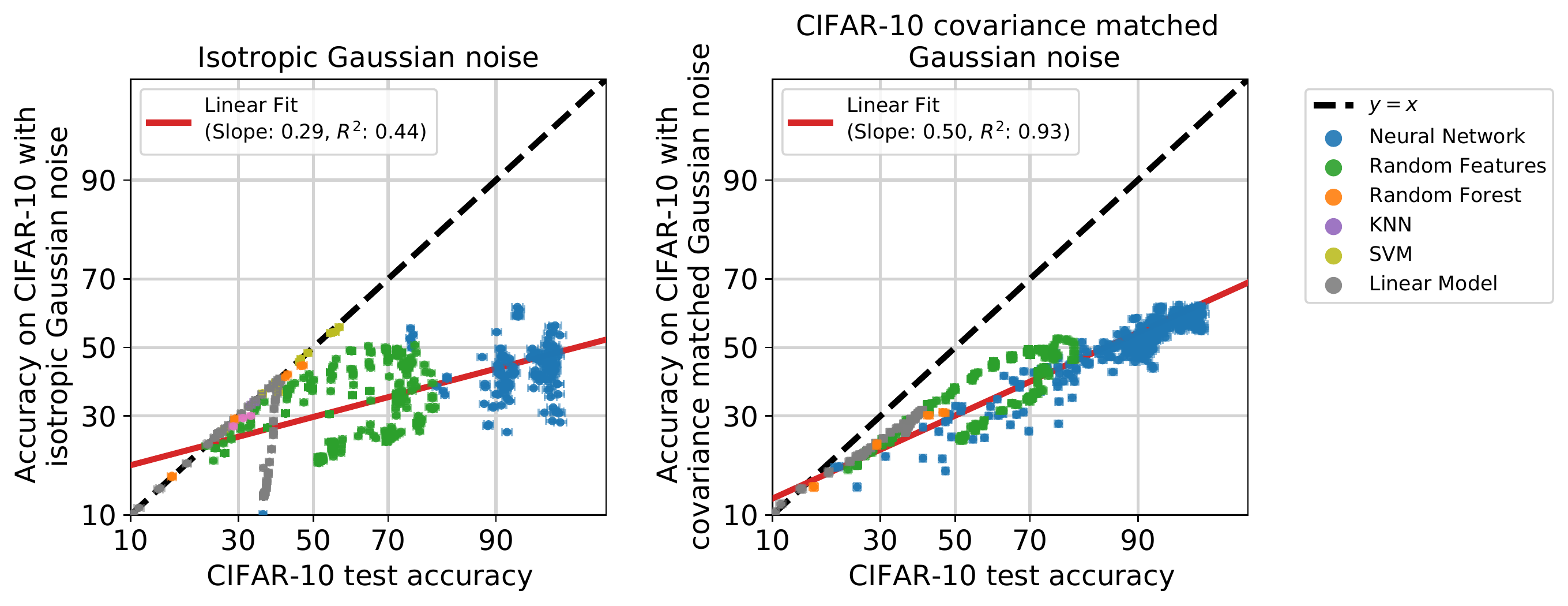}
    \caption{
        When the out-of-distribution data covariance matches the
        in-distribution data covariance, the linear fit is significantly
        better.
        \textbf{Left:} A collection of models trained on \cifarten and
        evaluated in-distribution on \cifarten and out-of-distribution on
        \cifarten images corrupted with \emph{isotropic} Gaussian noise.
        \textbf{Right:} The same collection of models evaluated
        out-of-distribution on \cifarten images corrupted with Gaussian
        noise with the \emph{same covariance as \cifarten}.
    }
    \label{fig:cifar10_gaussian_noise_comparison}
\end{figure*}

We also investigate how the relationship between the ID and OOD data covariances
impacts the linear trend. 
The theoretical model discussed in Section~\ref{sec:theory} predicts linear
fits occur if the data covariances between ID and OOD are the same up to a
constant scaling factor.  Thus, in
Figure~\ref{fig:cifar10_gaussian_noise_comparison}, we compare adding
\emph{isotropic} Gaussian noise to the \cifarten test set versus adding Gaussian
noise with the \emph{same covariance as data examples from \cifarten}.  We find
that when the OOD covariance matches the ID covariance the linear fit is
substantially better ($R^2 = 0.93$ vs.\ $R^2 = 0.44$).  This finding is
consistent with the theoretical model we propose and discuss in
Section~\ref{sec:theory}.

In Appendix \ref{app:corruptions}, we also compare \cifartenc{} to \imagenetc{}
and notice that each corruption displays a more linear trend on \imagenetc
compared to \cifartenc.  Investigating this discrepancy further is an
interesting direction for future work.
}{ 
\cifartenc{} \cite{hendrycks2018benchmarking} corrupts \cifarten{} test images
with various image perturbations.  The choice of corruption can have a
significant impact on the correlation between ID and OOD accuracy.
Interestingly, the mathematically simple corruption with Gaussian noise is one
of the corruptions with worst ID-OOD correlation. Appendix \ref{app:corruptions}
details experiments for each corruption.

In Appendix~\ref{app:cifar10cgaussian}, we also investigate how the relationship
between the ID and OOD data covariances impacts the linear trend. We find the
linear fit is substantially better when the ID and OOD covariances match up to a
scaling factor, which is consistent with the theoretical model we propose and
discuss in Section~\ref{sec:theory}.
%
}

\section{The effect of pretrained models} \label{sec:training_data}
In this section, we expand our scope to methods that leverage models pretrained on a third auxiliary distribution different from the ones we refer to in-distribution (ID) and out-of-distribution (OOD). 
Fine-tuning pretrained models on the task-specific (ID) training set is a central technique in modern machine learning \citep{decaf,cnnshelf,kornblith2019better,elmo,bert}, and zero-shot prediction (using the pretrained model directly without any task-specific training)  is showing increasing promise as well~\cite{brown2020language,radford2021learning}. Therefore, it is important to understand how the use of pretrained models affects the robustness of models to OOD data, and whether fine-tuning and zero-shot inference differ in that respect.

The dependence of the pretrained model on auxiliary data makes the ID/OOD distinction more subtle.
Previously, ``ID'' simply referred to the distribution of the training set, while OOD referred to an alternative distribution not seen in training.
In this section, the training set includes the auxiliary data as well, but we still refer to the \emph{task-specific} training set distributions as ID.
This means, for example, that when fine-tuning an \imagenet model on the \cifarten training set, we still refer to accuracy on the \cifarten test set as ID accuracy.
In other words, the ``ID'' distributions we refer to in this section are precisely the ``ID'' distributions of the previous sections (displayed on the $x$-axes in our scatter plots), but the presence of auxiliary training data alters the meaning of the term.

With the effect of auxiliary data on the meaning of ``ID'' in mind, it is reasonable to expect that ID/OOD linear trends observed when training purely on ID data will change or break down when pretrained models are used. In this section we test this hypothesis empirically and reveal a more nuanced reality: the task and the use of the pretrained model matter, and sometimes models pre-trained on seemingly broader distributions still follow the same linear trend as the models trained purely on in-distribution data.
We first present our findings for fine-tuning pretrained \imagenet models and
subsequently discuss results for zero-shot prediction. See
Appendix~\ref{app:more-data} for more experimental details.

\iftoggle{isarxiv}{}{\vspace{-0.8cm}}

\begin{figure*}
	\centering
\iftoggle{isarxiv}{
	 \includegraphics[height=5.5cm]{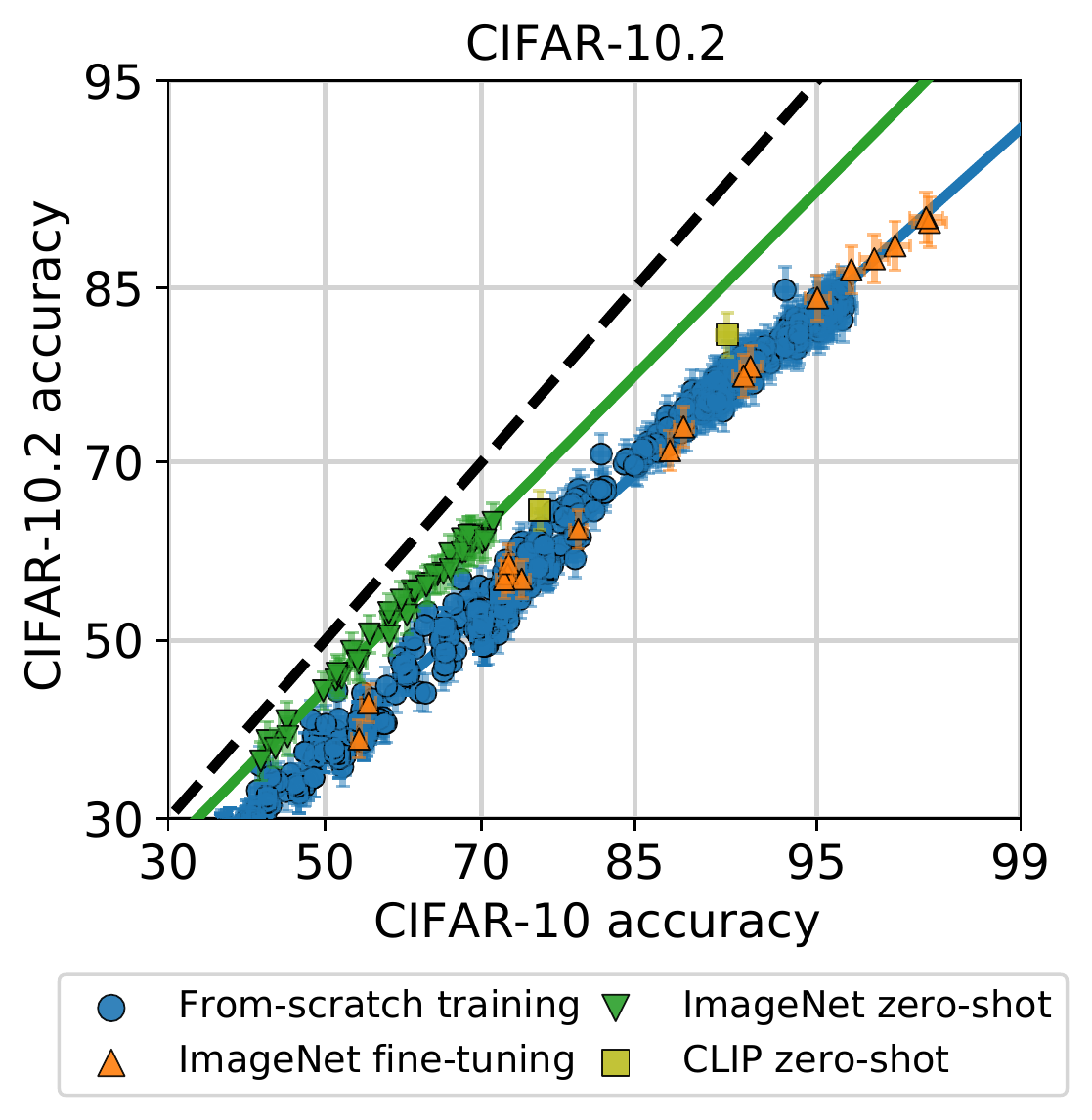}
	 \includegraphics[height=5.5cm]{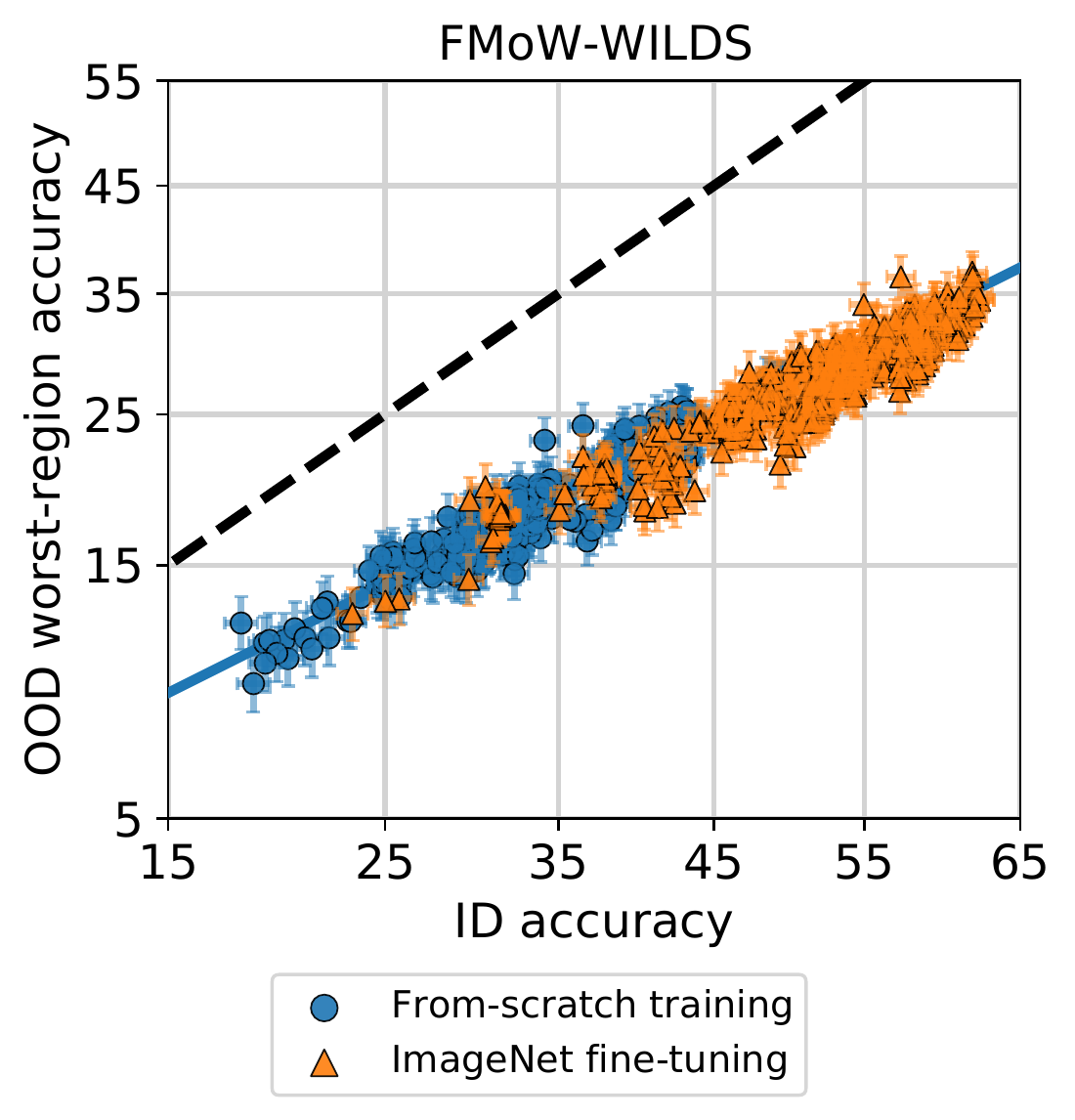}
	 \includegraphics[height=5.5cm]{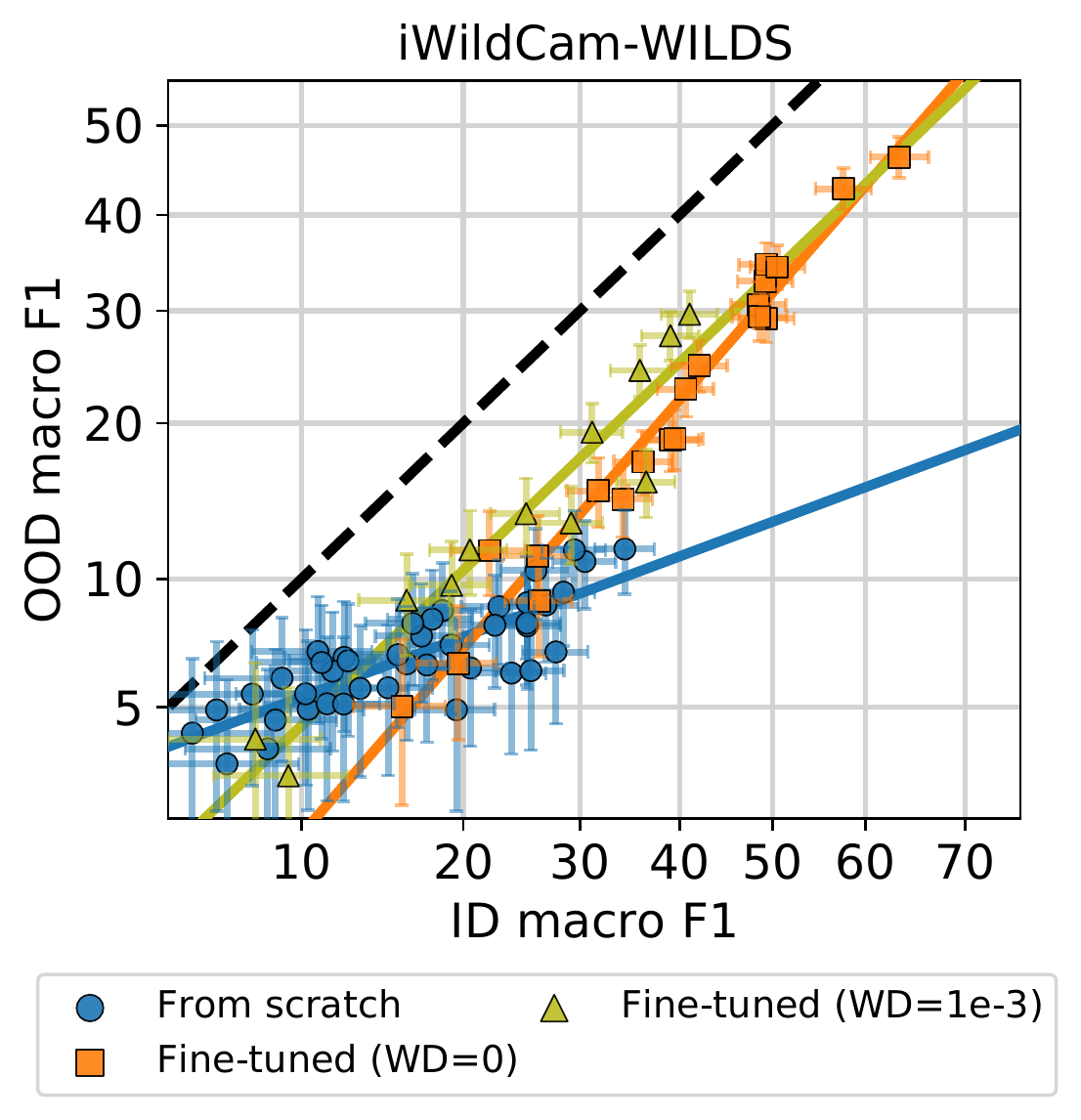}
 }{
	 \includegraphics[height=5.3cm]{figures/more_data_c10}
	 \includegraphics[height=5.3cm]{figures/more_data_fmow}
	 \includegraphics[height=5.3cm]{figures/more_data_iwc_v2}
     \vspace{-0.3cm}
    }
	\caption{The effect of pre-training with additional data on \cifartentwo{} (left), \fmow{} (middle), and \iwildcam{} (right).
    On \cifartentwo{} and \fmow, fine-tuning pretrained models moves the models
    along the predicted ID-OOD line.
    However, on \cifartentwo, zero-shot prediction using pretrained models deviates from this line.
    On \iwildcam, fine-tuning pretrained models changes the ID-OOD relationship observed for models trained from scratch. Moreover, the weight decay hyperparameter affects the ID-OOD relationship in fine-tuned models. 
  }
	\label{fig:more-data}
\end{figure*}

\paragraph{Fine-tuning pretrained models on ID data.}
Figure~\ref{fig:more-data} plots OOD performance vs.\ ID performance 
for models trained from-scratch (purely on ID data) and fine-tuned models whose initialization was pretrained on \imagenet.
Across the board, pretrained models attain better performance on both the ID and OOD test sets.
However, fine-tuning affects ID-OOD correlations differently across tasks.
In particular, for \cifarten reproductions and for \fmow, fine-tuning produces results that lie on the same ID-OOD trend as purely ID-trained models (Figure~\ref{fig:more-data} left and center).
On the other hand, a similar fine-tuning procedure yields models with a different ID-OOD relationship  on \iwildcam than models trained from scratch on this dataset.
Moreover, the weight decay used for fine-tuning seems to also affect the linear trend (Figure~\ref{fig:more-data} right).

One conjecture is that the qualitatively different behavior of fine-tuning on \iwildcam is related to the fact that \imagenet is a more diverse dataset that may encode robustness-inducing invariances that are not represented in the \iwildcam ID training set.
For instance, both \imagenet and \iwildcam contain high-resolution images of natural scenes, but the camera perspectives in \iwildcam may be more limited compared to ImageNet.
Hence ImageNet classifiers may be more invariant to viewpoint, which may aid generalization to previously unseen camera viewpoints in the OOD test set of \iwildcam.
On the other hand, the satellite images in \fmow are all taken from an overhead viewpoint, so learning invariance to camera viewpoints from ImageNet might not be as beneficial.
Investigating this and related conjectures (e.g., invariances such as lighting, object pose, and background) is an interesting direction for future work.

\iftoggle{isarxiv}{}{\vspace{-0.4cm}}

\paragraph{Zero-shot prediction on pretrained models.}
A common explanation for OOD performance drop is that training on the ID training set biases the model toward patterns that are more predictive on the ID test set than on its OOD counterpart. With that explanation in mind, the fact that fine-tuned models maintain the same ID/OOD linear trend as from-scratch models is surprising: once could reasonably expect that an initialization determined independently of either ID or OOD data would produce models that are less biased toward the former. Indeed, in the extreme scenario that no fine-tuning takes place, the model should have no bias toward either distribution, and we therefore expect to see a different ID/OOD trend.

The \cifarten allows us directly test this expectation directly by performing zero-shot inference on models pretrained on \imagenet: since the \cifarten classes form a subset of the \imagenet classes, we simply feed (resized) \cifarten images to these models, and limit the prediction to the relevant class subset.
The resulting classifiers have no preference for either the ID or OOD test set because they depend on neither distribution.
We plot the zero-shot prediction results in Figure~\ref{fig:more-data} (left) and observe that, as expected, they deviate from the basic linear trend.
Moreover, they form a different linear trend closer---but not identical---to $x=y$.
The fact that the zero-shot linear trend is closer to $x=y$ supports the hypothesis that the performance drop partially stems from bias in ID training.
However, the fact that this trend is still below $x=y$ suggests that the drop is also partially due to \cifarten reproductions being harder than \cifarten for current methods (interestingly, humans show similar performance on both test sets~\cite{recht2019imagenet,miller2020effect,humanaccuracy}).
These finding agree with prior work~\cite{lu2020harder}.

As another test of zero-shot inference, we apply two publically-available CLIP models on \cifarten by creating last-layer weights out of natural language descriptions of the classes~\cite{radford2021learning}. As Figure~\ref{fig:more-data} (left) shows, these models are slightly above the basic ID/OOD linear trend, but below the trend of zero-shot inference with \imagenet models. 

\iftoggle{isarxiv}{}{\vspace{-0.4cm}}

\paragraph{Additional experiments.}
In Appendix~\ref{app:more-data} we describe additional experiments
with pretrained models. To explore a middle ground between zero-shot prediction
and full-model fine-tuning, we consider a linear probe on CLIP for both
\cifarten and \fmow. For \cifarten, we also consider models trained on
a task-relevant subset of \imagenet classes~\cite{darlow2018cinic} and models
trained in a semi-supervised fashion using unlabeled data from 80 Million Tiny
Images~\cite{tinyimages,carmon2019unlabeled,augustin2020out}. Generally, we find
that, compared to zero-shot prediction, these techniques deviate less from the
basic linear trend. We also report results on additional OOD settings, namely
\cifartenone and different region subsets for \fmow, and reach similar
conclusions.

\section{Summary of empirical phenomena}
\label{sec:empirical-summary}
The previous sections have presented a variety of empirical phenomena concerning the relationship between in-distribution and out-of-distribution performance.
To summarize these phenomena, we now briefly highlight the key observations.
These observations will also guide the development of our theoretical model of distribution shift in the next section.

The key observations are:
\begin{enumerate}
    \item The linear trend between in-distribution and out-of-distribution performance applies to a wide
    range of model families and holds under variation in architecture,
    hyperparameters, and training duration (Section~\ref{sec:linear_trends}). 
    \item The linear trends are more precise after applying a probit or logit scaling on both axes of the scatter plotes. 
    \item Changing the amount of in-distribution training data does not affet 
    the linear trend (Section~\ref{sec:linear_trends-results}).
    On the other hand, pre-training on
    a different data distribution can -- but does not always -- yield a different trend
    (Section~\ref{sec:training_data}).
    \item Some distribution shifts show precise linear trends while others do so
    only for subsets of models or not at all (Section~\ref{sec:fit_failure}).
\end{enumerate}

\section{Theoretical models for linear fits}
\label{sec:theory}
In this section we propose and analyze a simple theoretical model that distills several of the empirical phenomena from the previous sections.
Our goal here is \emph{not} to obtain a general model that encompasses complicated real distributions such as the images in CIFAR-10.
Instead, our focus is on finding a simple model that is still rich enough to exhibit some of the same phenomena as real data distributions.

\subsection{A simple Gaussian distribution shift setting}\label{sec:theory-gaussian}

\newcommand{\acc}{\mathrm{acc}}
\renewcommand{\vec}[1]{\boldsymbol{#1}}
\newcommand{\vx}{\vec{x}}
\newcommand{\shiftvar}{\vec{\Delta}}
\newcommand{\meanvar}{\vec{\mu}}
\newcommand{\classifier}{\vec{\theta}}



We consider a simple binary classification problem where the label $y$ is distributed uniformly on $\{-1, 1\}$ both in the original distribution $D$ and shifted distribution $D'$. Conditional on $y$, we consider $D$ such that $\vx \in \R^d$ is an isotropic Gaussian, i.e.,\\[-.3cm]
\begin{equation*}
	\vx \, | \, y \; \sim \; \mc{N}( \meanvar \cdot y; \, \sigma^2 I_{d \times d} ),
\end{equation*}
for mean vector $\meanvar \in \R^d$ and variance $\sigma^2 > 0$.

We model the distribution shift as a change in $\sigma$ and $\meanvar$.
Specifically, we assume that the shifted distribution $D'$ corresponds to shifted parameters
\begin{equation}
  \label{eq:shift}
	\meanvar' = \alpha \cdot \meanvar + \beta \cdot \shiftvar
	~~\quad\mbox{and}\quad~~
	\sigma' = \gamma \cdot \sigma
\end{equation}
where $\alpha,\beta,\gamma>0$ are fixed scalars and $\shiftvar$ is \emph{uniformly distributed} on the sphere in $\R^d$. 
Note that in our setting $D'$ is a random object determined by the draw of $\shiftvar$.

Within the setup describe above, we focus on linear classifiers of the form $\vx \mapsto \sign ( \classifier^\top \vx )$. 
The following theorem states that, as long as $\classifier$ depends only on the training data and is \emph{thereby independent of the random shift direction} $\shiftvar$, the probit-transformed accuracies on $D$ and $D'$ have a near-linear relationship with slope $\alpha/\gamma$. (Recall that the probit transfrom is the inverse of the standard Normal cdf $\Phi(x)=\int_{-\infty}^x \frac{1}{\sqrt{2\pi}} e^{-t^2/2}\mathrm{d}t$).
The deviation from linearity is of order $d^{-1/2}$ and vanishes in high dimension. 

\begin{restatable}{theorem}{restateTheOneTheoremWeHave}\label{thm:theory}
	In the setting described above where $\shiftvar$ is independent of $\classifier$, let $\delta \in (0,1)$. 
	With probability at least $1-\delta$, we have
	\begin{equation*}
		\abs*{\Phi^{-1}\prn*{ \acc_{D'}(\classifier)} - \frac{\alpha}{\gamma} \, \Phi^{-1}\prn*{ \acc_{D}(\classifier)}} \; \le \; \frac{\beta}{\gamma \sigma}\sqrt{\frac{2\log\sfrac{2}{\delta}}{d}} \; .
	\end{equation*}
\end{restatable}
\noindent
The theorem is a direct consequence of the concentration of measure;
see proof in Appendix~\ref{app:theory-proof}.


We illustrate Theorem~\ref{thm:theory} by  simulating its setup and training different linear classifiers by varying the loss function and regularization. Figure~\ref{fig:theory-main} shows good agreement between the performance of linear classifiers and the theoretically-predicted linear trend. Furthermore,  conventional nonlinear classifiers (nearest neighbors and random forests) also satisfy the same linear relationship, which does not directly follow from our theory. Nevertheless, if the decision boundary of the nonlinear becomes nearly linear in our setting a similar theoretical analysis might be applicable. Our simple Gaussian setup thus illustrates how linear trends can arise across a wide range of models. 

\begin{figure}
	\centering
    \iftoggle{isarxiv}{
        \includegraphics[height=5cm]{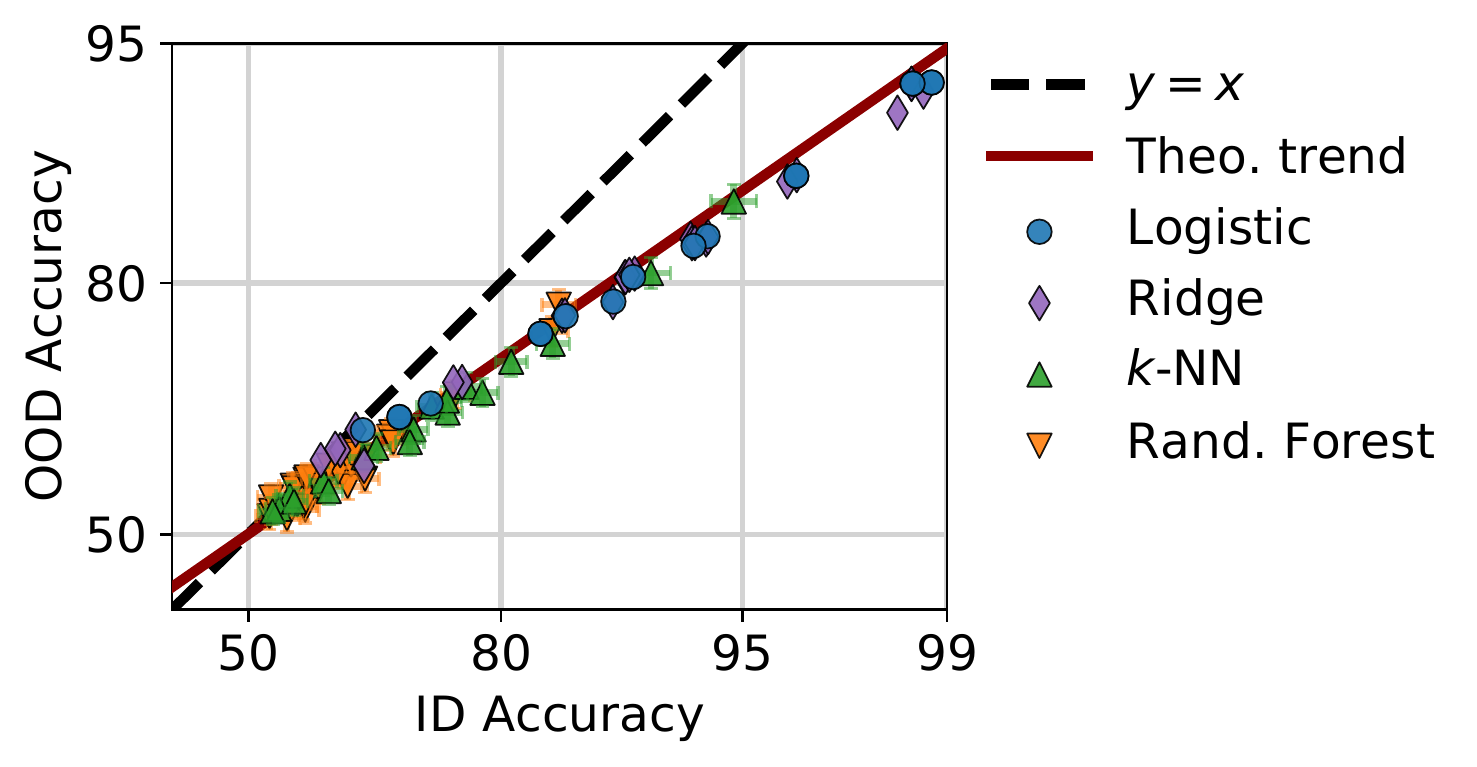}
    }{
        \includegraphics[width=\columnwidth]{figures/theory_main}
	    \vspace{-.5cm}
    }
	\caption{Illustration of the theoretical distribution shift model in Section~\ref{sec:theory-gaussian} with $d=10^{5}$, $\alpha=0.7$, $\beta=0.5$ and $\gamma=1$ (see Appendix~\ref{app:theory-figure-details} for details).
	The accuracies for linear models (logistic and ridge regression) agree with the prediction of Theorem~\ref{thm:theory}. Moreover,
	nonlinear models (nearest neighbors and random features) exhibit the same
    probit trend we prove for linear classifiers. \vspace{-.3cm}
	}
	\label{fig:theory-main}
\end{figure}

\subsection{Modeling departures from the linear trend}\label{sec:theory-departures}
In the previous section, we identified a simple Gaussian setting that showed linear fits across a large range of models. Now we discuss small changes to the setting that break linear trends and draw parallels to the empirical observations on complex datasets presented in this paper. In Appendix~\ref{app:theory-departures}, we discuss each of these modifications in further detail.

\textbf{Adversarial distribution shifts.} Previously, the direction $\shiftvar$ which determines the distribution shift as defined above in eq.~\eqref{eq:shift}, was chosen independent of the tested models $\classifier_1, \ldots, \classifier_k$. However, when $\shiftvar$ is instead chosen by an adversary with knowledge of the tested models, the ID-OOD relationship can be highly non-linear.
This is reminiscent of adversarial robustness notions where models with comparable in-distribution accuracies can have widely differing adversarial  accuracies depending on the training method.

\textbf{Pretraining data.} Additional training data from a \emph{different}
distribution available for pretraining could contain information about the shift
$\shiftvar$. In this case, the pretrained models are not necessarily independent
of $\shiftvar$ and these models could lie above the linear fit of classifiers
without pretraining. See Section~\ref{sec:training_data} for a discussion of when such behavior arises in practice. 

\textbf{Shift in covariance.} Previously, we assumed that $\vx \mid y$ is always an isotropic Gaussian. Instead consider a setting where the original distribution is of the form $\vx | y \sim \mc{N}( \meanvar y; \Sigma)$ where $\Sigma$ is not scalar (i.e., has distinct eigenvalues). Then, the linear trend breaks down even when the distribution shift is simple additive white Gaussian noise corresponding to $\vx | y \sim \mc{N}( \meanvar y; \Sigma + (\sigma')^2 I_{d\times d})$. For example, ridge regularization turns out to be an effective robustness intervention in this setting. However, if the shifted distribution is of the form $\vx | y \sim \mc{N}( \meanvar y; \gamma \Sigma)$ for some scalar $\gamma>0$, it is straightforward to see that a linear trend holds.

These theoretical observations suggest that a covariance change in ID/OOD the distribution shift could be a possible explanation for some departures from the linear trends such as additive Gaussian noise corruptions in \cifartenc.  To test this hypothesis, we created a new distribution shift by corrupting \cifarten with noise sampled from the same covariance as the original \cifarten distribution. As discussed in Section~\ref{ssec:cifarten-corrupted}, we find that the correlation between ID and OOD accuracy is substantially higher with the covariance-matched noise than with  isotropic Gaussian noise with similar magnitude.

While the theoretical setting we study in this work is much simpler than real-world distributions, the analysis sheds some light on when to expect linear trends and what leads to departures. Ideally, a theory would precisely explain what differentiates CIFAR-10.2, CINIC-10, and the CIFAR-10-C-Fog shift (see Figure \ref{fig:main_figure}) where we see linear trends from simply adding Gaussian noise to the images as in CIFAR-10-C-Gaussian where we do not observe linear trends. A possible direction may be to characterize shifts by their generation process, and we leave this to future work. 


\section{Related work}
\label{sec:related_work}
Due to the large body of research on distribution shifts, domain adaptation, and reliable machine learning, we only summarize the most directly related work here.
Appendix~\ref{app:related-work} contains a more detailed discussion of related work.

\textbf{Domain generalization theory.}
Prior work has theoretically characterized the performance of classifiers under
distribution shift.~\citet{bendavid2006analysis} provided the first
VC-dimension-based generalization bound. They bound the difference between a
classifier's error on the source distribution ($D$) and target distribution
($D'$) via a classifier-induced divergence measure.~\citet{mansour2009domain}
extended this work to more general loss functions and provided sharper
generalization bounds via Rademacher complexity. These results have been
generalized to include multiple sources~\citep{blitzer2007learning,
hoffman2018algorithms, mansour2008domain}. See the survey
of~\citet{redko2020survey} and the references therein for further discussion of
these results.  The philosophy underlying these works is that robust models
should aim to minimize the induced divergence measure and thus guarantee similar
OOD and ID performance.

The linear trends we observe in this paper are not captured by such analyses. As illustrated in Figure 1 (top-left), the bounds described above can only state that OOD performance is highly predictable from ID performance if they are equal (i.e., when the gray region is tight around the $x=y$ line). In contrast, we observe that OOD performance is \emph{both} highly predictable from ID performance and significantly different from it.
Our Gaussian model in Section~\ref{sec:theory-gaussian} demonstrates how the linear trend phenomenon can come about in a simple setting. However, unlike the above-mentioned domain generalization bounds, it is limited to particular distributions and the hypothesis class of linear classifiers.

\citet{mania2020classifier} proposed a condition that implies an approximately linear trend between ID and OOD accuracy, and empirically checked their condition in dataset reproduction settings.
The condition is related to model similarity, and requires the probability of certain multiple-model error events to not change much under distribution shift.
An interesting question for future work is whether their condition can shed light on what distribution shifts show linear trends, and what axes transformations lead to the most precise trends.

\textbf{Empirical observations of linear trends.}
Precise linear trends between in-distribution and out-of-distribution generalization were first discovered in the context of dataset reproduction experiments.
\citet{recht2018cifar10, recht2019imagenet,qmnist,miller2020effect} constructed new test sets for CIFAR-10 \cite{krizhevsky2009learning}, ImageNet \cite{deng2009imagenet,russakovsky2015imagenet}, MNIST \cite{mnist}, and SQuAD \cite{rajpurkar2016squad} and found linear trends similar to those in Figure 1.

However, these studies were limited in their scope, as they only focused on dataset reproductions.
\citet{taori2020measuring} later showed that linear trends still occur
for~\imagenet models on datasets like ObjectNet, Vid-Robust, and
YTBB-Robust~\citep{barbu2019objectnet,shankar2019image}.  On \imagenet,
\citet{shankar2020evaluating} showed that linear trends also occur between the original
top-k accuracy metrics and a multi-label accuracy metric based on
a new set of multi-label annotations.  All of these experiments, however, were
limited to \imagenet or \imagenet-like tasks.
We significantly broaden the scope of the linear trend phenomenon by including a
range of additional distribution shifts such as \cinicten, \stlten, \fmow, and
\iwildcam, as well as identifying negative examples like \camelyon and
some \cifartenc shifts.
In addition, we also include a pose estimation task with \ycb.
The results show that linear trends not only occur in academic 
benchmarks but also in distribution shifts coming from applications ``in the wild.''
Furthermore, we show that linear trends hold across different learning approaches, training durations, and hyperparameters.

\citet{kornblith2019better} study linear fits in the context of transfer
learning and train or fine-tune models on the distribution corresponding to the
y-axis in our setting. On a variety of image classification tasks, they show
that a model's \imagenet test accuracy linearly correlates with the model's
accuracy on the new task after fine-tuning.
The similarity between their results
and ours suggest that they may both be part of a broader
phenomenon of predictable generalization in machine learning.

In concurrent work, \citet{andreassen2021evolution} study the impact of fine-tuning on the effective robustness of pre-trained models, i.e., how much a pre-trained model is above the linear trend given by models trained only on in-distribution data \citep{taori2020measuring}.
At a high level, this investigation is similar to Section \ref{sec:training_data} in our paper, but the datasets are complementary: \citet{andreassen2021evolution} focus on distribution shifts in the context of \cifarten and \imagenet while we also study the effect of pre-training on the three distribution shifts from WILDS (\fmow, \iwildcam, and \camelyon).
\citet{andreassen2021evolution} measure how the effective robustness of a model evolves during fine-tuning and find that models gain accuracy but lose effective robustness over the course of fine-tuning.
In addition, \citet{andreassen2021evolution} investigate how data diversity in
the pre-training data along with other factors like model size impact the
effective robustness achieved by fine-tuning.
%
%
%
%

\section{Discussion}
\label{sec:discussion}
Initial research on dataset reproductions found that many neural networks follow a linear trend in scatter plots relating in-distribution to out-of-distribution performance.
Our paper and concurrent work on object detection, natural language processing,
and magnetic resonance imaging~\citep{caine2021pseudo, liu2021can, darestani2021measuring} 
show that such linear trends are not a peculiarity of dataset reproductions but occur for many types of models and distribution shifts.
The striking regularity of these trends raises the possibility that for certain classes of distribution shifts and models, out-of-distribution performance is solely a function of in-distribution performance.

While our paper and related work give many examples of model types and distribution shifts with such a universal trend, our experiments have also demonstrated datasets where linear trends do \emph{not} occur.
This naturally raises the question for what model types and distribution shifts out-of-distribution performance is a function of in-distribution performance.
As a starting point for future work in this direction, we now formalize the precise relationship between in-distribution and out-of-distribution performance as \emph{correlation property}.

\begin{definition}[Correlation property]
A pair of distributions $D$, $D'$ and a family of models $\mathcal{M}$ have the $\alpha$-approximate correlation property under loss function $\ell$ (e.g., accuracy) and monotone transform function $\gamma : \R \rightarrow \R$ (e.g., a linear function in probit domain) if for all models $f \in \mathcal{M}$ we have
\[
    \left| \, \gamma\left(\mathop{\mathbb{E}}_{x, y \sim D} \left[ \ell(f(x), y) \right] \right) \; - \; \mathop{\mathbb{E}}_{x, y \sim D'} \left[ \ell(f(x), y) \right] \, \right|  \; \leq \; \alpha \; .
\]
    
\end{definition}

With the correlation property in place, we can now state a candidate hypothesis for specific distribution shifts such as the shift from \cifarten to \cifartenone.
\begin{conjecture}
\label{conj:main}
    The distribution shift from \cifarten to \cifartenone (or \imagenet to \imagenettwo, or \fmow, etc.) has the 1\%-approximate correlation property with loss function accuracy and transform
    \[
        \gamma(l) \; = \; \Phi\left(\text{slope} \cdot \Phi^{-1}(l) + \text{offset} \right)
    \]
    for models that are trained with empirical risk minimization (ERM) on the respective training distribution.
\end{conjecture}
The restriction to models trained on in-distribution data is important since for instance \imagenet models applied to \cifarten in a zero-shot manner follow a different trend (recall Section \ref{sec:training_data}).
Our experiments with a wide range of ERM models give evidence for these conjectures, but are not ultimate proof.
An interesting direction for future work is understanding which distributions shifts have the correlation property for ERM models.

\begin{question}
    \label{q:dist}
    What pairs of distributions $D$, $D'$ have the correlation property for models trained via empirical risk minimization only on $D$?
\end{question}

Beyond exploring the data distribution question, another dimension of the correlation property is what models it applies to.
There is evidence that statements similar to Conjecture \ref{conj:main} hold for a wider range of models than ERM.
For instance, nearest neighbor models follow the same trend as ERM models on several distribution shifts studied in this paper (see Figure \ref{fig:main_figure}).
Moreover, earlier work found that a wide range of robustness interventions (e.g., adversarial training, data augmentation, filtering layers, distributionally robust optimization) do not improve over ERM baselines on the \imagenettwo, ObjectNet, and \fmow distribution shifts \citep{taori2020measuring,koh2020wilds,lostgeneralization}.
On the other hand, it is also easy to construct models that do \emph{not} follow the same trend as the ERM baseline.
Interpolating between a \cifarten model on the linear trend and a random classifier yields models above the linear trend.\footnote{To interpolate between the two classifiers, draw a sample from a Bernoulli with success probability $p$. If the sample is 0, classify with the original classifier. Otherwise classify with the random classifier. Varying $p$ in $[0, 1]$ interpolates between the two classifiers.} 
This leads to the following question:


\begin{question}
    \label{q:model}
    For the \cifarten $\rightarrow$ \cifartenone distribution shift (or \imagenet $\rightarrow$ \imagenettwo, or \fmow, etc.), what models satisfy the correlation property?
\end{question}

Beyond interpolating with a random classifier, we are currently not aware of models that violate the correlation property on the aforementioned distribution shifts by a substantial amount.
Interpolating with a random classifier is a peculiar intervention since it decreases \emph{both} in-distribution and out-of-distribution performance.
Hence a stronger version of Conjecture \ref{conj:main} that applies to a wider range or even all ``useful'' models trained only on in-distribution data is plausible.
However, we currently do not have a satisfying way to make this stronger conjecture precise.
We hope that future work further investigates Questions \ref{q:dist} and \ref{q:model} (and their combination) both empirically and theoretically to shed light on the relationship between in-distribution and out-of-distribution generalization.

\subsection{Possible implications}
If the correlation property holds for relevant distribution shifts and models, it can be a valuable guide for building reliable machine learning systems.
An important point here is that -- at least empiricially -- the correlation property often holds not only for a single pair of distributions, but for an entire range of distribution shifts (e.g., from \cifarten to \cifartenone, \cifartentwo, \cinicten, and \stlten or for temporal and spatial distribution shifts in \fmow).
So when a practitioner encounters a linear trend between in-distribution and out-of-distribution performance, it is reasonable to expect that similar distribution shifts will also exhibit a linear trend.

We now briefly describe three implications that arise when the correlation property holds for a range of distribution shifts.
We note that these implications are conditional and more research is needed to understand their extent.
\begin{itemize}
    \item \textbf{Model selection.} Practitioners are often faced with the challenge of selecting a model that performs well not only on a specific test set but also on future unseen data that may come from different distributions.
    If the shifted distributions have the correlation property with respect to the training distribution, selecting the best model under these distribution shifts reduces to selecting the best model on the in-distribution test set.

    \item \textbf{Baseline for measuring out-of-distribution robustness.} A central goal of research on reliable machine learning is to develop models that perform well on out-of-distribution data.
    There are two natural ways to quantify this goal: (i) performance on out-of-distribution data, and (ii) the gap between in-distribution and out-of-distribution performance.

    The correlation property for empirical risk minimization implies that optimizing for in-distribution performance also provides corresponding gains in out-of-distribution performance.
    Hence existing work on improving in-distribution performance already improves the robustness of a model according to criterion (i) without explicitly targeting robustness.
    So if a proposed training technique claims to improve the robustness of a model as a quantity distinct from in-distribution performance, the proposed technique should not only improve out-of-distribution performance, but also reduce the gap between in-distribution and out-of-distribution performance -- criterion (ii) -- beyond what current methods optimizing for in-distribution performance achieve.
    Graphically in terms of our scatter plots, the proposed technique promoting robustness should produce a model that lies \emph{above} the linear trend given by empirical risk minimization.\footnote{Improving the out-of-distribution performance of a model by improving its in-distribution performance is also clearly a valid way to improve robustness according to criterion (i). But the proposed technique should then be compared to existing methods for improving in-distribution performance (architecture variations, training schedules, etc.) and ideally improve over the out-of-distribution performance achieved by state-of-the-art methods for in-distribution performance.}

    To better compare new training techniques to prior work in terms of robustness, we recommend that papers illustrate the effect of their technique with a scatter plot of relevant models (e.g., the evaluations in \citet{taori2020measuring}, our paper, and Section 3.3 of \citet{radford2021learning}).
    In addition, papers should report \emph{both} in-distribution and out-of-distribution performance of their technique so that the effect on both quantities is clear.
    We also refer the reader to \citet{taori2020measuring}, who formalize the concept of ``robustness beyond a baseline'' as ``effective robustness''.


    \item \textbf{Guide for algorithmic interventions to improve robustness.} If we can characterize for what set of training approaches the correlation property holds, research aiming to decrease the gap between in-distribution and out-of-distribution performance can focus on other approaches.
    For instance, our experiments suggest that architecture variations in neural network may not affect the gap between in-distribution and out-of-distribution performance, but better pre-training datasets can at least sometimes reduce this gap.
\end{itemize}

\subsection{Frequently asked questions}
In conversations with colleagues and reviewers, certain questions about our work appeared repeatedly.
To clarify our perspective on these issues, we answer the three most common questions below:

\paragraph{Q:} \textbf{Do all distribution shifts have linear trends?} \\[.2cm]
No.
Section \ref{sec:fit_failure} gives concrete examples of distribution shifts that do not follow a linear trend.
Before our work, it was already clear that adversarial distribution shifts such as $\ell_\infty$-robustness do not follow a linear trend because models trained with ERM usually have little to no robustness to adversarial examples while multiple approaches give non-trivial robustness, e.g., \citep{madry2017towards,randomizedsmoothing,certifiedaditi,pmlr-v80-wong18a}.
Since multiple non-adversarial and natural (i.e., not synthetically constructed) distribution shifts \emph{do} show linear trends for a wide range of models, the main question is \emph{what} distribution shifts have linear trends.

\paragraph{Q:} \textbf{Should we only work on improving in-distribution performance?} \\[.2cm]
No.
As mentioned above, not all distribution shifts have linear trends.
Moreover, more work is needed to understand whether new robustness interventions can improve over the linear trends observed for empirical risk minimization and existing robustness interventions \citep{taori2020measuring,koh2020wilds}.
In addition, Section \ref{app:more-data} suggests pre-training as a promising direction for improving out-of-distribution performance, as also shown by the recent CLIP model \citep{radford2021learning}.

\paragraph{Q:} \textbf{Is it is possible to construct models that violate the linear trend?}\\[.2cm]
Yes, when the linear trend is on a non-linear transformation of the accuracy such as the probit transform we use in this paper.
This is due to the fact that we can always construct a family of models with a \emph{linear} ID/OOD performance relationship (without the probit transform) by randomly switching between two base models.

Concretely, let $f$ be a model with non-trivial performance and consider the following \emph{interpolations} between $f$ and a trivial random classifier: given input $x$, output $f(x)$ with probability $p$ and output a random class label with probability $1-p$.
Let $C$ be the number of classes and let $\mathrm{acc}_D(f)$ and $\mathrm{acc}_{D'}(f)$ be the in-distribution and out-of-distribution accuracies of $f$, respectively.
Then, as we vary $p$ from $0$ to $1$, the in- and out-of-distributions accuracies of the interpolating model trace a line from $(1/C, 1/C)$ to $(\mathrm{acc}_D(f), \mathrm{acc}_{D'}(f))$.
Therefore, if we apply a non-linear transformation to these accuracies (such as a probit transform), the in- and out-of-distribution performance of these models no longer follows a linear trend.
Characterizing for which models the linear trend (approximately) holds is an important direction for future work.

%

\section*{Acknowledgements}
We would like to thank
Sara Beery,
Moritz Hardt,
Gabriel Ilharco,
Daniel Levy,
Mike Li,
Horia Mania,
Yishay Mansour, 
Henrik Marklund,
Hongseok Namkoong,
Ben Recht, 
Max Simchowitz,
Kunal Talwar, 
Nilesh Tripuraneni,
Mitchell Wortsman,
Michael Xie,
Steve Yadlowsky,
and 
Bin Yu
for helpful discussions while working on this paper.
We also thank
Benjamin Burchfiel,
Eric Cousineau,
Kunimatsu Hashimoto,
Russ Tendrake,
and Vickie Ye 
for assistance and guidance in setting up the \ycb testbed.

This work was funded by an Open Philanthropy Project Award.
JM was supported by the National Science Foundation Graduate Research
Fellowship Program under Grant No. DGE 1752814.
RT was supported by the National
Science Foundation Graduate Research Fellowship Program under Grant No. DGE
1656518.
AR was supported by the Google PhD Fellowship and the Open Philanthropy AI Fellowship.
SS was supported by the Herbert Kunzel Stanford Graduate Fellowship.
YC was supported by Len Blavatnik and the Blavatnik Family foundation, and the Yandex Machine Learning Initiative for Machine Learning.

\bibliographystyle{plainnat}
\bibliography{linearfits}

\clearpage
\newpage

\onecolumn

\appendix

\etocdepthtag.toc{mtappendix}

\iftoggle{isarxiv}{
\etocsettagdepth{mtsection}{none}
\etocsettagdepth{mtappendix}{subsubsection}
\tableofcontents
}{
}

\section{Experimental testbeds}
\label{app:testbed}
A rigorous empirical investgation of the correlation between in-distribution and out-of-distribution performance requires a broad set of experiments. To measure the behavior of many models on a variety of datasets, we utilized
three different experimental ``testbeds.'' A testbed consists of a collection of
one or more ``dataset universes'' and a compatible set of models that can be
trained and evaluated on these ``universes.'' Each dataset universe itself
consists of a training set (e.g. \cifarten train), an in-distribution test-set
(\cifarten test), and a collection of out-of-distribution test-sets (e.g.
\cifartentwo, \cifartenc, etc). Within a universe, models trained on one dataset
can be tested on all other datasets, with each test set representing a different
distribution. The three testbeds we use are:
\begin{enumerate}
    \item A new custom-built test for experiments with~\cifarten and WILDS (\fmow, \camelyon, and \iwildcam)
    \item An \imagenet testbed based on \citet{taori2020measuring}, and
    \item A testbed for pose estimation in the context of the \ycb dataset \citep{calli2015benchmarking}.
\end{enumerate}

In the rest of this section, we first detail the custom-built \cifarten and
WILDS testbed since it forms the basis for most experiments in this paper.  We
then describe our modifications to the \imagenet testbed of
\citet{taori2020measuring} in Section \ref{app:imagenettestbed}, and finally we
describe our testbed for \ycb in Section \ref{app:ycbtestbed}.

\subsection{\cifarten and WILDS testbed}
\label{app:testbed_datasets}
We now describe the datasets in our main testbed and summarize the models it contains.
Our main testbed contains four distinct ``universes.'' Each universe consists of
at least three datasets that we use for training and testing models both
in-distribution and out-of-distribution.

The four universes are \cifarten, \fmow, \camelyon, and \iwildcam, which we now describe in more detail. The latter three datasets are taken from the WILDS benchmark \citep{koh2020wilds}, and we use the  train/test splits and  evaluation procedures therein.

\subsubsection{\cifarten and related datasets}
The \cifarten universe comprises $32 \times 32$ pixel color images used in an image classification task.
The ten classes are airplane, automobile, bird, cat, deer, dog, frog, horse, ship, and truck.
The \cifarten universe contains the following datasets:

\begin{itemize}
    \item \textbf{\cifarten} is the main dataset in the \cifarten universe and was introduced by \citet{krizhevsky2009learning}.
    \cifarten is derived from the larger Tiny Images dataset \cite{tinyimages}.
    Since its introduction, \cifarten has become one of the most widely used image classification benchmarks.

    \item \textbf{\cifartenone} is a reproduction of the \cifarten dataset.
    \citet{recht2019imagenet} closely followed the dataset creation process of \cifarten and assembled a new dataset also using Tiny Images as a source.
    \cifartenone contains only about 2,000 images and is therefore usually used only as a test set.
    The distribution shift from \cifarten to \cifartenone poses an interesting challenge since many parameters of the data generation process are held constant but a standard ResNet model still sees an 8 to 9 percentage points accuracy drop.

    \item \textbf{\cifartentwo} is a second reproduction of the \cifarten dataset.
    \citet{lu2020harder} again closely followed the dataset creation process of
    \cifarten to assemble a new dataset from Tiny Images, this time with
    different annotators compared to \cifartenone. \cifartentwo contains 12,000
    images with a suggested split into 10,000 training images and 2,000 test
    images. We conduct all of our experiments using the 2,000 image test set.
    Similar to \cifartenone, \cifartentwo is a distribution shift arising from changes in the filtering process conducted by the human annotators.

    \item \textbf{\cinicten} \citep{darlow2018cinic} is a dataset in \cifarten format that supplements \cifarten with additional images from the full \imagenet dataset (not only the 2012 competition set).
    In total, \cinicten contains 270,000 images.
    Here, we limit \cinicten to the images coming from \imagenet in order to keep the distribution more clearly separate from \cifarten.
    The resulting test set has size 70,000.
    \cinicten represents a distribution shift because the source of the images changed from Tiny Images to \imagenet.

    \item \textbf{\stlten} \citep{coates2011analysis} is another \cifarten-inspired dataset derived from \imagenet.
    Since the focus of \stlten is unsupervised learning, the dataset contains 100,000 unlabeled and 13,000 labeled images.
    We only use the labeled subset because we are mainly interested in \stlten as a test set with distribution shift (as in \cinicten, the data source changed from Tiny Images to \imagenet).
    The class structure of \stlten is slightly different from the \cifarten classes: instead of the frog class, \stlten contains a monkey class.
    When experimenting with \stlten, we therefore limit the dataset to the remaining nine classes.
    This yields an overall test set size of 11,700.

    \item \textbf{\cifartenc} contains a range of synthetic distribution shifts derived from \cifarten.
    \citet{hendrycks2018benchmarking} created \cifartenc by applying perturbations such as Gaussian noise, motion blur, or synthetic weather patterns (fog, snow, etc.) to the \cifarten test set.
    In total, \cifartenc contains 19 different perturbations, each with five different severity levels.
\end{itemize}

\subsubsection{\fmow}
In \fmow, which is adapted from the Functional Map of the World dataset \citep{christie2018functional}, the task is to classify land or building use from satellite images taken in different geographical regions (Africa, Americas,
Oceania, Asia, and Europe) and in different years.
Specifically, the input is an RGB satellite image and the label is one of 62 different land or building use categories (e.g., `shopping mall' or `road bridge').

The training set comprises 76,863 images taken around the world between 2002 and 2013. The in-distribution test set comprises 11,327 images from the same distribution, i.e., also taken around the world between 2002 and 2013, and we evaluate models by their average accuracy.
The out-of-distribution validation set comprises 19,915 images taken around the
word between 2013 and 2016, and the out-of-distribution test set consists of
22,108 images taken around the world between 2016 and 2018.

We evaluate models out-of-distribution by either their average accuracy or their worst accuracy over all five geographical regions.
When evaluating models using their worst-region accuracy, the out-of-distribution test set reflects both a distribution shift across time (from 2002--2013 to 2016--2018) and across regions (from images that are distributed across the world to images that are only from a given region).
In our experiments, the worst-performing region is generally Africa, which has the second smallest number of training examples, ahead of Oceania.

\subsubsection{\camelyon}
In \camelyon, which is a patch-based variant of the CAMELYON17 dataset
\citep{bandi2018camelyon}, the task is to classify whether a given patch of tissue contains any tumor tissue. Specifically, the input is a 96 $\times$ 96 patch of tissue extracted from a whole-slide image (WSI) of a breast cancer metastasis in a lymph node section, and the label is whether any pixel in the central 32 $\times$ 32 region of the patch has been labeled as part of a tumor in the ground-truth pathologist annotations.

The training set comprises 302,436 patches taken from 30 WSIs across 3 hospitals (10 WSIs per hospital). The in-distribution test set comprises 33,560 patches taken from the same set of 30 WSIs; this corresponds to the ``in-distribution validation set'' in the WILDS benchmark \citep{koh2020wilds}.
The out-of-distribution test set comprises 85,054 patches taken from 10 WSIs from a different hospital.
All of the above sets are class-balanced.
We evaluate models by their average accuracy; performance on the out-of-distribution test set reflects a model's ability to generalize to different hospitals from the ones it was trained on.

\subsubsection{\iwildcam}\label{app:datasets-iwc}
In \iwildcam, which is adapted from the iWildCam 2020 Competition Dataset
\citep{beery2020iwildcam}, the task is to classify which animal species (if any) is present in a camera trap photo. Specifically, the input is a (resized) $448\times 448$ pixel color image from a camera trap, and the label is one of 182 animal species (including ``no animal'').

 The training set comprises 129,809 images taken by 243 camera traps; the in-distribution test set comprises 8,154 images taken by those same 243 camera traps; and the out-of-distribution training set comprises 42,791 images taken by 48 different camera traps.
 As images taken by different camera traps can vary greatly in terms of camera angle, illumination, background, and animal distribution, the performance on the out-of-distribution test set reflects a model's ability to generalize to different camera traps from the ones it was trained on.

 While we study the current version of \iwildcam unless noted otherwise, we also study an earlier version of \iwildcam with a different in-distribution train-test split in \ref{app:linear-trends-iwc}.

\paragraph{Evaluation metric and confidence interval calculation.}
 Following~\citet{koh2020wilds}, we evaluate models by their macro F1 score, as this better captures model performance on rare species. Macro F1 is the average of the per class F1 scores for all classes appearing in the test data.
 We obtain confidence interval for this metric using the following heuristic. Suppose class $i$ has empirical F1 score $f_{i}$ and $n_i$ examples in the test set. As an approximate confidence interval for the F1 score of this class, we consider $[f_i - \delta_i, f_i+\delta_i]$ where $\delta_i$ is such that $[0.5-\delta_i,0.5+\delta_i]$ is a 95\% Clopper-Pearson confidence interval for a Bernoulli success probability given $n_i/2$ positive observation out of $n_i$ total observations. The size of this confidence interval is guaranteed to be larger than the size of the confidence intervals for both recall and precision for this class. Since the F1 score is the harmonic mean of recall and precision, the interval should provide adequate coverage for the F1 score as well. Finally, we combine the per class intervals to obtain a macro F1 confidence interval of the form $[\bar{f}-C^{-1/2}\bar{\delta},\bar{f}+C^{-1/2}\bar{\delta}]$, where $C$ is the number of classes in the test data and $\bar{f},\bar{\delta}$ are the averages of $f_1,\ldots,f_C$ and $\delta_1,\ldots, \delta_C$, respectively. This expression makes the approximation that individual F1 estimates are independent, which is not entirely accurate because the per-class precision estimates rely on overlapping samples.

 As Figures~\ref{fig:main_figure} and \ref{fig:iwc-arch} show, the confidence intervals computed with the above heuristic are fairly large. This may be in part due to a somewhat pessimistic approximation of the individual F1 confidence intervals, but it also reflects the fact that many of the \iwildcam classes are very rare (with 10 or fewer examples in the test set), and we simply do not have a good estimate for how models perform on them. The high level of class rarity was also the reason we chose not to use a bootstrap confidence interval: re-sampling the test sets with replacement leads to entire classes being dropped and biases the macro F1 estimates.

\paragraph{Label noise reduction.}
 The \iwildcam labels contain errors that stem primarily due to the fact they are derived from video-level annotation indicating whether a motion-activation event contains a particular animal. As consequence, many video frames that are in fact empty (showing no animal) are mislabeled as containing an animal that appeared in a temporally adjacent frame. To reduce this noise, we used auxiliary the auxiliary MegaDetector data provided as part of iWildCam 2020 Competition Dataset. More specifically, we performed our evaluation only on frames that were either labeled as empty or contained a MegaDetector detection with confidence at least 0.95.\footnote{
We still performed the model \emph{training} using precisely the same data, splits and labels as~\citet{koh2020wilds}; the filtering step was done at the evaluation stage only.} This filtering step provided a modest improvement to strength of the observed correlations (with $R^2$ increasing by one or two points).

\subsubsection{Models for \cifarten and WILDS Experiments}
\label{app:testbed_models}
To probe how widely the linear trend phenomena apply, we integrated a large number of classification models into our testbed.
At a high level, we divide these models into two types: deep neural networks (predominantly convolutional neural networks) and classical approaches. Due to the wide range of neural network architectures and training approaches emerging over the past decade, we further subdivide the neural network models based on their training set.

\paragraph{Convolutional neural networks for CIFAR-10.}
We integrated the following model architectures into our testbed.
Unless noted otherwise, we used the implementations from \url{https://github.com/kuangliu/pytorch-cifar}.
The models span a range of manually designed architectures and the results of automated architecture searches.
We refer the reader to the respective references for details about the individual architectures.
\begin{itemize}
    \item \textbf{DenseNet}, with depths 121 and 169~\citep{huang2017densely}.
    \item \textbf{Dual Path Networks (DPN)}, with depths 26 and 92~\citep{chen2017dual}.
    \item \textbf{EfficientNet}, specifically the B0 variant~\citep{tan2019efficientnet}.
    \item \textbf{GoogLeNet}, a member of the Inception family~\citep{szegedy2015going}.
    \item \textbf{MobileNet}, both the original and the MobileNetV2 variant~\citep{sandler2018mobilenetv2}.
    \item \textbf{MyrtleNet}, which are optimized for particularly fast training times. The code for these networks is from \url{https://github.com/davidcpage/cifar10-fast}.
    \item \textbf{PNASNet}, both A and B variants~\citep{liu2018progressive}.
    \item \textbf{RegNet}, configurations X\_200, X\_400, and Y\_400~\citep{radosavovic2020designing}.
    \item \textbf{ResNet}, varying the number of layers (18, 34, 50, and 101),
    and including the PreAct variant for each depth~\citep{he2016deep,he2016identity}.
    \item \textbf{ResNeXT} models with various widths and depths (2x64d, 32x4d, 4x64d)~\citep{xie2017aggregated}.
    \item \textbf{Squeeze-and-Excitation Networks} with 18 layers~\citep{hu2018squeeze}.
    \item \textbf{ShuffleNet}, specifically the G2, G3, and V2 variants, with network scale factors 0.5, 1, 1.5, and 4 for the ShuffleNetV2 architecture~\citep{zhang2018shufflenet, ma2018shufflenet}.
    \item \textbf{VGG} with 11, 13, 16, and 19 layers~\citep{simonyan2014very}.
\end{itemize}

\paragraph{Convolutional neural networks pre-trained on ImageNet.}
We explored the use of models pre-trained on ImageNet both in the \cifarten-universe and in the WILDS datasets.
In some experiments, we also trained the following architectures from scratch to quantify the effect of pre-training in detail (see Section \ref{sec:training_data}).
The code for the following models is from \url{https://github.com/creafz/pytorch-cnn-finetune}.
\begin{itemize}
    \item \textbf{AlexNet}~\citep{alexnet}.
    \item \textbf{DenseNet} with 121, 161, 169, and 201 layers~\citep{huang2017densely}
    \item \textbf{Dual Path Networks (DPN)}, variants 68, 68b, and 92~\citep{chen2017dual}.
    \item \textbf{GoogLeNet}, a member of the Inception family~\citep{szegedy2015going}.
    \item \textbf{MobileNetV2}~\citep{sandler2018mobilenetv2}.
    \item \textbf{Neural Architecture Search Networks (NASNets)}, specifically
    NASNet-A-Large and PNASNet-5-Large~\citep{zoph2017learning, liu2018progressive}
    \item \textbf{ResNet} with 18, 34, 50, 101, and 152 layers~\citep{he2016deep, he2016identity}.
    \item \textbf{ResNeXT}, configurations 50\_32x4d and 101\_32x4d~\citep{xie2017aggregated}.
    \item \textbf{Squeeze-and-Excitation Networks}, specifically se\_resnext50\_32x4d and se\_resnext101\_32x4d~\citep{hu2018squeeze}.
    \item \textbf{ShuffleNetV2}, scale factors 0.5 and 1~\citep{zhang2018shufflenet,ma2018shufflenet}.
    \item \textbf{SqueezeNet}, version 1.0 and 1.1~\citep{i2016squeezenet}.
    \item \textbf{VGG} with 11, 13, and 16 layers, including variants with batch normalization for 13 and 16 layers~\citep{simonyan2014very}.
\end{itemize}

\paragraph{Convolutional neural networks only trained on ImageNet.}
For the zero-shot generalization experiments in Section \ref{sec:training_data}, we also utilized a set of models trained on \imagenet without any further fine-tuning to \cifarten.
As above, the models are from \url{https://github.com/creafz/pytorch-cnn-finetune}.
\begin{itemize}
    \item \textbf{AlexNet}~\citep{alexnet}
    \item \textbf{DenseNet} with 121, 161, 169, and 201 layers~\citep{huang2017densely}.
    \item \textbf{Dual Path Networks (DPN)}, variants 68, 68b, 92, 98, 107, and 131~\citep{chen2017dual}.
    \item \textbf{Inception} models: GoogleNet, InceptionV3, and InceptionResNetV2~\citep{szegedy2015going, szegedy2015rethinking,
    szegedy2016inceptionv4}.
    \item \textbf{MobileNetV2}~\citep{sandler2018mobilenetv2}.
    \item \textbf{PolyNet}~\citep{zhang2016polynet}.
    \item \textbf{ResNet} with 18, 34, 50, 101, and 152 layers~\citep{he2016deep,he2016identity}.
    \item \textbf{Squeeze-and-Excitation Networks} specifically senet154, se\_resnet50, se\_resnet101, se\_resnet152, se\_resnext50\_32x4d, and se\_resnext101\_32x4d~\citep{hu2018squeeze}.
    \item \textbf{ShuffleNetV2}, scale factors 0.5 and 1~\citep{zhang2018shufflenet,ma2018shufflenet}.
    \item \textbf{SqueezeNet}, version 1.0 and 1.1~\citep{i2016squeezenet}.
    \item \textbf{ResNeXT}, configurations 50\_32x4d, 101\_32x4d, 101\_32x8d, and 101\_64x4d~\citep{xie2017aggregated}.
    \item \textbf{VGG} with 11, 13, 16, and 19 layers, all with and without batch normalization~\citep{simonyan2014very}.
    \item \textbf{Xception}~\citep{chollet2017xception}.
\end{itemize}

\paragraph{Further models trained on extra data.}
To measure the effect of extra training data more broadly than only relying on ImageNet for pre-training, we also included the following three models utilizing data from different sources:
\begin{itemize}
    \item \textbf{CLIP}: We evaluate the two publicly released CLIP models \cite{radford2021learning}.
    These models were trained with 400 million image-caption pairs scraped from the web.
    We evaluate the two ResNet50 and VisionTransformer variants released by the CLIP team.
    CLIP models are particularly interesting since they can be evaluated zero-shot:
    image classification labels can be turned into textual prompts so that the model
    can be evaluated on downstream tasks without needing to look at the training data.
    \item \textbf{Self-training on 80 Million Tiny Images:} \citet{carmon2019unlabeled} introduced robust self-training (RST) and showed that unlabeled data can improve adversarial robustness.
    In the context of their work, they also trained baseline CIFAR-10 models that used data from 80 Million Tiny Images~\citep{tinyimages} in addition to the standard CIFAR-10 training set.
    This baseline model is an interesting addition to our testbed since the extra training data from a potentially more diverse source may move the model away from the linear trend given by models only trained on CIFAR-10.
    \item \textbf{Out-distribution aware self-training (ODST):} \citet{augustin2020out} develop an iterative self-training approach to leverage unlabeled data when some of the unlabeled data is not relevant to the classification task of interest.
    They also instantiate their approach on CIFAR-10, using 80 Million Tiny Images as an unlabeled data source.
    As before, the ODST models are relevant for our experiments because they use extra training data beyond the standard CIFAR-10 training set.
\end{itemize}

\paragraph{Classical methods.}
In addition to the neural network methods discussed previously, we also
integrated several classical, non-neural network methods into our testbed.
Unless noted otherwise, we used the implementations from
scikit-learn~\citep{scikitlearn}. Each of these methods works directly on the image
pixels, which are each scaled to have zero-mean and unit variance on the
training set. We included the following methods into our testbed:
\begin{itemize}
    \item Random features~\citep{coates2012learning}, using the implementation from \url{https://github.com/modestyachts/nondeep}.
    \item AdaBoost from~\citet{hastie2009multi}, using an scikit-learn decision tree classifier to build the boosted ensemble.
    \item Ridge regression classifiers with varying $\ell_2$ regularization parameter
    \item Support vector machines with linear, gaussian, and polynomial kernels and varying regularization penalty term.
    \item Logistic regression with varying regularization parameter and using both $\ell_1$ and $\ell_2$ regularization.
    \item Quadratic discriminant analysis.
    \item Random forests~\citep{breiman2001random} with varying maximum tree
    depth, number of trees in the forest, and using both entropy and gini
    impurity as the splitting criterion.
    \item Nearest neighbor with varying number of $k$ nearest-neighbors and
    using $\ell_2$ distance between points.
\end{itemize}

\subsubsection{Summary statistics}
The following two tables give a brief overview of the number of experiments we ran with our testbed.
Table \ref{tab:model_counts} shows how many distinct models we trained for each of our training sets (a total of about 3,000).
Each of these models was then evaluated on a range of test sets to generate the scatter plots in this paper.

Table \ref{tab:eval_counts} shows the total number of evaluations for each family of datasets.
Besides being tested on multiple datasets, one trained model can also have led to several evaluations since we sometimes evaluated all training checkpoints of a model on multiple datasets as well to study whether the linear trends are reliable when varying training duration (see Section \ref{sec:linear_trends-results}).
This lead to an overall total of about 100,000 model evaluations, each of which
corresponds to one point in a scatter plot in our paper.

\begin{table}
    \centering
    \rowcolors{2}{gray!15}{white}
\renewcommand{\arraystretch}{1.3}
\begin{tabular}{l r }
\toprule
Dataset & Number of trained models \\[.1cm]
\midrule
  \cifarten & \numprint{1,895} \\
  \iwildcam & \numprint{197} \\
  \fmow & \numprint{592} \\
  \camelyon & \numprint{461} \\
\bottomrule
\end{tabular}
\caption{Number of trained models (of all types) by training set. The model counts include only fully trained models, not intermediate checkpoints.}

    \label{tab:model_counts}
\end{table}

\begin{table}
    \centering
    \rowcolors{2}{gray!15}{white}
\renewcommand{\arraystretch}{1.3}
\begin{tabular}{l r }
\toprule
Dataset & Number of model evaluations \\[.1cm]
\midrule
  \cifarten & \numprint{6,814} \\
  \cifartenone & \numprint{5,315} \\
  \cifartentwo & \numprint{11,212} \\
  \cifartenc & \numprint{39,677} \\
  \cinicten & \numprint{4,259} \\
  \stlten & \numprint{507} \\
  \iwildcam & \numprint{15,147} \\
  \fmow & \numprint{12,127} \\
  \camelyon & \numprint{7,056} \\
\bottomrule
\end{tabular}
\caption{Number of model evaluations by test set type. Some of the rows, e.g., \cifartenc, correspond to multiple individual test sets. We count evaluations of a model and its training checkpoints separately here.}
    \label{tab:eval_counts}
\end{table}

\subsection{\imagenet Testbed}
\label{app:imagenettestbed}

\subsubsection{Datasets}
We include all of the natural distribution shifts from the testbed of \citet{taori2020measuring},
excluding the consistency shifts since those are somewhat adversarial in nature.
\paragraph{\imagenettwo.}
\imagenettwo is a reproduction of the \imagenet test set collected by \citet{recht2019imagenet}.
\paragraph{ObjectNet.}
ObjectNet is a test set of objects in a variety of scenes, poses, and lighting conditions
with 113 classes that overlap with \imagenet \cite{barbu2019objectnet}.
\paragraph{ImageNet-Vid-Anchors and YTBB-Anchors.}
These are two datasets introduced by \citet{shankar2019image} that measure accuracy on frames taken from video clips.
They contain  30 and 24 super-classes of the \imagenet class hierarchy, respectively.
For evaluation, we measure accuracy using the pm-0 metric as defined in \citet{shankar2019image},
which measures accuracy over the anchor frames of the video clips.
\paragraph{ImageNet-A.}
ImageNet-A \cite{hendrycks2021natural} is an adversarially collected dataset, constructed by downloading labeled images from a variety of online sources
and then selecting the subset that was misclassified by a ResNet-50 model.

\subsubsection{\imagenet Testbed Models}
We include all of the existing models in the testbed from \citet{taori2020measuring}, and add a few others:
\begin{enumerate}
\item \textbf{CLIP:} We add the two CLIP models released by \citet{radford2021learning} and evaluate them
zero-shot using the publicly released textual prompts.
\item \textbf{Self-supervised models:} We add models trained using a few different self-supervised methods:
SimSiam \cite{chen2020exploring}, SimCLRv2 \cite{chen2020big}, and SwAV \cite{caron2021unsupervised}.
For SimSiam and SwAV, we use the ResNet50 variants pretrained on \imagenet without labels and then
final-layer finetuned on \imagenet. For SimCLRv2, we use a ResNet50 and a ResNet152 variant, and for each
use a model final-layer finetuned and whole-network finetuned on \imagenet.
\item \textbf{Classical models:} We add four low-accuracy classical models: random features \cite{coates2012learning},
random forests \cite{breiman2001random}, one-nearest neighbors, and a linear model trained with least squares.
The random forests model, the linear model, and the one-nearest neighbors model were trained directly on pixels of images downsampled to 32x32.
\item \textbf{Low accuracy ConvNets:} We add a multitude of low-accuracy ResNet models trained for various epochs and
on various subsets of the training set.
\end{enumerate}

\subsection{\ycb Testbed}
\label{app:ycbtestbed}
We describe the 6D pose estimation task, our synthetic dataset, and the models
in our testbed below.

\subsubsection{6D Pose Estimation}
In 6D pose estimation, the task is to determine the three-dimensional
position and orientation of an object in a scene. Concretely, for our purposes,
models are given as input a single $128 \times 128$ RGB image of an object and
must determine the object's 6 degree-of-freedom pose (rotation and translation)
relative to the scene. For more background on pose estimation, see~\citet{lepetit2005monocular}
or~\citet{xiang2017posecnn} and the references therein.

We evaluate each model using the accuracy metric
from~\citet{hinterstoisser2012model}. Specifically, given a ground truth
rotation $R$ and translation $t$, estimated rotation $\tilde{R}$ and translation
$\tilde{t}$, and a 3D model $\mathcal{M}$ consisting of $m$ points $x \in
\mathcal{M}$, then average distance (ADD) metric
of~\citet{hinterstoisser2012model} is the mean of the distances between 3D model
points transformed under the ground-truth and estimated poses
\begin{align*}
    \mathrm{ADD} = \frac{1}{m} \sum_{x \in \mathcal{M}} \|(Rx + t) - (\tilde{R}x + \tilde{t})\|_2.
\end{align*}
An estimated 6D pose is consider to be \emph{correct} if the ADD is less than
10\% of the diameter of the 3D model $\mathcal{M}$.

\subsubsection{\ycb Datasets.}
Similar to~\citet{xiang2017posecnn} and~\citet{tremblay2018deep},
we construct a synthetic datasets for 6D pose estimation by rendering images of
known object models from~\citet{calli2015benchmarking}
and~\citet{hashimotokosnet} using Blender~\citep{blender}. We use the subset of
16 non-symmetric YCB objects from~\citet{xiang2017posecnn}, as well as the two
non-symmetric objects from~\citet{hashimotokosnet} in our experiments.

In our datasets, each object is placed on a plane with one of 60 textures from
\url{texturehaven.com} and rendered with lighting from one of 60 HDRIs from
\url{hdrihaven.com}. To generate distribution shift, we separate the textures
into two, non-overlapping subsets based on their material properties. The
in-distribution test set uses one subset of textures, and the
out-of-distribution test set uses the other. See
Figure~\ref{fig:ycb_example_images} for example images corresponding to the in
and out-of-distribution textures and corresponding datasets.

We generate datasets by uniformly sampling an object, a background lighting
environment, a background texture (from the in or out-of-distribution subset),
an object pose, and a camera pose. We generate in-distribution training sets of
50,000 and 100,000 images and both in and out-of-distribution test sets of
10,000 images. In this section, we use the 50,000 example training set for our
experiments. We use the 100,000 example training set to explore the effect of
adding more i.i.d. training set in Appendix~\ref{app:linearfit_variations}.

In simulation, the object model, the object pose, and the camera pose are all
known in advance, so we can easily compute a ground truth pose for each object
relative to the scene.  We additionally annotate each image in our dataset with
9 2D keypoints corresponding to projection of the 3D bounding box and the 3D
center of the object onto the 2D image. Figure~\ref{fig:ycb_example_images}
visualizes these annotations for a random sample of images from the training
set.

\subsubsection{\ycb Models}
The neural pose estimation models in our testbed are all based on semantic
segmentation networks for predicting 2D keypoints. In essence, each network
takes as input the entire image of the scene and predicts the nine keypoints
previously described and shown in Figure~\ref{fig:ycb_example_images}. Some
implementations in the literature first estimate bounding boxes of the objects
in the scene before passing the images to the keypoint prediction network.
Since our scenes only contain a single object, the networks in our testbed do
not perform this step.

Given 2D keypoints predictions, each model then uses the PnP
algorithm~\citep{lepetit2009epnp} to recover the 3D object pose. This approach,
developed by~\citet{rad2017bb8} and~\citet{pavlakos20176} is also used in
high-performing implementations like~\citet{tremblay2018deep}.  Our testbed
contains several models for the semantic segmentation backbone.  Unless
otherwise noted, the implementation is taken
from~\url{https://github.com/qubvel/segmentation_models.pytorch}.
\begin{enumerate}
\item UNet~\citep{ronneberger2015u} with ResNet~\citep{he2016deep},
MobileNet~\citep{sandler2018mobilenetv2}, and EfficientNet-b7~\citep{tan2019efficientnet} as the encoder.
\item UNet{++}~\citep{zhou2018unet++}.
\item FCN\_ResNet with varing depths 18, 34, 50, and
101~\citep{long2015fully}, using the implementation from
\url{https://github.com/pytorch/vision/tree/master/torchvision/models/segmentation}.
\item LinkNet~\citep{chaurasia2017linknet}.
\item PSPNet~\citep{zhao2017pyramid}.
\item PoseNet~\citep{kendall2015posenet}.
\item 2-layer CNN
\end{enumerate}
Each of these models outputs a set of nine heatmaps, one corresponding to each
keypoint prediction. For each model, we use the PnP implementation
from~\citet{opencv_library}.

\section{The linear trend phenomenon}
\label{app:more_linear_trends}
In this section, we present additional examples of linear trends between
in-distribution and out-of-distribution performance across each of the testbeds
discussed in Appendix~\ref{app:testbed}. In Appendix~\ref{app:linear_trends_examples},
we first highlight examples of linear trends across a variety of distribution shifts
for models in each of the \cifarten, \fmow, \imagenet, and the \ycb
``universes'' discussed previously. Then, in
Appendix~\ref{app:linearfit_variations}, we show these linear trends are
invariant to changes in model hyperparameters, training duration, and training
set size.

\subsection{Further examples of linear trends}
\label{app:linear_trends_examples}

\subsubsection{\cifarten}
\label{app:linearcifarten}
\paragraph{Dataset reproduction shifts.}
In Figure~\ref{fig:cifar10_benchmark_fits}, we plot out-of-distribution test
accuracy vs. in-distribution \cifarten test accuracy for each of the \cifarten
testbed models described in Appendix~\ref{app:testbed_models} on two different
dataset reproduction shifts: \cifartenone, \cifartentwo. For each
shift, the relationship between in-distribution and out-of-distribution test
accuracy for both classical and neural models is well captured by a linear fit,
and the corresponding $R^2$ statistic is greater than $0.99$ for each example.

\paragraph{Distribution shifts between machine learning benchmarks.}
In Figure~\ref{fig:cifar10_benchmark_fits}, we also plot out-of-distribution test
accuracy vs. in-distribution \cifarten test accuracy for each of the \cifarten
testbed models described in Appendix~\ref{app:testbed_models} on two different
machine learning benchmark shifts: \cinicten, and \stlten. 
The class structure of \stlten differs slightly from \cifarten and
includes a monkey class instead of a frog class. For the \stlten experiment we
therefore consider nine-class variants of \stlten and \cifarten, omitting
instances with monkey or frog labels, and, for each model, we mask the frog
class (or logit) and predict only among the remaining nine classes. The
relationship between ID and OOD accuracy is well-captured by a linear fit and
the $R^2$ statistic is greater than $0.99$ in each case.

\paragraph{Synthetic perturbations.}
In Figure~\ref{fig:cifar10_cifar10c_fits}, we plot out-of-distribution test
accuracy vs. in-distribution \cifarten test accuracy for the same collection of
\cifarten testbed models on a subset of eight different synthetic dataset shifts
from \cifartenc~\citep{hendrycks2018benchmarking} where very clean linear trends
occur--- fog, brightness, snow, defocus blur, spatter, elastic transform, frost,
and saturate. For each shift, the linear fit well-approximates the relationship
between in-distribution and out-of-distribution accuracy, and the $R^2$
statistic is greater than $0.94$ for each example. However, the fits are not as
clean as the machine learning benchmark shifts discussed previously, and,
moreover, for several of the synthetic perturbations in \cifartenc, there is no
linear trend at all. We discuss examples from \cifartenc where linear trends
fail to hold further in Section~\ref{sec:fit_failure} and
Appendix~\ref{app:fit_failure}.

\begin{figure*}[ht!]
    \centering
    \includegraphics[width=\linewidth]{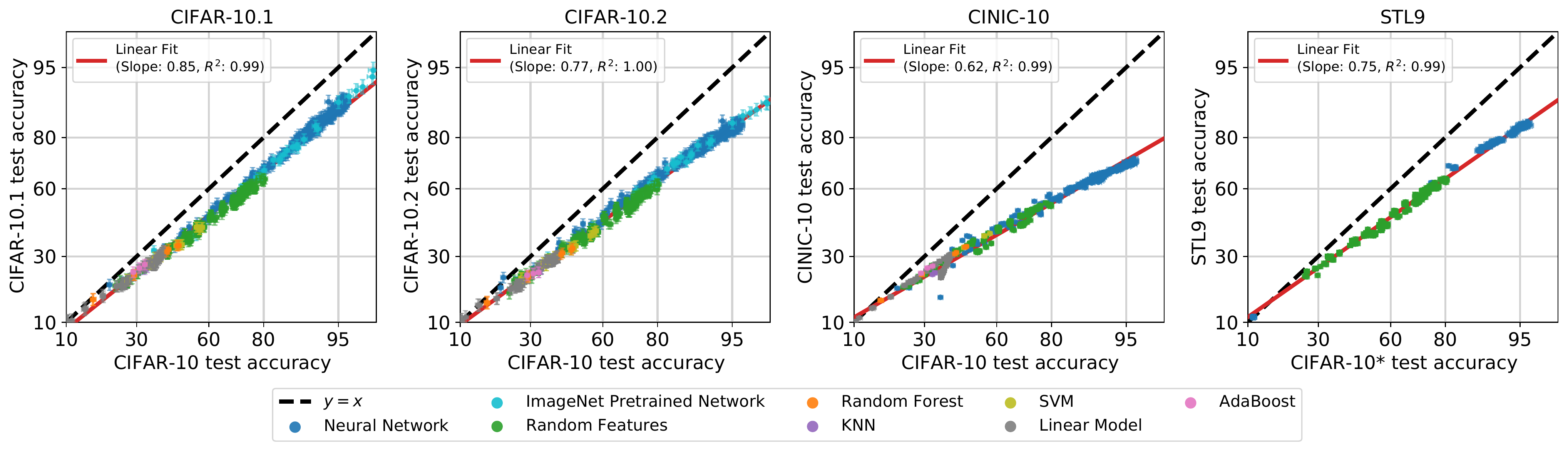}
    \caption{
        Out-of-distribution accuracies vs. in-distribution \cifarten test
        accuracies for a wide range of models across two different dataset
        reproduction shifts, \cifartenone and \cifartentwo, as well as two
        different shifts between machine learning benchmarks, 
        \cinicten, and \stlten.
        Each point corresponds to a model evaluation, shown with 95\%
        Clopper-Pearson confidence intervals (mostly covered by the point
        markers). For the \stlten experiment, we consider nine-class subsets of
        both \stlten and \cifarten, omitting the monkey and frog class,
        respectively, and restrict each model to predict only from the remaining
        nine-classes.
    }
    \label{fig:cifar10_benchmark_fits}
\end{figure*}
\begin{figure*}[ht!]
    \centering
    \includegraphics[width=\linewidth]{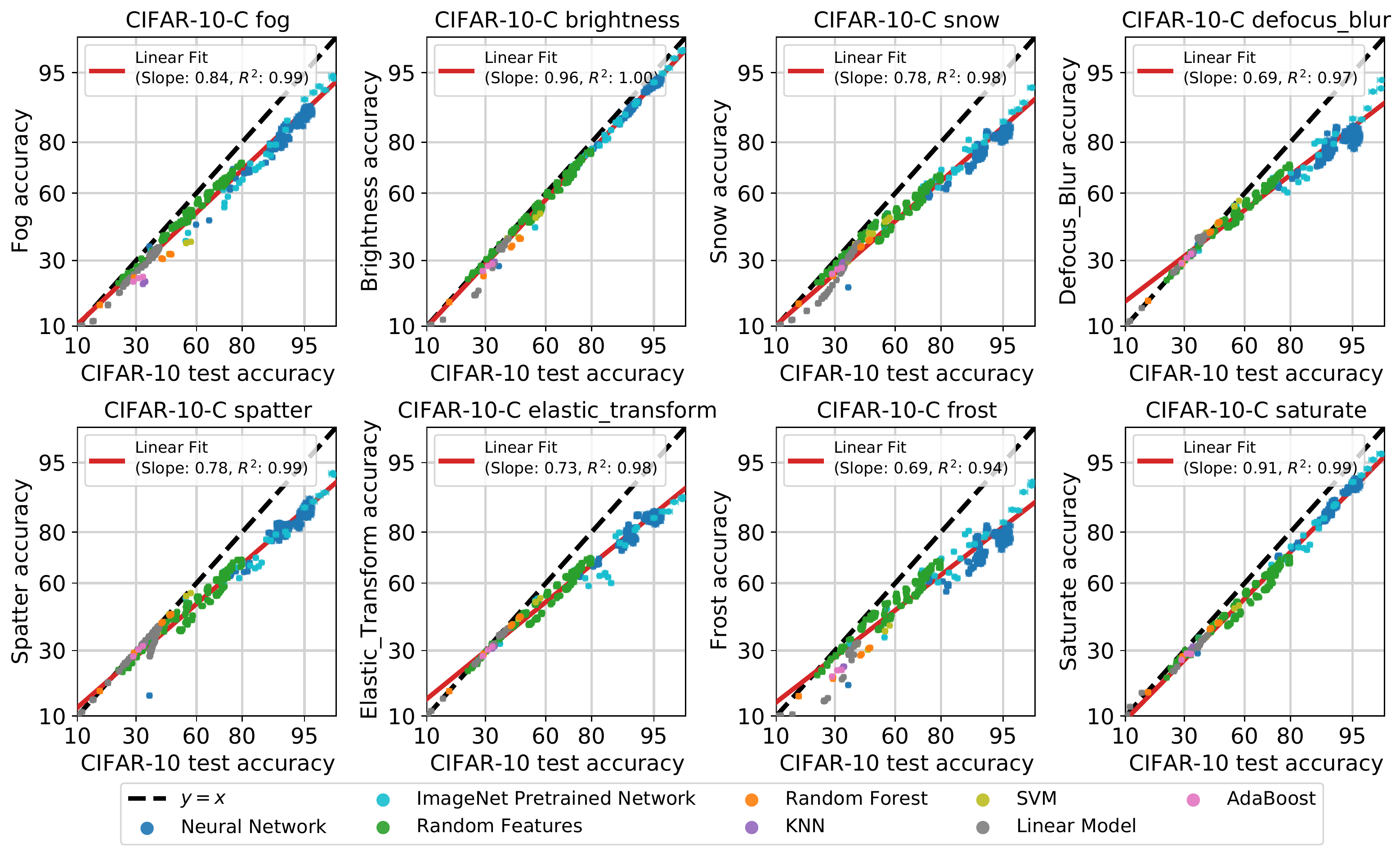}
    \caption{
        Out-of-distribution accuracies vs. in-distribution \cifarten test
        accuracies for a wide range of models from our \cifarten testbed
        across eight \emph{synthetic perturbation} shifts from \cifartenc.
        Each point corresponds to a model evaluation, shown with 95\%
        Clopper-Pearson confidence intervals (mostly covered by the point
        markers).
    }
    \label{fig:cifar10_cifar10c_fits}
\end{figure*}

\subsubsection{\fmow}
\label{app:linearfmow}
In Figure~\ref{fig:fmow_linear_fits}, we plot out-of-distribution test accuracy
vs. in-distribution \fmow test accuracy for both the classical methods and the
\imagenet networks from the main testbed described in
Appendix~\ref{app:testbed}. We evaluate each model on both the
out-of-distribution validation and the out-of-distribution test set from \fmow
using two metrics: average accuracy and worst accuracy over five geographical
regions (for more details on \fmow, see Appendix~\ref{app:testbed}). To remove
noise from very low accuracy models, we restrict our attention to models with
\fmow test set accuracy at least 10\%. Across both out-of-distribution datasets
and both metrics, the linear fit well-captures the relationship between in and
out-of-distribution performance with an $R^2$ statistic of a least $0.98$.

\paragraph{Experimental details.}
Below, we provide additional technical details about our \fmow experiments.
\begin{itemize}
    \item \textbf{Datasets.} We train each model on the training split of the
    \fmow dataset~\cite{christie2018functional} defined by~\citet{koh2020wilds},
    and perform testing on the in-distribution (ID) and out-of-distribution
    (OOD) validation and test splits defined by~\citet{koh2020wilds}.
    \item \textbf{Worst-region accuracy confidence intervals.}
    We heuristically obtain confidence intervals for the worst-region accuracy
    by computing standard 95\% Clopper-Pearson confidence intervals for accuracy
    in the region with the lowest accuracy on the test set for each model.
    \item \textbf{Training hyperparameters.} Unless otherwise noted, we train
    all of the neural models using learning rate $10^{-4}$ and weight decay $0$
    for 50 epochs. We use Adam throughout with all other parameters set to
    their default PyTorch values.
\end{itemize}

\begin{figure*}[ht!]
    \centering
    \includegraphics[width=\linewidth]{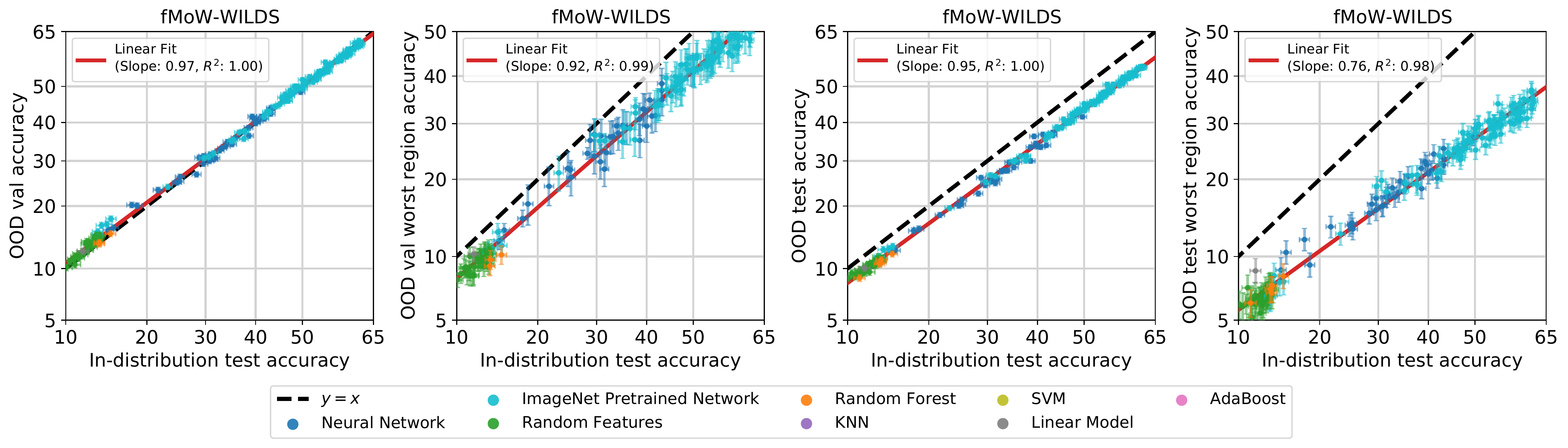}
    \caption{
        Out-of-distribution accuracies vs. in-distribution \fmow test
        accuracies for a wide range of classical methods and \imagenet networks
        from our main testbed. Each point corresponds to a model evaluation,
        shown with 95\% Clopper-Pearson confidence intervals (mostly covered by
        the point markers). \textbf{Left}: In the left two plots, we evaluate
        each model on the \fmow OOD \emph{validation set} using both average and
        worst-region accuracy. \textbf{Right}: In the right two plots, we
        evaluate each model on the \fmow OOD \emph{test set} using both average and
        worst-region accuracy. In all four cases, a linear fit well captures the
        relationship between in-distribution and out-of-distribution performance
        with $R^2$ statistics greater than $0.98$ in each setting.
    }
    \label{fig:fmow_linear_fits}
\end{figure*}
\FloatBarrier

\subsubsection{\imagenet}
\label{app:linearimagenet}

\begin{figure*}[ht!]
    \centering
    \includegraphics[width=\linewidth]{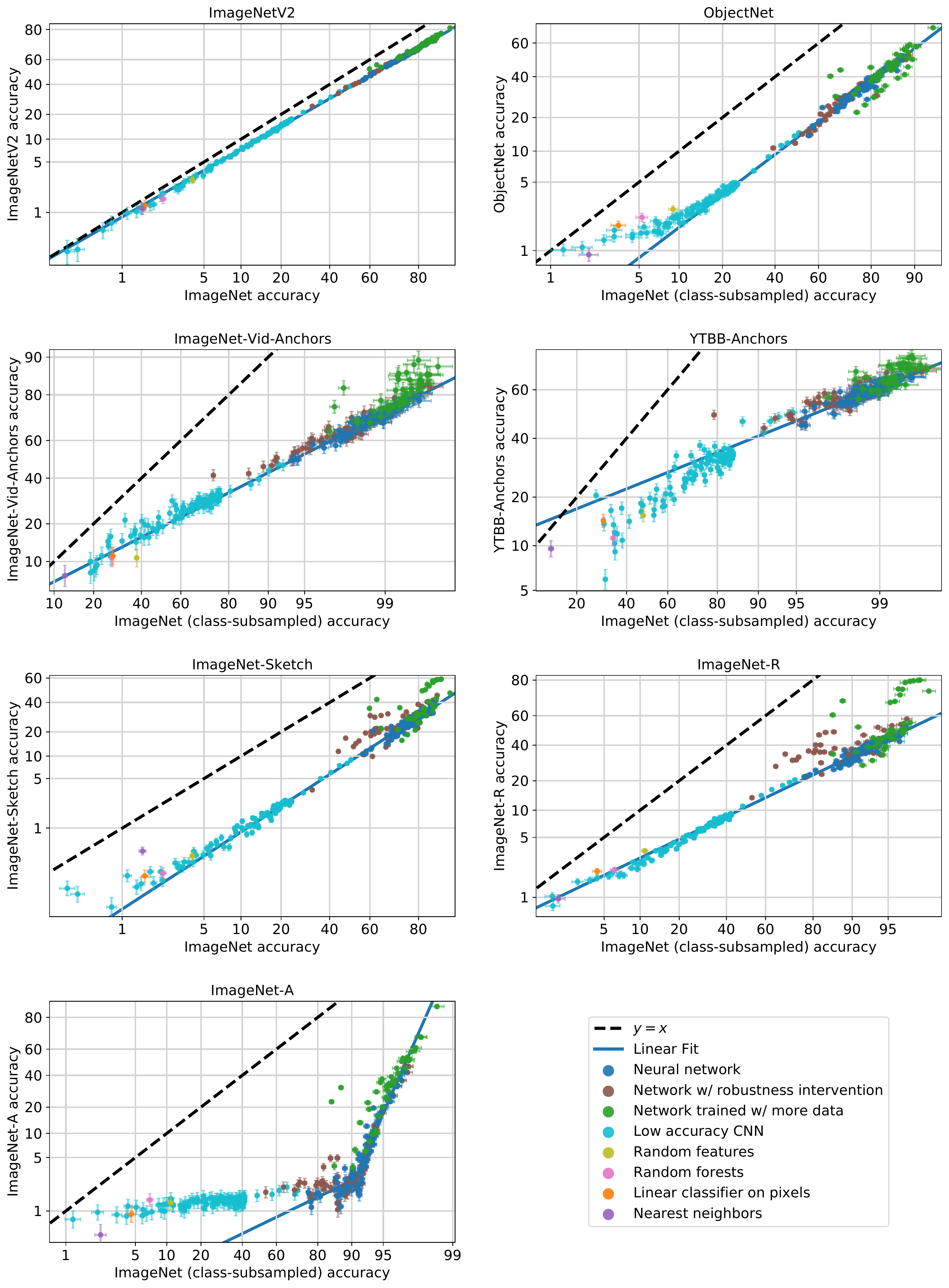}
    \vspace{-0.2cm}
    \caption{
        Model accuracies on the ImageNet natural distribution shifts \cite{taori2020measuring}:
        ImageNetV2, ObjectNet, ImageNet-Vid-Anchors, YTBB-Anchors, ImageNet-Sketch, ImageNet-R, and ImageNet-A.
        Each point corresponds to a model evaluation,
        shown with 95\% Clopper-Pearson confidence intervals.
        The axes are scaled via logit scaling, and the linear fit is fit 
        only to standard networks following \cite{taori2020measuring}.
    }
    \label{fig:imagenet_all_shifts}
\end{figure*}

In Figure \ref{fig:imagenet_all_shifts}, we plot the existing models from
\citet{taori2020measuring} alongside the new models in our testbed (CLIP,
self-supervised models, and classical methods like random features, random
forests, nearest neighbors, and linear regression)
on the ImageNet natural distribution shifts.  
First, we observe that the two CLIP models are significantly robust on all shifts
(these models are the two green points above the line at around 60\% \imagenet{}
accuracy).  
This is interesting and is in line with our conclusions that
pretraining on extra data can increase model robustness to distribution shift.
Second, we observe that all three low-accuracy models lie relatively near the
predicted fit line for ImageNetV2, ImageNet-Vid-Anchors, ImageNet-Sketch, and ImageNet-R.
Note that this line is fit only to the standard neural networks (blue points).
Understanding why the fit isn't as predictive in the low-accuracy regime for 
ObjectNet, YTBB-Anchors, and ImageNet-A is an interesting direction for future work.
Note that the fit to ImageNet-A is performed piecewise around the ResNet-50 model accuracy
following the procedure in \citet{taori2020measuring}.
\FloatBarrier

\subsubsection{\iwildcam}
\label{app:iwcv2}

\paragraph{Experiment details.}
The models reported in the \iwildcam panel of Figure~\ref{fig:main_figure} were  obtained using the following parameters. We trained 10 neural network architectures on the \iwildcam training set (see legend of Figure~\ref{fig:iwc-arch} below). For each architecture, we perform both training from scratch and fine-tuning from a model pretrained on ImageNet. The fine-tuning configurations are similar to the setting of \citet{koh2020wilds}: we train for 12 epochs with batch size 16 using Adam and sweep over  learning rate and weight decay values in the grid $\{ 3\cdot 10^{-5}, 10^{-4}, 10^{-3}, 10^{-2}\}\times \{ 0, 10^{-3}, 10^{-2}\}$. The other Adam parameters were set to the Pytorch defaults. For models trained from scratch we use the same hyperparameter grid except we train for 15 epochs, which seems to suffice for convergence for each model with at least some of the learning rates. For details on error bar calculation and label noise reduction, see Appendix~\ref{app:datasets-iwc}. 

\paragraph{Architecture variation with fixed weight decay.}
Figure~\ref{fig:iwc-arch} provides a more detailed view of the \iwildcam experiments, wherein we plot the final epoch performance of the models we train, where the weight decay is set to zero. As the figure shows, the error bars for all model intersect the fitted linear trend line (in probit domain), with the exception of AlexNet when training from scratch, which is slightly below the linear trend. Varying the weight decay parameter appears to affect the ID/OOD trend of fine-tuned model; see Section~\ref{sec:training_data} and Appendix~\ref{app:more-data-iwc} for additional discussion and plots.

\begin{figure*}
	\centering
	\includegraphics[width=0.9\linewidth]{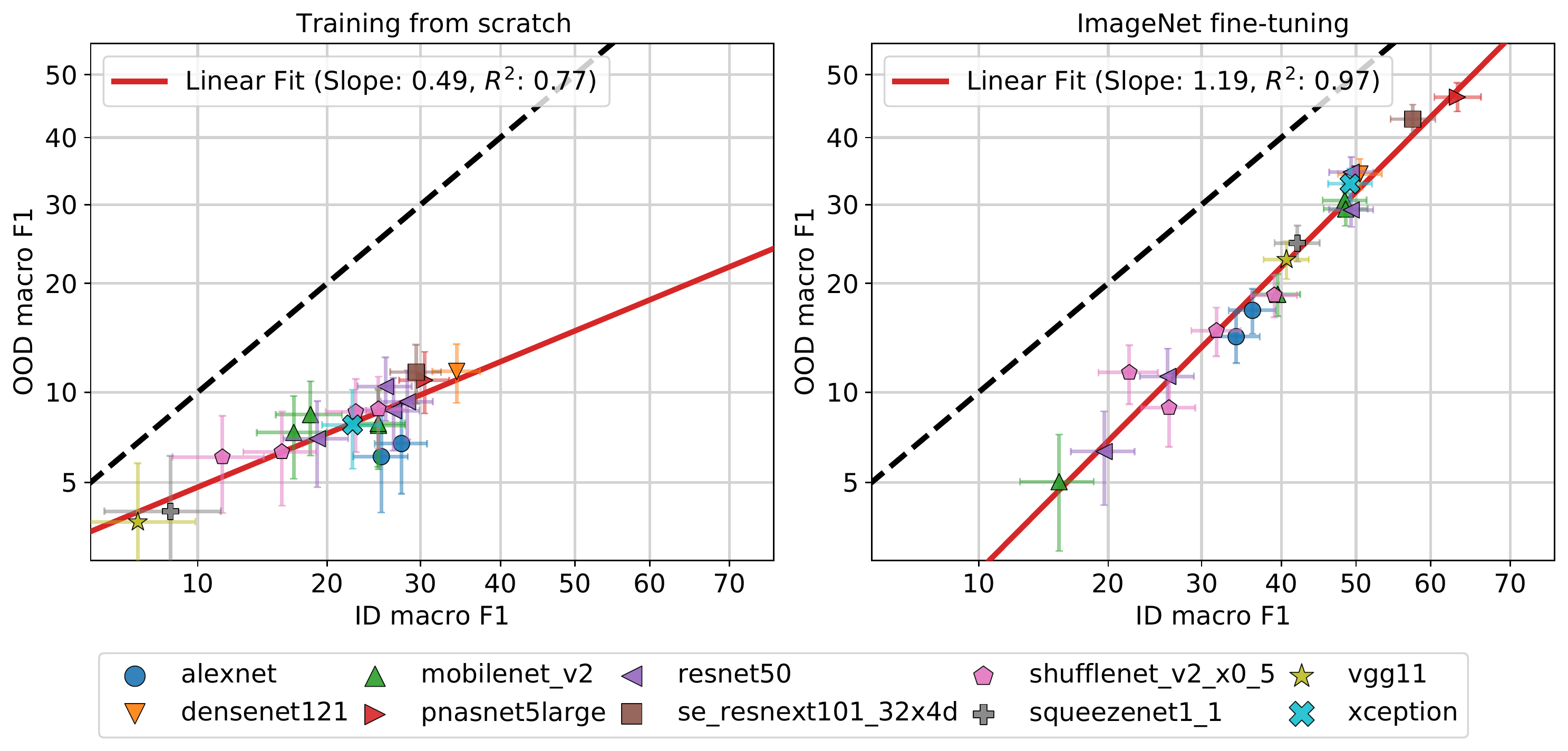}
	\caption{
		OOD vs.\ ID macro F1 scores for \iwildcam models trained from scratch
        (left) or fine-tuned from pretrained ImageNet models (right), with
        varying model architecture and learning rate, but weight decay fixed to
        zero. Contrast with Figure~\ref{fig:iwc-app-wd} for results when varying the weight decay parameter.
	}
    \label{fig:iwc-arch}
\end{figure*}
\FloatBarrier

\subsubsection{\ycb}
\label{app:linearycb}
In Figure~\ref{fig:ycb_appendix_fits}, we plot out-of-distribution accuracy
versus in-distribution accuracy for a synthetic 6D pose estimation task using
the YCB object models from~\citet{calli2015benchmarking} and a testbed of neural
models for 6D pose estimation. As in the previous examples, a linear fit
well-approximates the relationship between in and out-of-distribution accuracy
with an $R^2$ statistic of $0.99$.

\begin{figure*}[ht!]
    \centering
    \includegraphics[width=0.5\linewidth]{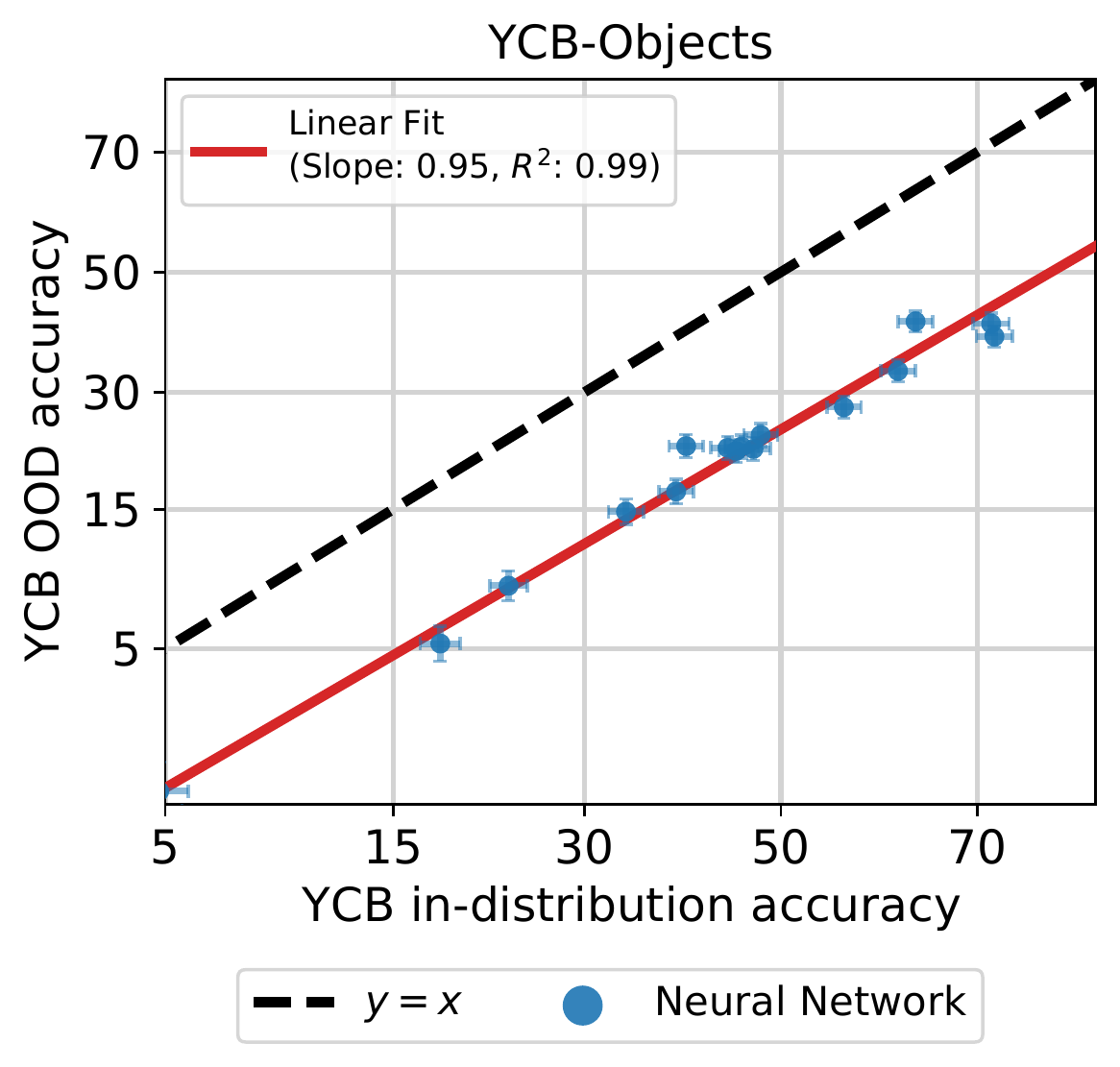}
    \caption{
        Out-of-distribution accuracy vs.\ in-distribution accuracy for a
        synthetic 6D pose estimation task based on the YCB object models
        from~\citet{calli2015benchmarking} across a testbed of neural pose
        estimation networks. Each point corresponds to a model evaluation shown
        with 95\% Clopper-Pearson confidence intervals. The distribution shift
        corresponds to varying the background texture for rendered images of the
        YCB objects. See Figure~\ref{fig:ycb_example_images} for example images
        both in and out-of-distribution. Appendix~\ref{app:linearycb} describes the
        dataset and the model testbed in more detail.
    }
    \label{fig:ycb_appendix_fits}
\end{figure*}

\begin{figure*}[ht!]
    \centering
    \begin{subfigure}{0.5\textwidth}
        \includegraphics[width=\linewidth]{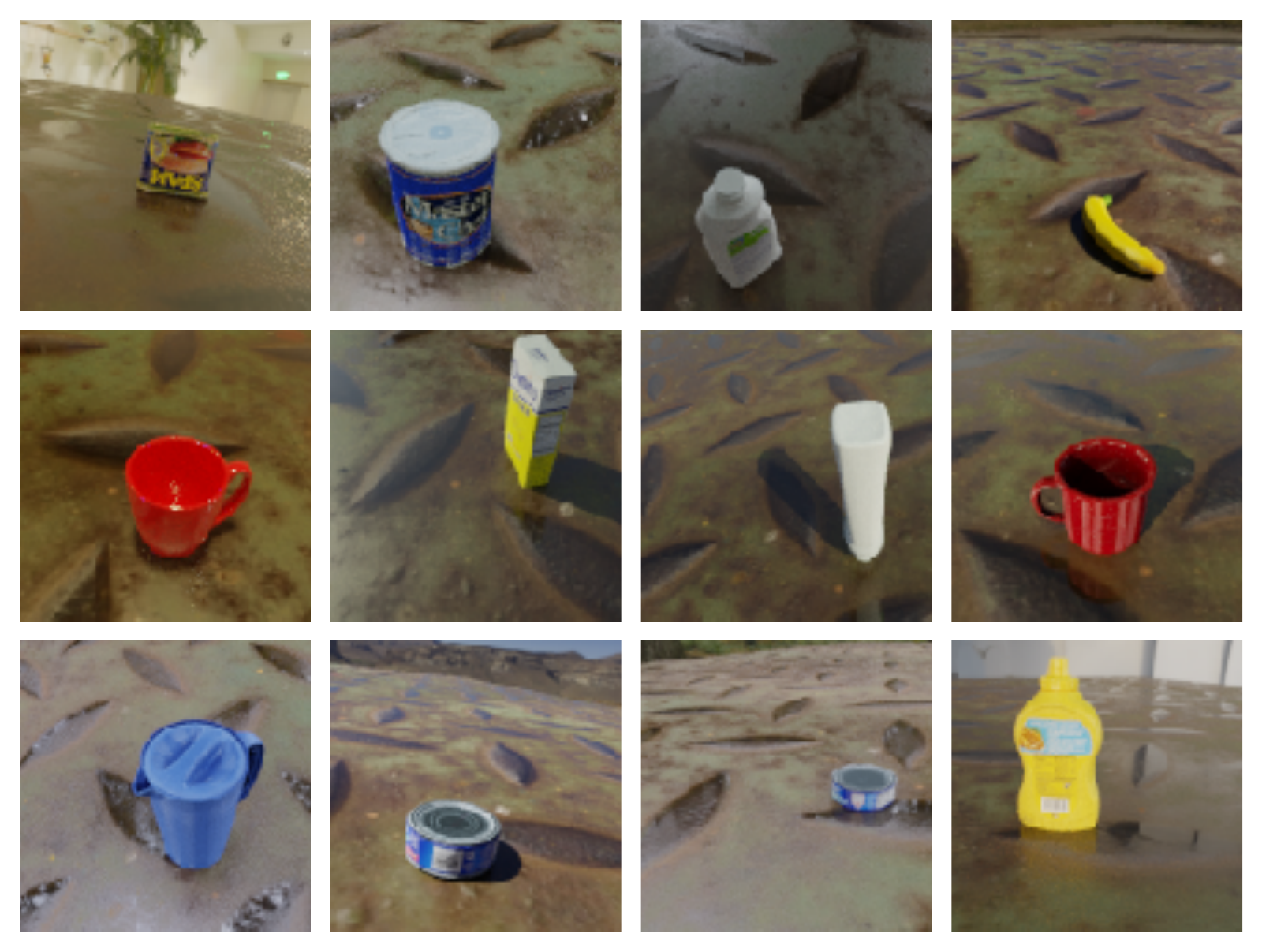}
        \caption{Example images from the \ycb in-distribution test set. Each
        object is rendered on a background whose texture has similar material
        properties.}
    \end{subfigure} \\
    \begin{subfigure}{0.5\textwidth}
        \includegraphics[width=\linewidth]{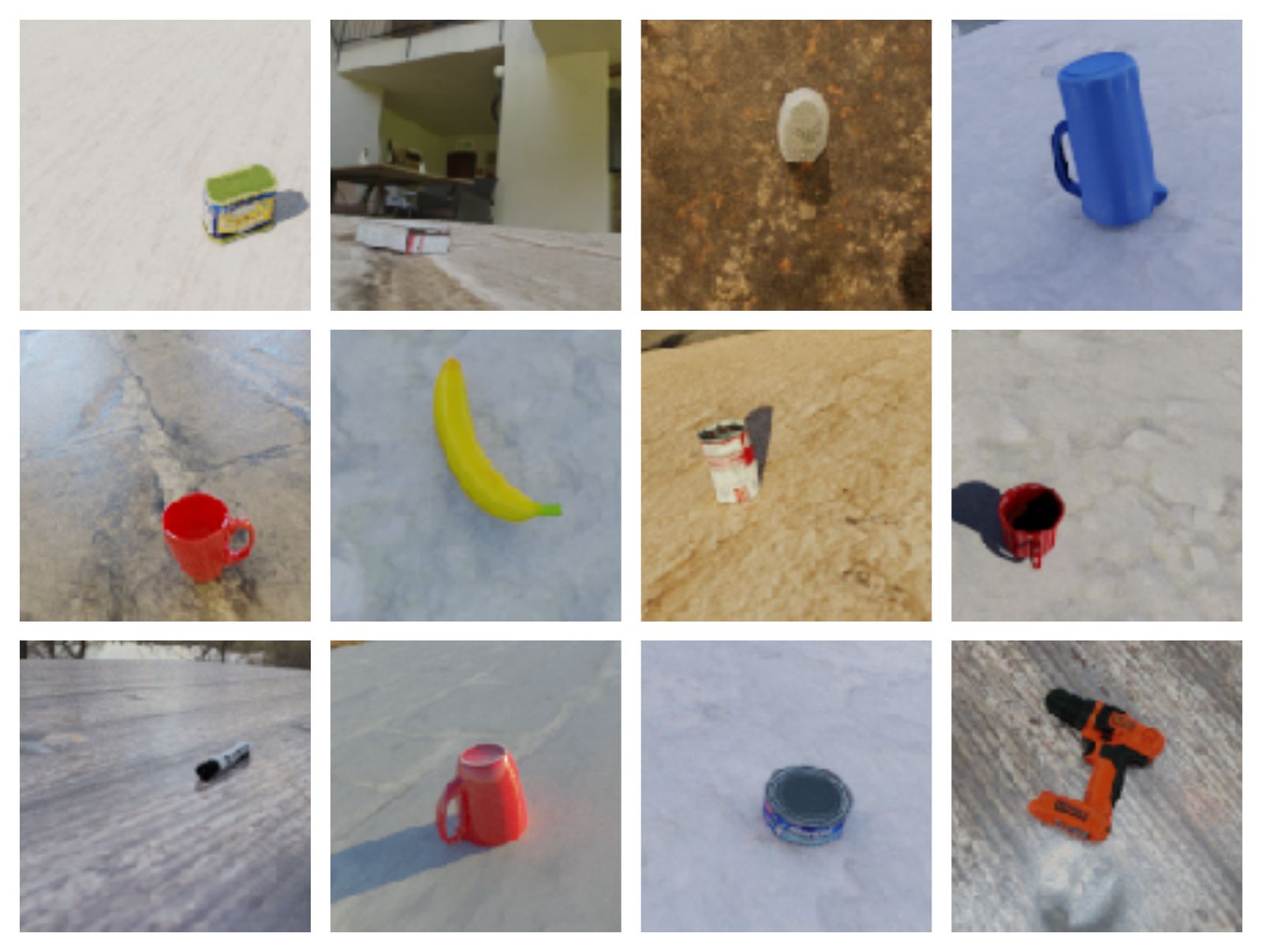}
        \caption{Example images from the \ycb out-of-distribution test set. The
        distribution shift corresponds to rendering objects on a held out set of
        textures with a different set of material properties than the
        in-distribution textures. Aside from the texture change, the set of
        objects, the lighting environments, and the sampling distribution for
        objects, poses, and lighting is held fixed between datasets.}
    \end{subfigure} \\
    \begin{subfigure}{0.6\textwidth}
        \includegraphics[width=\linewidth]{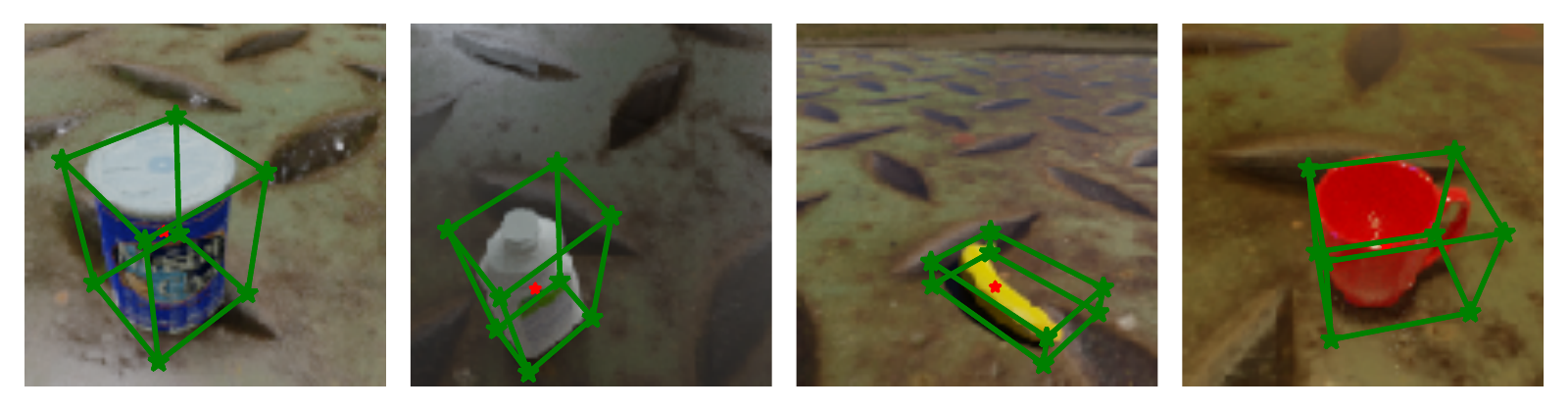}
        \caption{Examples images from the \ycb in-distribution test set shown
        with keypoint annotations. Each image is annotated with nine keypoints
        corresponding to the corners of the 3D bounding box and the object
        center. Models in the testbed predict these keypoints, and the object's
        6D pose is recovered from keypoints using the PnP algorithm~\citep{lepetit2009epnp}.}
    \end{subfigure}
    \caption{
        Examples images and keypoint annotations from the \ycb in-distribution
        and out-of-distribution (texture shift) datasets.
    }
    \label{fig:ycb_example_images}
\end{figure*}

\paragraph{Experimental details.}
We train two variants of each model. The first variant is trained with standard
$\ell_2$ loss on the distance between the predicted heatmap and the ground truth
keypoint location (with a Gaussian blur of $\sigma=0.2$). These models predict
keypoint locations by taking an $\arg\max$ over the heatmap. The other variant
is trained with and makes predictions using the integral pose regression
technique of~\citep{sun2018integral}. We train each model using SGD with
momentum and learning rate annealing.  For each model, we optimized the learning
rate in $[10^{-4}, 10^{-1}]$ and weight decay in $[10^{-4}, 1]$.
\FloatBarrier

\subsection{Variations in model hyperparameters, training duration, and training dataset size}
\label{app:linearfit_variations}
In this section, we explore the sensitivity of the linear trends discussed in
Appendix~\ref{app:linear_trends_examples} to variation in model hyperparameters,
training duration, and training set size.

We focus much of our exploration on two datasets \cifarten and \fmow. We
selected \cifarten for ease of experimentation, and we selected \fmow
in order to understand the sensitivity of the linear trends outside the context
of machine learning benchmark or synthetic shifts.

\subsubsection{\cifarten}

In Figures~\ref{fig:app_cifar102_variation},~\ref{fig:app_cinic_variation},
and~\ref{fig:app_cifar10c_fog_variation}, we probe the sensitivity of the
linear trend between in and out-of-distribution test accuracy for \cifarten
models to three types of variation: variation in hyperparameters, variation in
training duration, and variation in training set size. For ease of
visualization, we focus our experiments on three model families spanning low,
moderate, and high accuracy regimes: a ridge regression classifier on image
pixels, the random feature model from~\citet{coates2011analysis}, and
a ResNet~\citep{he2016identity}. The results are virtually identical, but
harder to visualize, when considering a larger number of model families
simultaneously.

We systematically vary the hyperparameters, number of training epochs (for the
ResNets), and the size of the training set for models from each class. We plot
model evaluations on the same linear trend line as found in
Appendix~\ref{app:linear_trends_examples}.  We show variation along these three
dimensions moves models along the linear trend line for each dataset, but does
not change the linear fit. For each of the dataset reproduction shift
\cifartentwo  (Figure~\ref{fig:app_cifar102_variation}), the benchmark shift
\cinicten (Figure~\ref{fig:app_cinic_variation}), and the synthetic \cifartenc
fog shift (Figure~\ref{fig:app_cifar10c_fog_variation}), the $R^2$ statistic of
the fit is greater than $0.99$.

\paragraph{Experimental details.}
We briefly provide details about the specific variations we consider for each
model class.
\begin{enumerate}
    \item \textbf{Hyperparameter variation:} For the ridge regression
    classifier, we vary the $\ell_2$ regularization parameter in $[10^{-6},
    10^{10}]$. For the random features models, we vary the $\ell_2$
    regularization parameter in $[10^{-4}, 10^{6}]$ and the number of random
    features in $[2^0, 2^{14}]$. For the ResNet model, we vary network depth in
    $\{18, 34, 50, 101\}$, learning rate in $[10^{-5}, 10]$, momentum in $[0.33,
    0.99]$, and weight decay in $[10^{-5}, 10^{5}]$.
    \item \textbf{Training duration variation:} To understand sensitivity to
    training duration, we save and evaluate each ResNet model after every epoch
    of training. We train each model for 350 epochs, giving 350 evaluations per
    run.
    \item \textbf{Training set size variation:} To understand sensitivity to the
    amount of training data, we subsample the \cifarten dataset from the
    original 50,000 samples to i.i.d. class-balanced subsets of size 1000, 5000,
    10000, 15000, 25000, and 40000 examples. We train each of the hyperparameter
    configurations for each model class on each of the 6 subsets of the original
    dataset and evaluate them on the same in and out-of-distribution test sets
    as before.
\end{enumerate}

\begin{figure*}[ht!]
    \centering
    \includegraphics[width=\linewidth]{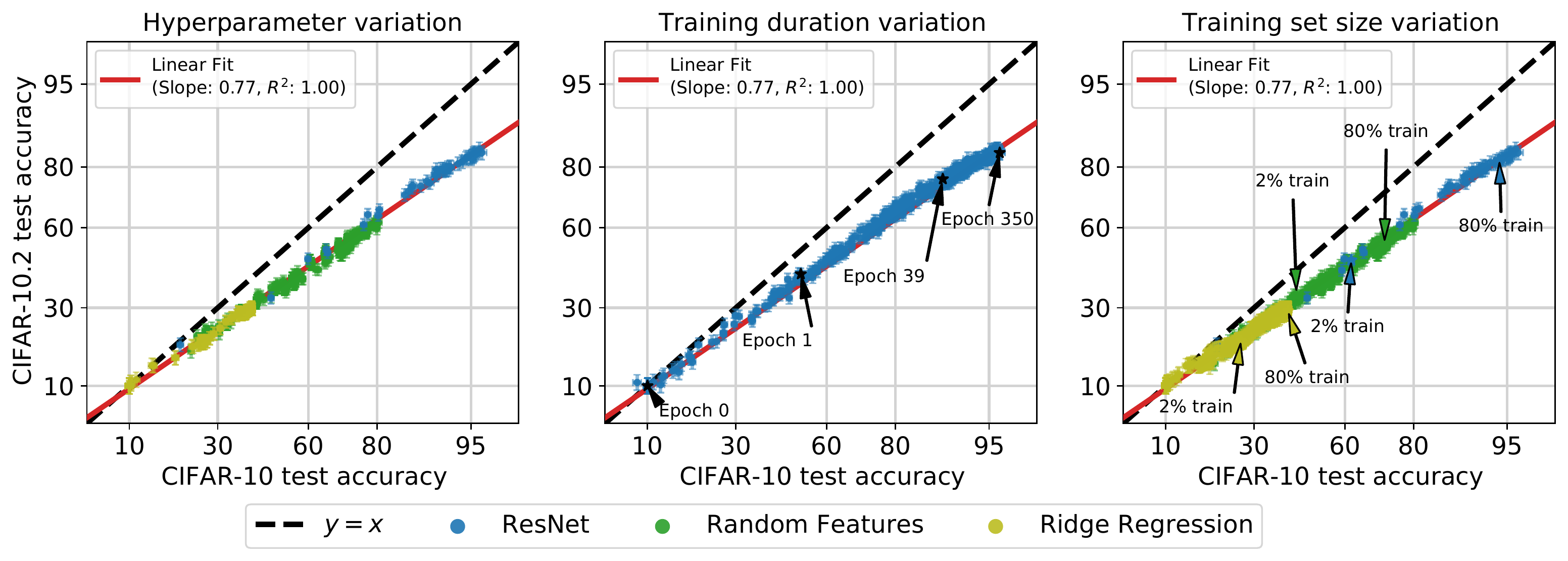}
    \caption{
        Out-of-distribution \cifartentwo accuracies vs. in-distribution \cifarten test
        accuracies under variations in model hyperparameters, training duration,
        and the size of the training set.  Each point corresponds to a model
        evaluation, shown with 95\% Clopper-Pearson confidence intervals (mostly
        covered by the point markers).
        In each panel, we compare models with the linear trend line from
        Appendix~\ref{app:linear_trends_examples}. \textbf{Left:}
        For each model family, we
        vary model-size, regularization, and optimization hyperparameters.
        \textbf{Middle:} We evaluate each network after every epoch of training.
        \textbf{Right:} We train models on randomly sampled subsets of the
        training data, ranging from 2\% to 80\% of the original \cifarten
        training set size. In each setting, variation in hyperparameters,
        training duration, or training set size moves models along the trend
        line, but does not affect the linear fit.
    }
    \label{fig:app_cifar102_variation}
\end{figure*}

\begin{figure*}[ht!]
    \centering
    \includegraphics[width=\linewidth]{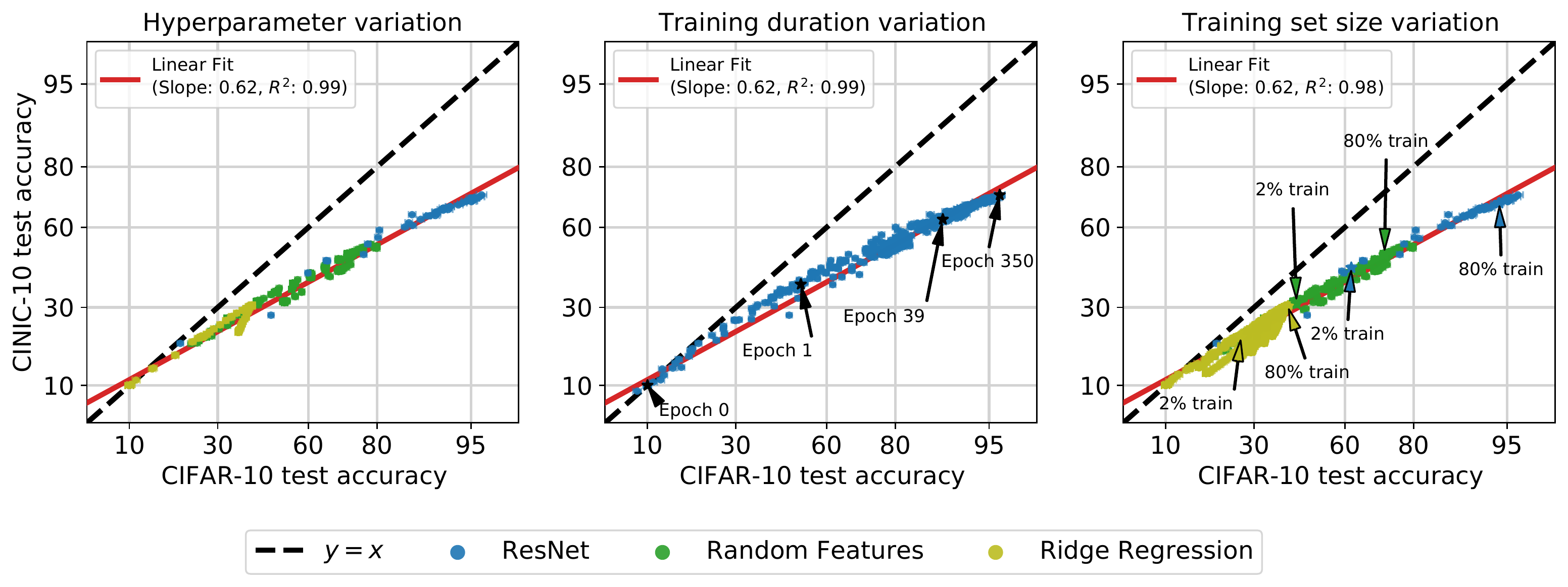}
    \caption{
        Out-of-distribution \cinicten accuracies vs. in-distribution \cifarten test
        accuracies under variations in model hyperparameters, training duration,
        and the size of the training set.  Each point corresponds to a model
        evaluation, shown with 95\% Clopper-Pearson confidence intervals (mostly
        covered by the point markers).
        In each panel, we compare models with the linear trend line from
        Appendix~\ref{app:linear_trends_examples}. \textbf{Left:}
        For each model family, we
        vary model-size, regularization, and optimization hyperparameters.
        \textbf{Middle:} We evaluate each network after every epoch of training.
        \textbf{Right:} We train models on randomly sampled subsets of the
        training data, ranging from 2\% to 80\% of the original \cifarten
        training set size. In each setting, variation in hyperparameters,
        training duration, or training set size moves models along the trend
        line, but does not affect the linear fit.
    }
    \label{fig:app_cinic_variation}
\end{figure*}

\begin{figure*}[ht!]
    \centering
    \includegraphics[width=\linewidth]{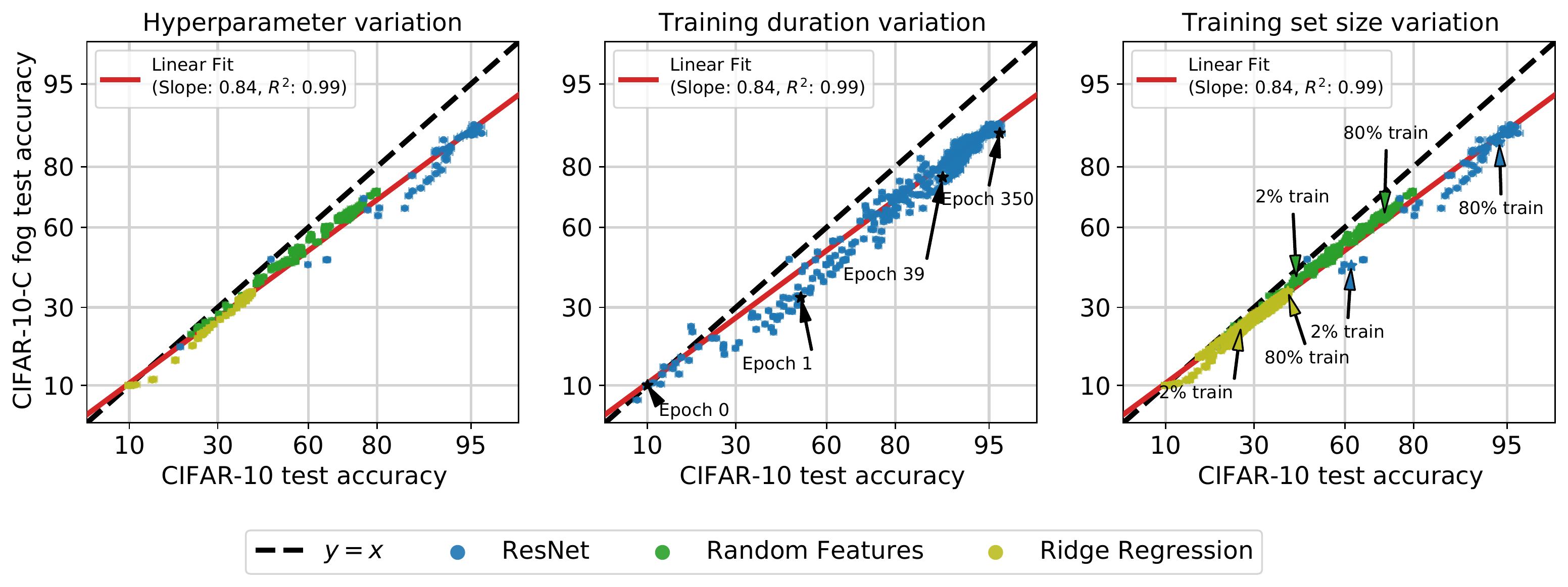}
    \caption{
        Out-of-distribution \cifartenc fog accuracies vs. in-distribution \cifarten test
        accuracies under variations in model hyperparameters, training duration,
        and the size of the training set.  Each point corresponds to a model
        evaluation, shown with 95\% Clopper-Pearson confidence intervals (mostly
        covered by the point markers).
        In each panel, we compare models with the linear trend line from
        Appendix~\ref{app:linear_trends_examples}. \textbf{Left:}
        For each model family, we
        vary model-size, regularization, and optimization hyperparameters.
        \textbf{Middle:} We evaluate each network after every epoch of training.
        \textbf{Right:} We train models on randomly sampled subsets of the
        training data, ranging from 2\% to 80\% of the original \cifarten
        training set size. In each setting, variation in hyperparameters,
        training duration, or training set size moves models along the trend
        line, but does not affect the linear fit.
    }
    \label{fig:app_cifar10c_fog_variation}
\end{figure*}
\FloatBarrier

\subsubsection{\fmow}
As in the previous section, in Figure~\ref{fig:app_fmow_variation}, we probe the
sensitivity of the linear trend for \fmow models to variation in
hyperparameters, variation in training duration, and variation in training set
size. For ease of visualization, we focus our experiments on three model
families spanning low, moderate, and high accuracy regimes: a random forest
model on image pixels, the random feature model
from~\citet{coates2011analysis}, and a ResNet~\citep{he2016identity}.
We plot model evaluations on the same linear trend line as found in
Appendix~\ref{app:linear_trends_examples}.  We show variation along these three
dimensions moves models along the linear trend line for each dataset, but does
not change the linear fit: the $R^2$ statistic of the fit is greater than $0.99$
for every setting under the accuracy metric and greater than $0.91$ for the
worst-region accuracy metric.

\paragraph{Experimental details.}
We briefly provide details about the specific variations we consider for each
model class.
\begin{enumerate}
\item \textbf{Hyperparameter variation:} For the random forest classifier, we
vary the maximum depth in $\{1, 3, 10, 20\}$, the number of trees in $\{10, 20,
50, 200\}$, and the splitting criterion between entropy and gini impurity.
For the random features models, we vary the $\ell_2$ regularization parameter in
$[10^{-4}, 10^{6}]$ and the number of random features in $[2^0, 2^{8}]$. For
the ResNet model, we vary network depth in $\{18, 34, 50, 101\}$, learning rate
in $[10^{-5}, 10]$, momentum in $[0.33, 0.99]$, and weight decay in $[10^{-5},
10^{5}]$.
\item \textbf{Training duration variation.} We train each configuration of the
ResNet for 70 epochs and evaluate each model after every epoch of training.
\item \textbf{Training set size variation.} We i.i.d. subsample the \fmow train
dataset from the original 76,863 examples to subsets of size 1000, 5000, 10000,
20000, and 50000 examples. We train each of the hyperparameter configurations
for each model class on each of the 5 subsets of the original dataset and
evaluate them on the same in and out-of-distribution test sets as before.
\end{enumerate}

\begin{figure*}[ht!]
    \centering
    \includegraphics[width=\linewidth]{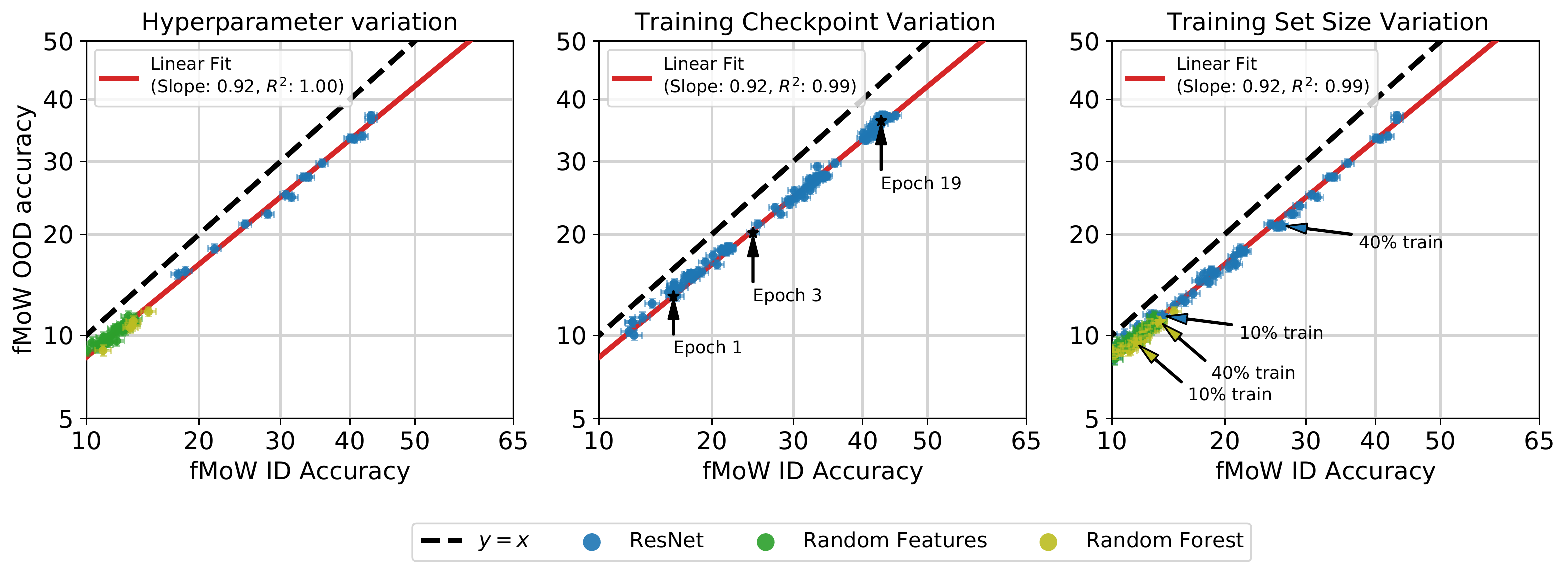}
    \includegraphics[width=\linewidth]{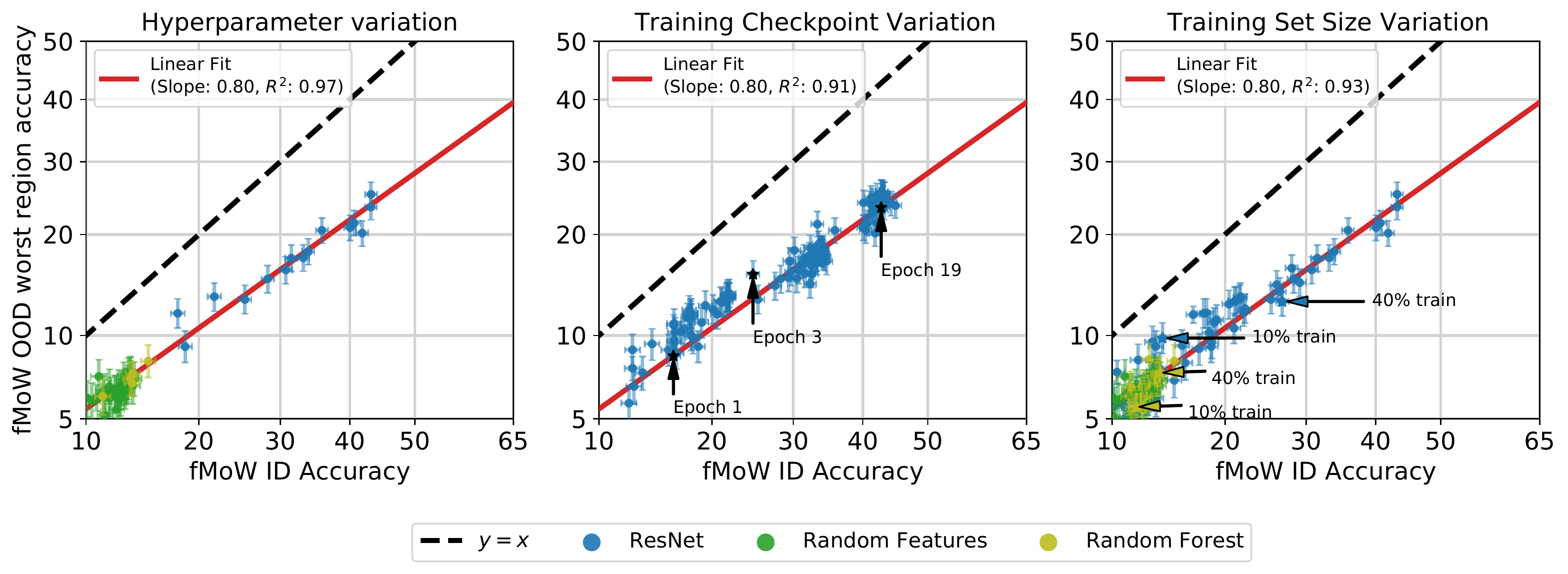}
    \vspace{-0.8cm}
    \caption{
        Out-of-distribution \fmow test accuracies vs. in-distribution \fmow test
        accuracies under variations in model hyperparameters, training duration,
        and the size of the training set.  Each point corresponds to a model
        evaluation, shown with 95\% Clopper-Pearson confidence intervals.
        In each panel, we compare models with the linear trend line from
        Appendix~\ref{app:linear_trends_examples}. The top row compares model
        trends using average accuracy as the OOD metric, and the bottom rows
        uses worst-region accuracy as the OOD metric. \textbf{Left:}
        For each model family, we vary model-size, regularization, and
        optimization hyperparameters.  \textbf{Middle:} We evaluate each network
        after every epoch of training.  \textbf{Right:} We train models on
        randomly sampled subsets of the training data, ranging from 2\% to 80\%
        of the original \cifarten training set size. In each setting, variation
        in hyperparameters, training duration, or training set size moves models
        along the trend line, but does not affect the linear fit.
    }
    \label{fig:app_fmow_variation}
\end{figure*}
\FloatBarrier

\subsubsection{\ycb}
In Figure~\ref{fig:app_ycb_variation}, we see that the linear fit for the \ycb
experiment from Appendix~\ref{app:linearycb} is also invariant to changes in the
amount of training data.

\begin{center}
\begin{figure*}[ht!]
    \centering
    \includegraphics[width=0.8\linewidth]{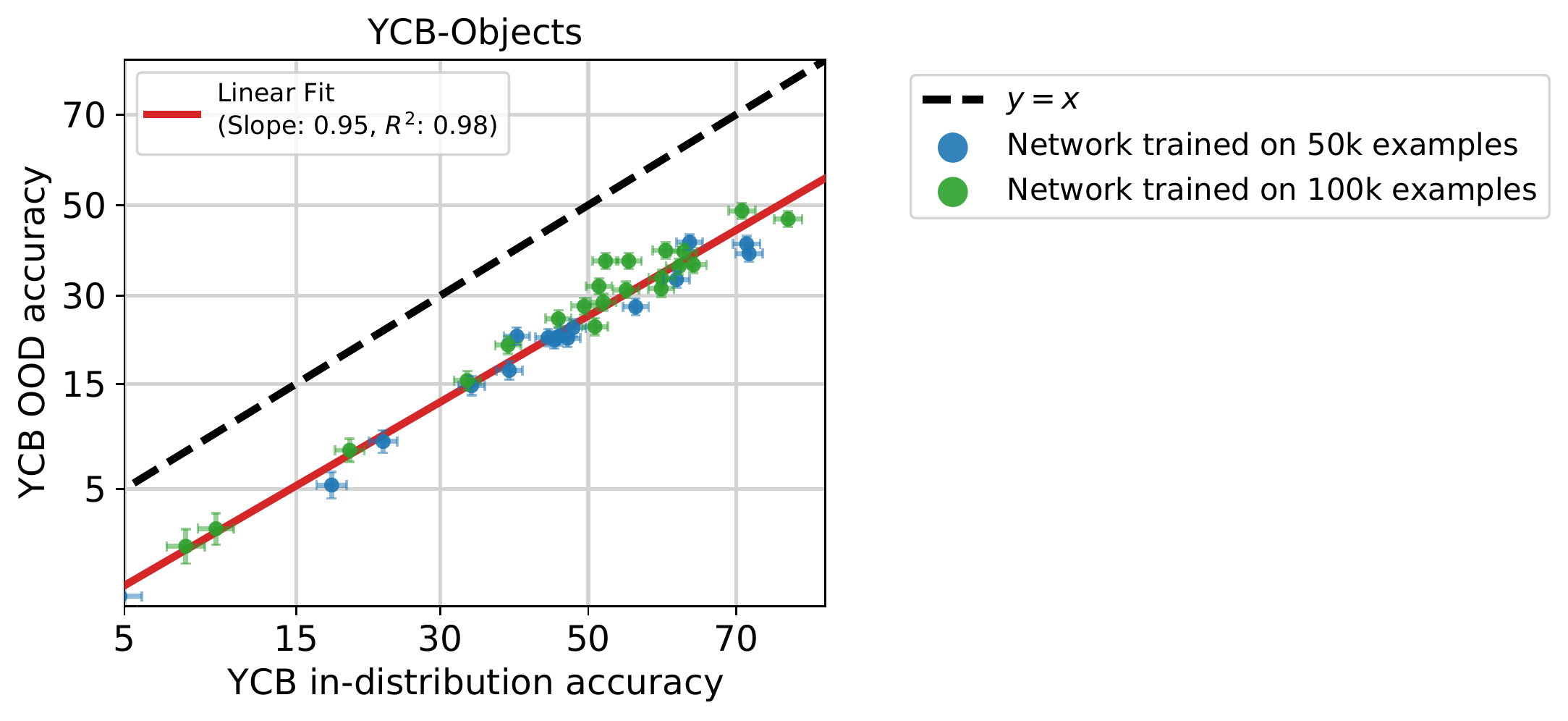}
    \caption{
        Out-of-distribution \ycb test accuracies vs. in-distribution \ycb test
        accuracies under variations in training set size.  Each point
        corresponds to a model evaluation, shown with 95\% Clopper-Pearson
        confidence intervals. The linear trend line is the same as
        Figure~\ref{fig:ycb_appendix_fits}. The linear trend still well explains
        the data ($R^2=0.98$), and increasing training set size moves models
        along the linear trend, but does not affect the linear fit.
    }
    \label{fig:app_ycb_variation}
\end{figure*}
\end{center}
\FloatBarrier

\subsection{Comparison of axis scaling}
\begin{center}
\begin{figure*}[ht!]
    \centering
    \includegraphics[width=0.8\linewidth]{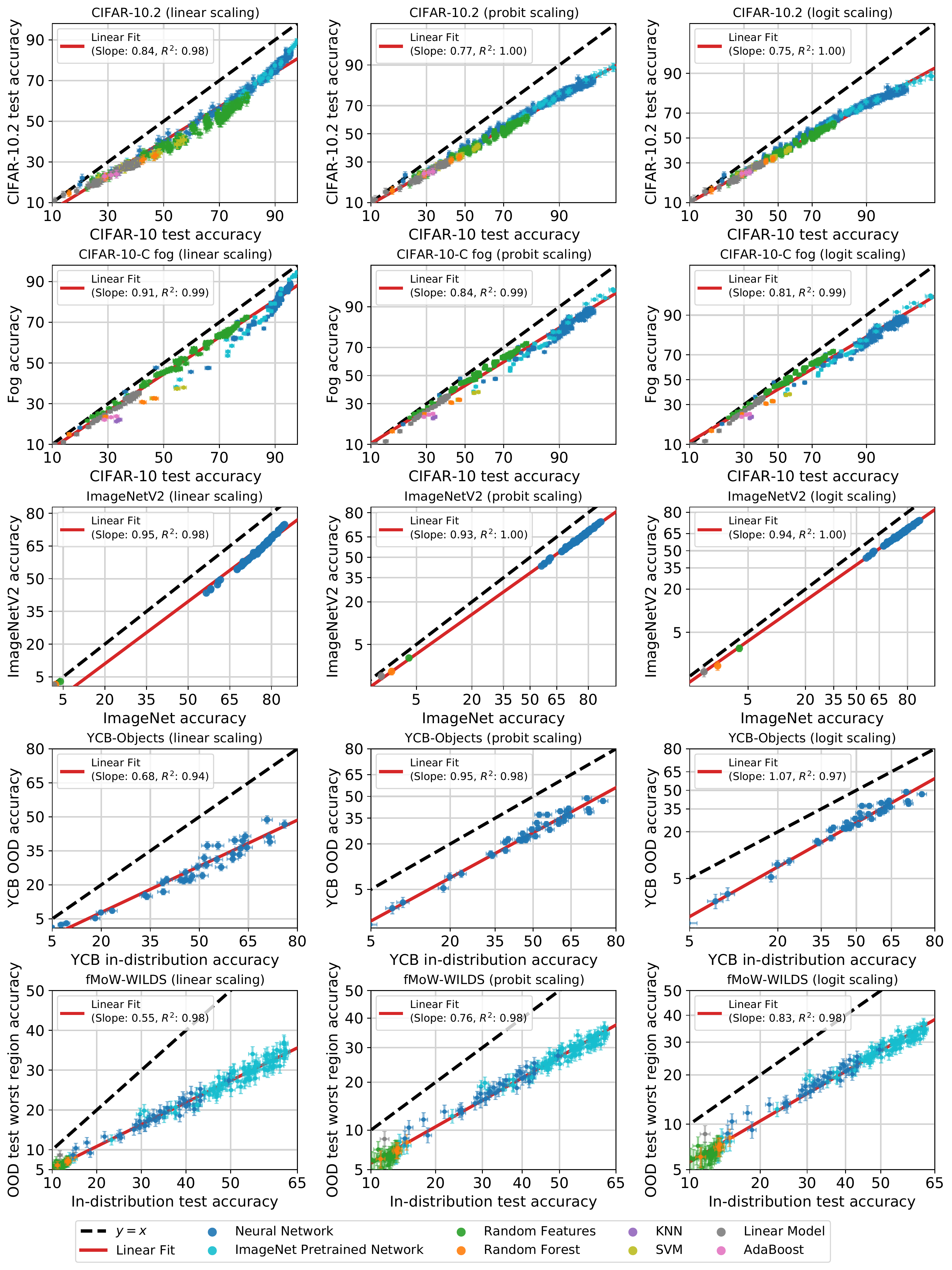}
    \caption{
        OOD test accuracies vs. ID test accuracies for several pairs of ID and
        OOD test sets visualized with three different axis scalings.
        \textbf{Left:} The left column shows model accuracies with a linear axis
        scaling.  \textbf{Middle:} The middle column shows model accuracies with
        a probit scale on both axes. In other words, model accuracy $x$ appears
        at $\Phi^{-1}(x)$ where $\Phi^{-1}$ is the inverse Gaussian CDF.
        \textbf{Right:} The right column shows model accuracies with a logit
        scale on both axes: model accuracy $x$ appears at $\sigma^{-1}(x)$ where
        $\sigma^{-1}$ is the inverse logistic function.  Visual inspection shows
        the linear fit is better in the logit or probit domain, especially when
        model accuracies span a wide range.  Quantitatively, the R2 statistics
        are higher in the probit or logit domains than with linear axis scaling.
        For instance, on ImageNetV2 and CIFAR-10.2, the $R^2$ is $0.98$ in the
        linear domain compared to $1.0$ in the probit or logit domains.
    }
    \label{fig:app_scaling_comparison}
\end{figure*}
\end{center}
\FloatBarrier

\section{Distribution shifts with weaker correlations}
\label{app:fit_failure}

\subsection{\camelyon}
\label{app:camelyon}

In this section, we first explore the role of training randomness on the observed ID-OOD correlation for \camelyon.
Remember that in Figure \ref{fig:camelyon}, we found a very high degree of variability between
ID and OOD performance.
To see if the performance variation was due to training randomness, we train each model
ten times and then average the final model accuracies together.
The result of these averaged runs is displayed in Figure \ref{fig:camelyon_radomness} (left).
The $R^2$ value for the averaged runs comes in at $R^2=0.39$, which is approximately
equivalent to the $R^2$ value in Figure \ref{fig:camelyon} ($R^2=0.40$);
this suggests that training randomness is not enough to account for the performance variability.

In Figure \ref{fig:camelyon_radomness} (right), we also attempted early-stopping each trained model
on a separate OOD validation set
(different in distribution from the OOD test set), as is recommended in \cite{koh2020wilds},
before averaging model accuracies; the result is largely unchanged and comes in at $R^2=0.46$.
Early-stopping on the in-distribution validation set, however, does increase ID-OOD correlation
significantly to $R^2=0.77$, as seen in Figure \ref{fig:camelyon_radomness} (middle);
further investigating the mechanisms at play here is an interesting direction for future work.

\begin{figure*}[ht!]
    \centering
    \includegraphics[width=\linewidth]{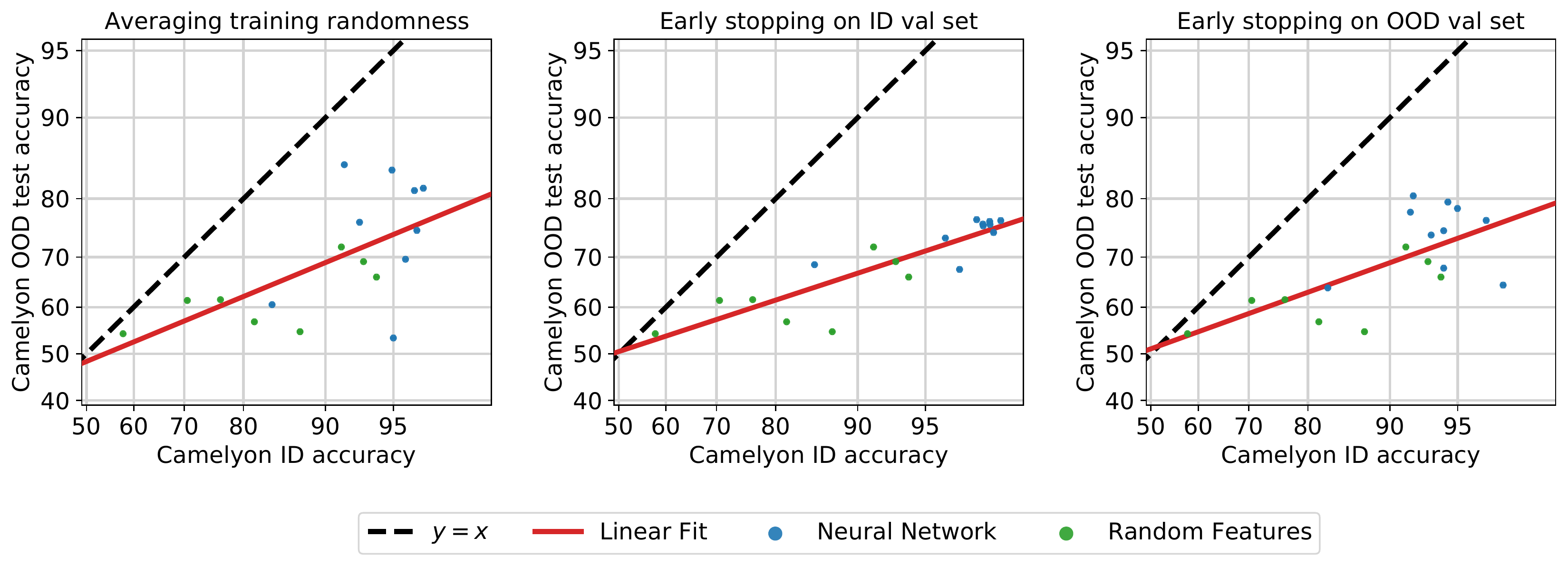}
    \vspace{-0.6cm}
    \caption{
		Model accuracies on the \camelyon distribution shift.
		Each point gives average accuracies for models trained with ten different random seeds,
		and error bars give the standard deviation.
		\textbf{Left:} Models trained to convergence and then averaged over seeds.
		\textbf{Middle:} Each model is early-stopped on the ID validation set then averaged over seeds.
		\textbf{Right:} Each model is early-stopped on the OOD validation set then averaged over seeds.
    }
    \label{fig:camelyon_radomness}
\end{figure*}

\begin{figure*}[ht!]
    \centering
    \includegraphics[width=\linewidth]{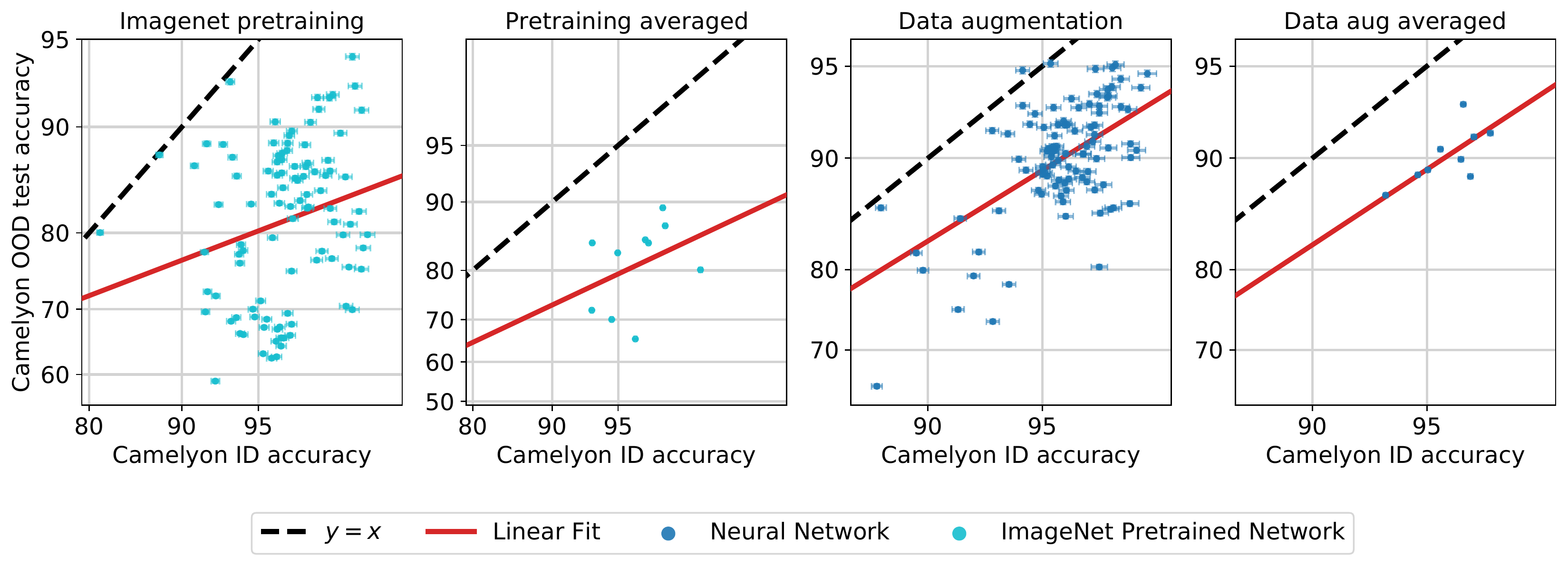}
    \vspace{-0.6cm}
    \caption{
		Model accuracies on the \camelyon distribution shift.
		\textbf{Left:} ImageNet pretrained models finetuned to convergence.
		\textbf{Middle Left:} ImageNet pretrained model accuracies averaged over ten random seeds.
		\textbf{Middle Right:} Models trained with targeted color jitter data augmentation.
		\textbf{Right:} Data augmentation model accuracies averaged over ten random seeds.
    }
    \label{fig:camelyon_ablations}
\end{figure*}

As a next step, to study the role of the training data distribution on the observed trends,
we conduct two specific training-time interventions: pretraining on \imagenet, and training using a
specific color-jitter data augmentation.

We show results for models pretrained on \imagenet{} in Figure \ref{fig:camelyon_ablations} (left).
As is evident, the variability in model performance is still extremely high ($R^2=0.05$).
Averaging over training randomness does not seem to help either ($R^2=0.14$).

We also train using a domain-specific color-jitter data augmentation designed to mimic the visual differences in samples from different hospitals,
a technique that has previously been found to have been beneficial on a similar task \cite{tellez2018whole,tellez2019quantifying}.
As seen in Figure \ref{fig:camelyon_ablations} (middle right), training with the data augmentation both considerably increases
average OOD performance and significantly reduces the amount of OOD accuracy variation ($R^2=0.77$).
However, even with the targeted data augmentation, large OOD accuracy fluctuations still exist.
Averaging over training randomness greatly increases the correlation further and mitigates these fluctuations ($R^2=0.95$),
as seen in Figure \ref{fig:camelyon_ablations} (right);
however, the data augmentation causes all models to have relatively high ID accuracy, and it is unclear whether this tight trend
would hold for models in the low accuracy regime as well.

One possible reason for the high variation in accuracy is the correlation across image patches. 
Image patches extracted from the same slides and hospitals are correlated because patches from the same slide are from the same lymph node section, and patches from the same hospital were processed with the same staining and imaging protocol. 
In addition, patches in \camelyon are extracted from a relatively small number of slides (the dataset includes 50 slides total, and there are 10 slides from a single hospital in the OOD test set \cite{koh2020wilds}).
Prior work in the context of natural language processing tasks have shown that these correlations can result in instabilities in both training and evaluation \citep{zhou2020curse}, and investigating their effect on OOD variation in \camelyon is interesting future work.


As an initial exploration of the effect of highly correlated test examples, we observe that correlated examples can result in high OOD variation in a simulated environment on \cifarten and \cifartentwo.
Concretely, we subsample \cifarten and \cifartentwo and then apply data augmentation to each example to generate a test set of the same size as the original but with significant correlation between examples. 
In each panel in Figure \ref{fig:camelyon_effective_test_size}, we train models on \cifarten and then evaluate them on \cifarten
and \cifartentwo with effective test size $k$ for varying $k$.
Concretely, we subsample $k$ images from each class, and then apply
RandAugment {\tt rand-m9-mstd0.5-inc1}~\citep{cubuk2020randaugment} to
each example to generate test sets of size 10,000. 
We work with a binary version of \cifarten and \cifartentwo,
restricting both datasets to two classes: {\tt airplanes} and {\tt
cats}. 
When the effective test set size is small, e.g. $k=1$ or
$k=2$, the linear fit is very poor. 
However, as the effective test set size $k$
increases to $k=100$ or $k=500$, the linear fit is much better ($R^2 =
0.94$ vs. $R^2 = 0.66$), and the variance between model evaluations is
substantially smaller.

In contrast to highly correlated test examples, highly correlated
\emph{training} examples appears to have substantially less effect on the amount
of OOD variation or the quality of the linear fit. Using the same simulated
\cifarten and \cifartentwo environment as the previous paragraph, we generate a
sequence of training sets with varying degrees of correlation between training
examples. Concretely, we subsample the \cifarten and \cifartentwo training sets and then
apply data augmentation (RandAugment~\citep{cubuk2020randaugment}) to each
example to generate a training set of the same size as the original but with
significant correlation between examples.  In each panel in Figure
\ref{fig:camelyon_effective_train_size}, we train models on \cifarten and then
evaluate them on \cifarten. Even with the effective training set size is small,
e.g. $k=2$, the linear fit is fairly good ($R^2 = 0.89$),
and there is substantially smaller variance between model evaluations than in
the corresponding effective test-size experiment.

\begin{figure*}[ht!]
    \centering
    \includegraphics[width=\linewidth]{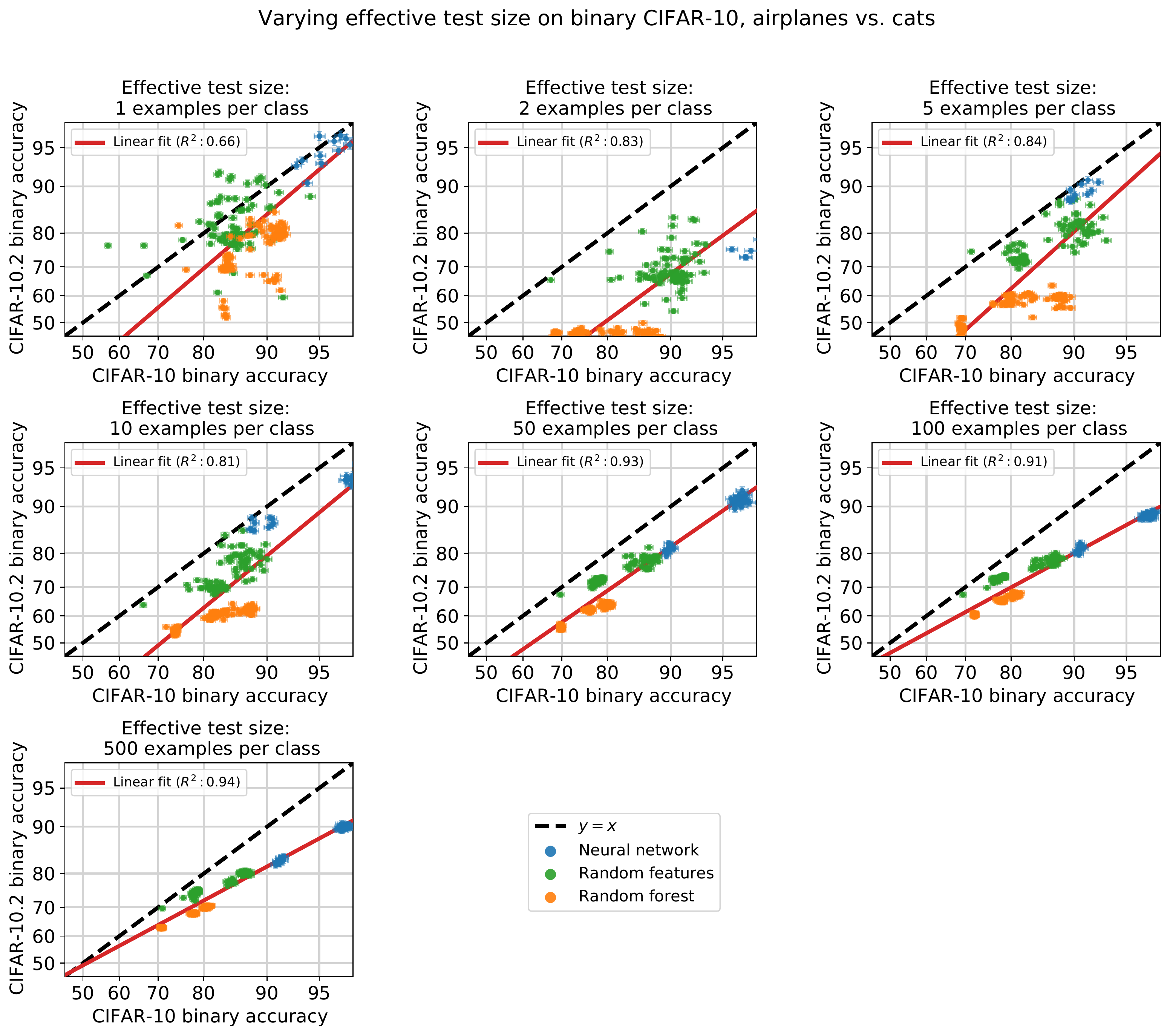}
    \caption{
        Models trained on \cifarten and evaluated on \cifartentwo 
        for binary classification: {\tt airplanes} vs {\tt cats}.
        Each panel depicts evaluating the models with varying effective test set sizes $k$,
        where $k$ images are subsampled from each class and then repeated
        data-augmented using RandAugment~\citep{cubuk2020randaugment} to generate a
        consistent test set size of 10,000 examples.  For smaller effective test
        set sizes, the linear fit is very poor, and this variance decreases
        substantially for larger $k$. 
    }
    \label{fig:camelyon_effective_test_size}
\end{figure*}

\begin{figure*}[ht!]
    \centering
    \includegraphics[width=\linewidth]{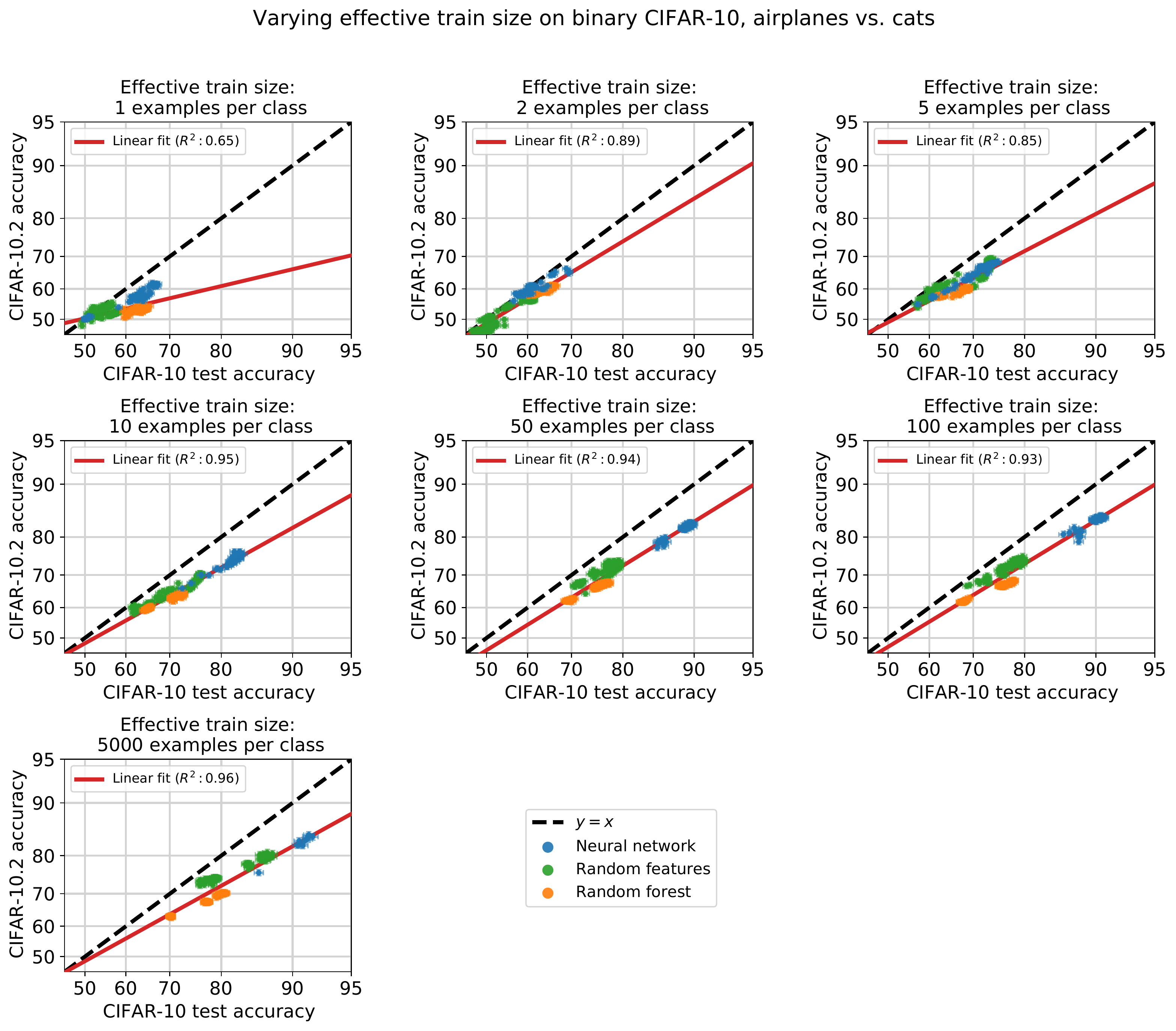}
    \caption{
        Models trained on \cifarten and evaluated on \cifartentwo 
        for binary classification: {\tt airplanes} vs {\tt cats}.
        Each panel depicts evaluating the models with varying effective train set sizes $k$,
        where $k$ images are subsampled from each class and then repeatedly data-augmented
        using RandAugment~\citep{cubuk2020randaugment} to generate a consistent train set
        size of 50,000 examples.  In contrast to varying the effective test set
        size (see Figure~\ref{fig:camelyon_effective_train_size}), varying the
        effective train set has little effect on the quality of the linear fit.
        For instance, with as few as two effective examples per class, the linear fit is
        fairly precise ($R^2 = 0.89$).
    }
    \label{fig:camelyon_effective_train_size}
\end{figure*}
\FloatBarrier

\subsection{\cifartenc}
\label{app:corruptions}

In this section, we look at distribution shifts induced by image corruptions in more detail.
Specifically, in Figures \ref{fig:corruptions1}--\ref{fig:corruptions5}, we plot neural networks
trained on either \cifarten or \imagenet and evaluated on a similar set of image corruptions.
Interestingly, the choice of corruption can have a significant effect on the strength of the linear
trend between ID and OOD accuracy, as we have already explored in Sections \ref{sec:linear_trends}
and \ref{sec:fit_failure}.
Comparing the plots in Figures \ref{fig:corruptions1}--\ref{fig:corruptions5} side-by-side,
we also observe that many corruptions behave more linearly on \imagenetc{} than on \cifartenc.
Investigating this discrepancy further is an interesting direction for future work.

\begin{figure*}[ht!]
    \centering
	\includegraphics[width=0.95\linewidth]{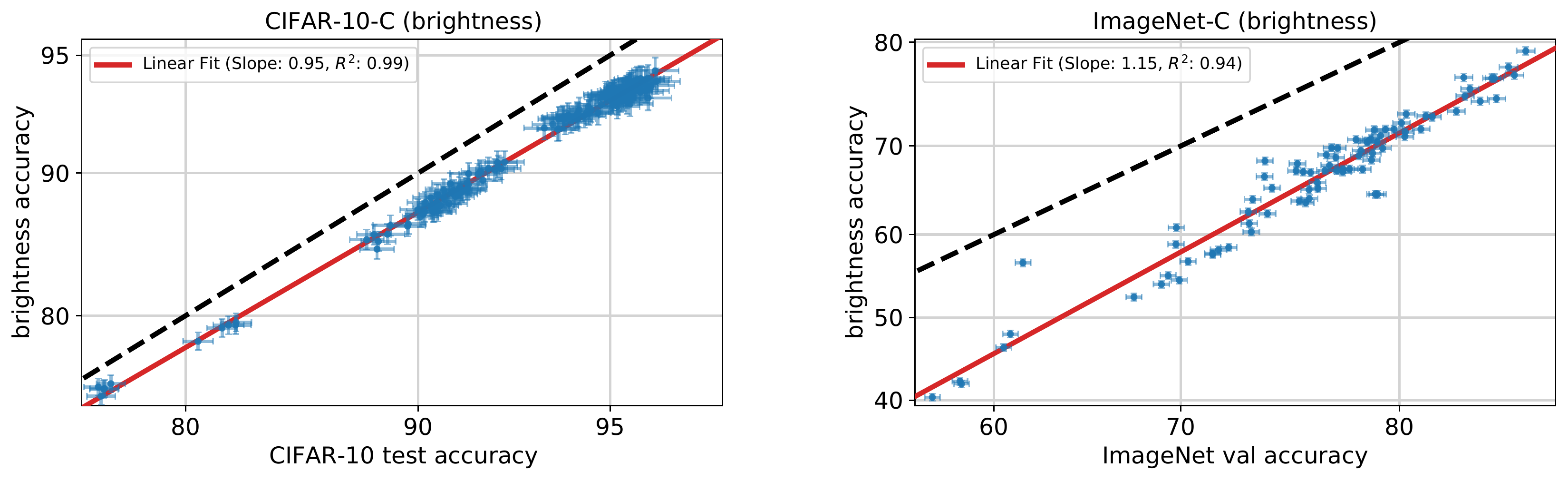}
	\includegraphics[width=0.95\linewidth]{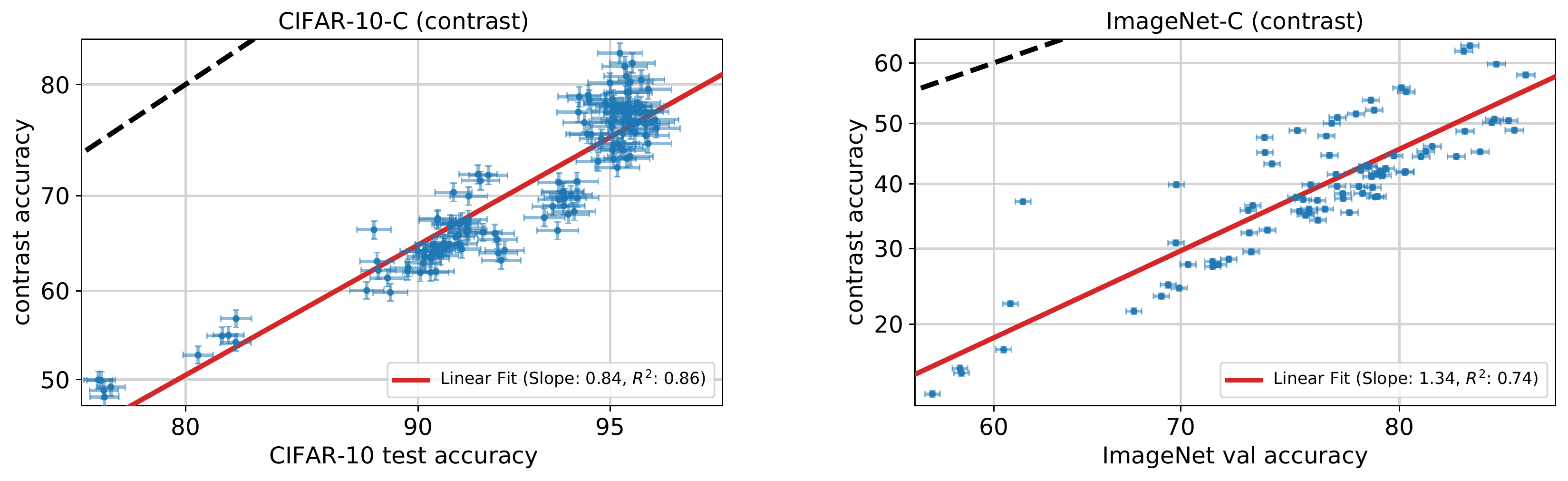}
	\includegraphics[width=0.95\linewidth]{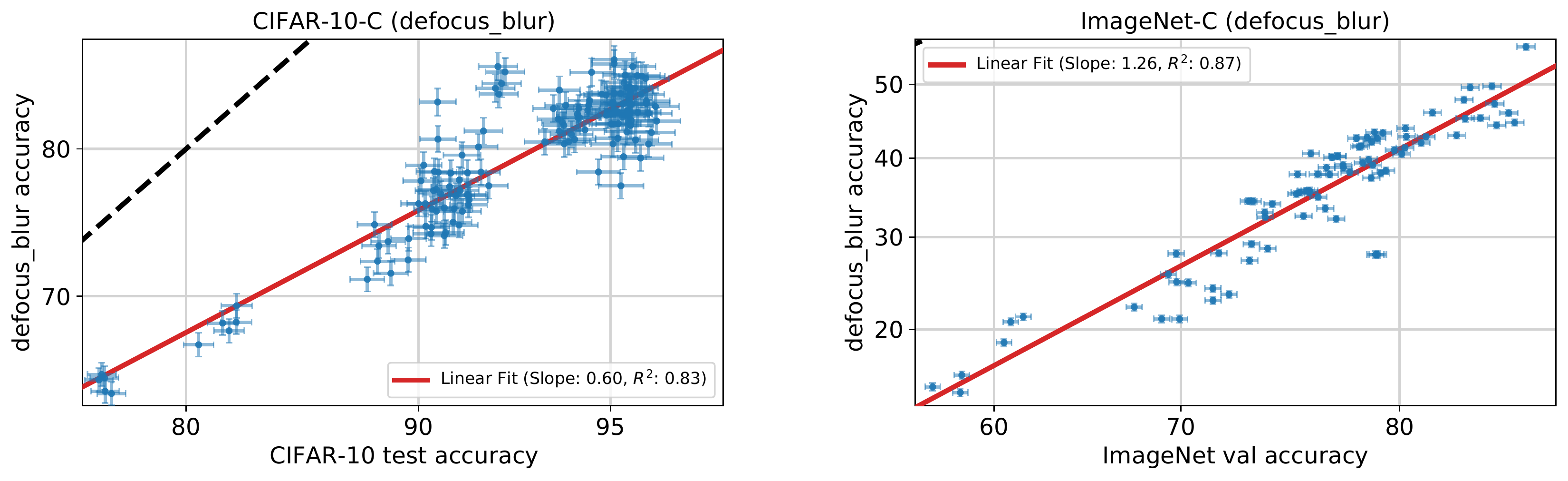}
	\includegraphics[width=0.95\linewidth]{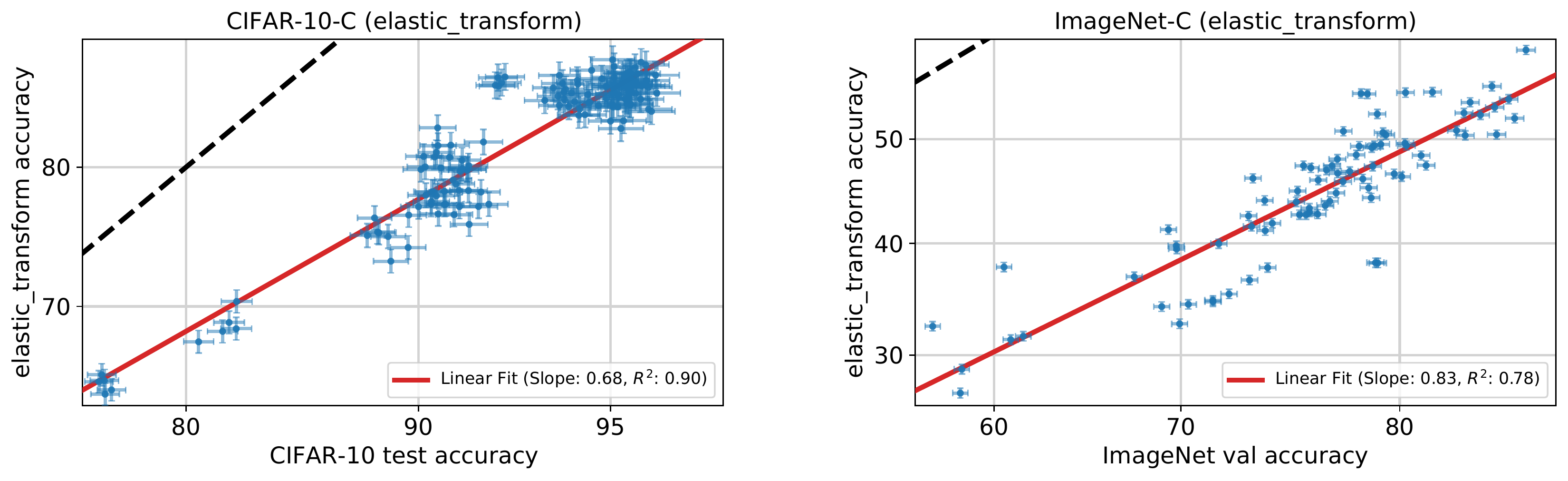}
	\includegraphics[width=0.3\linewidth]{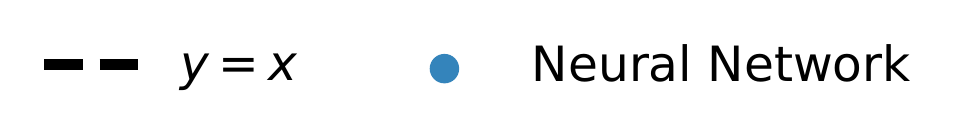}
    \vspace{-0.2cm}
    \caption{
		Models trained on either \cifarten{} (left) or \imagenet{} (right) and evaluated
		under distribution shift due to image corruptions.
		This figure continues for the next few pages.
    }
    \label{fig:corruptions1}
\end{figure*}

\begin{figure*}[ht!]
    \centering
	\includegraphics[width=0.95\linewidth]{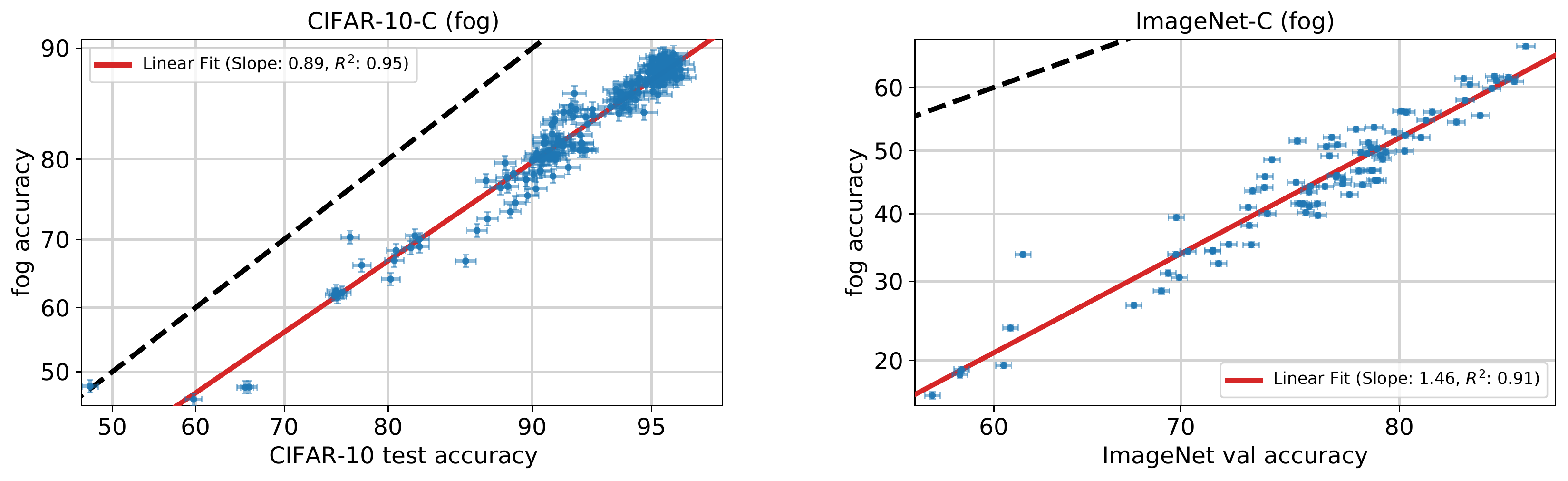}
	\includegraphics[width=0.95\linewidth]{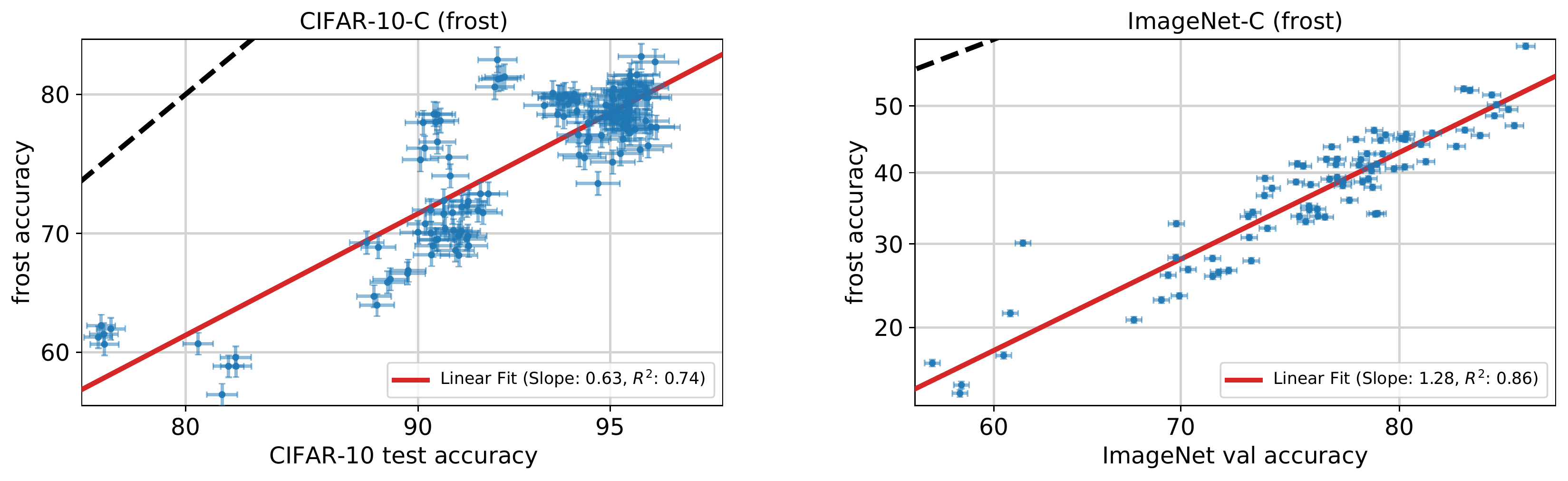}
	\includegraphics[width=0.95\linewidth]{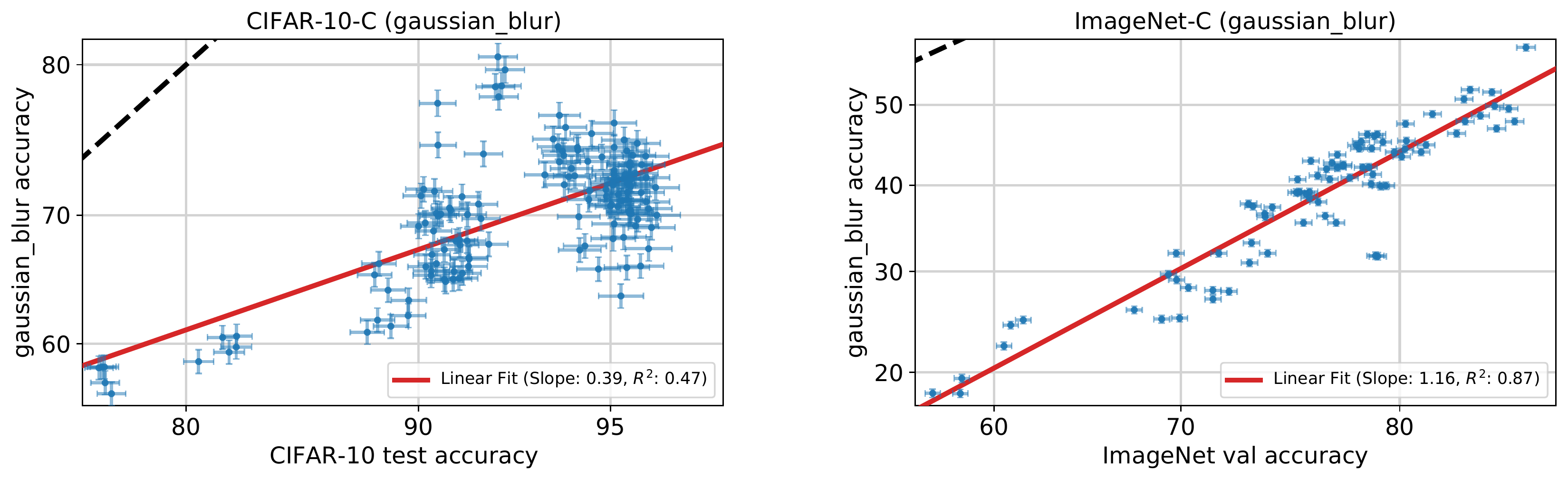}
	\includegraphics[width=0.95\linewidth]{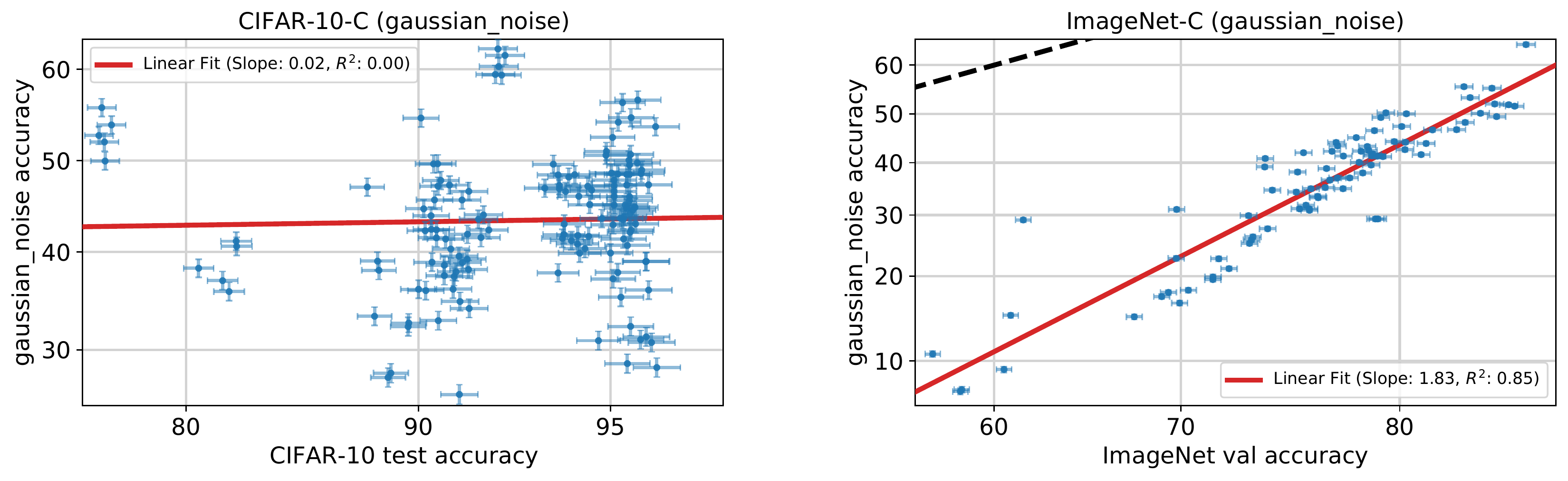}
	\includegraphics[width=0.3\linewidth]{figures/corruptions/corruption_legend.pdf}
    \vspace{-0.2cm}
    \caption{
		Continuation of the corruption plots.
    }
    \label{fig:corruptions2}
\end{figure*}

\begin{figure*}[ht!]
    \centering
	\includegraphics[width=0.95\linewidth]{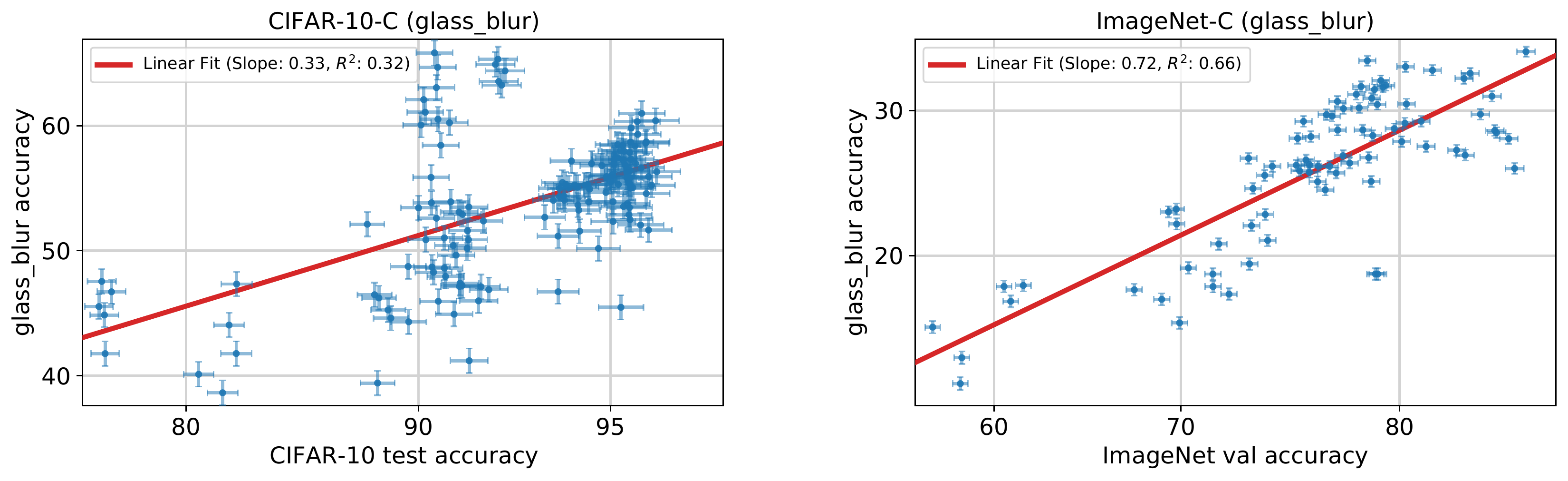}
	\includegraphics[width=0.95\linewidth]{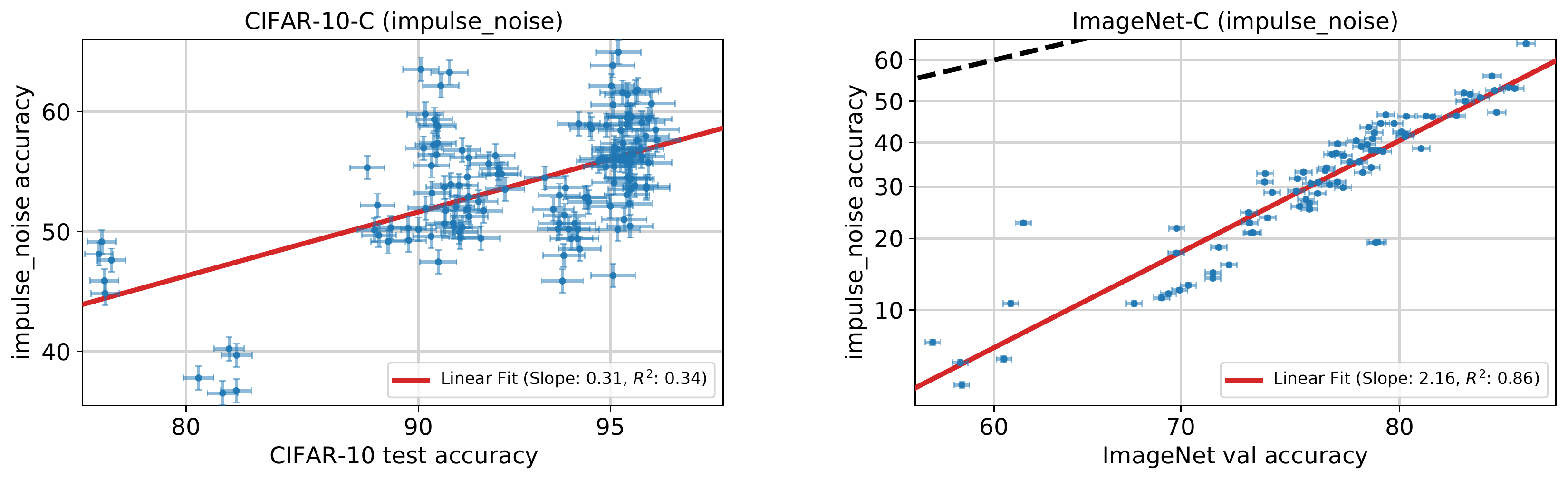}
	\includegraphics[width=0.95\linewidth]{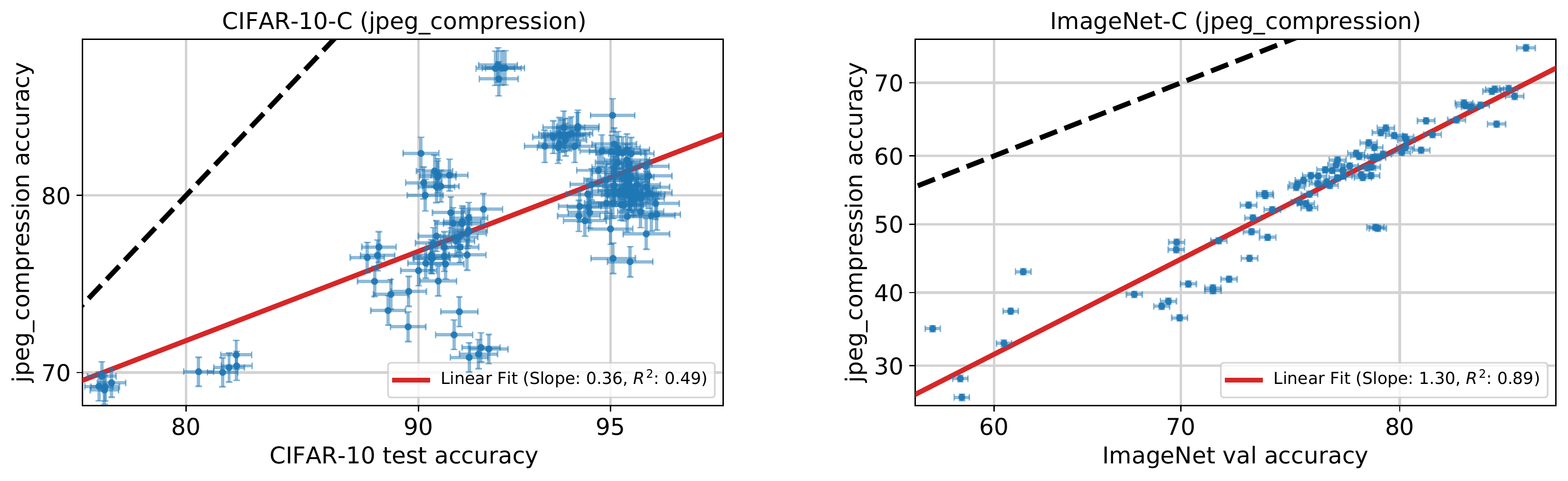}
	\includegraphics[width=0.95\linewidth]{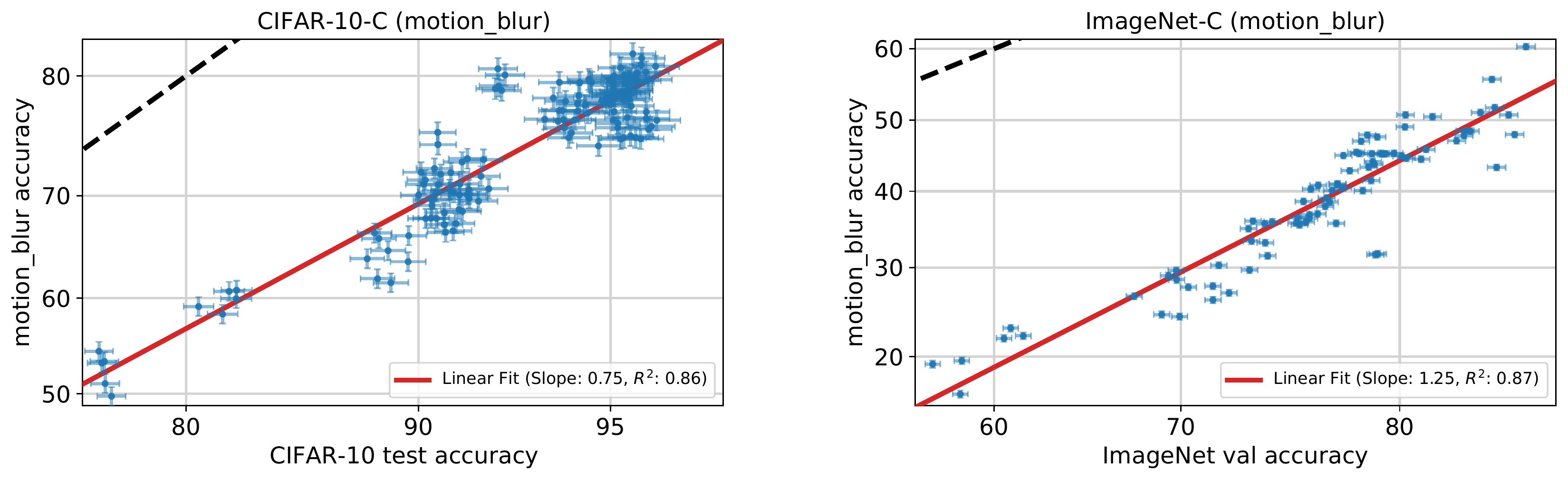}
	\includegraphics[width=0.3\linewidth]{figures/corruptions/corruption_legend.pdf}
    \vspace{-0.2cm}
    \caption{
		Continuation of the corruption plots.
    }
    \label{fig:corruptions3}
\end{figure*}

\begin{figure*}[ht!]
    \centering
	\includegraphics[width=0.95\linewidth]{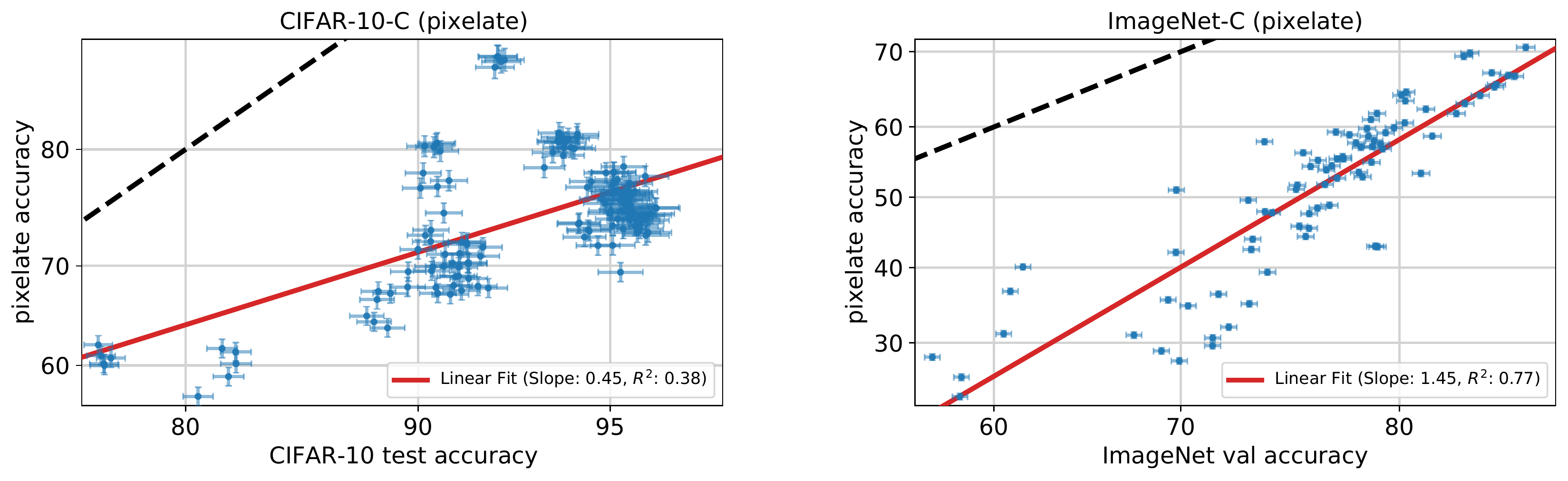}
	\includegraphics[width=0.95\linewidth]{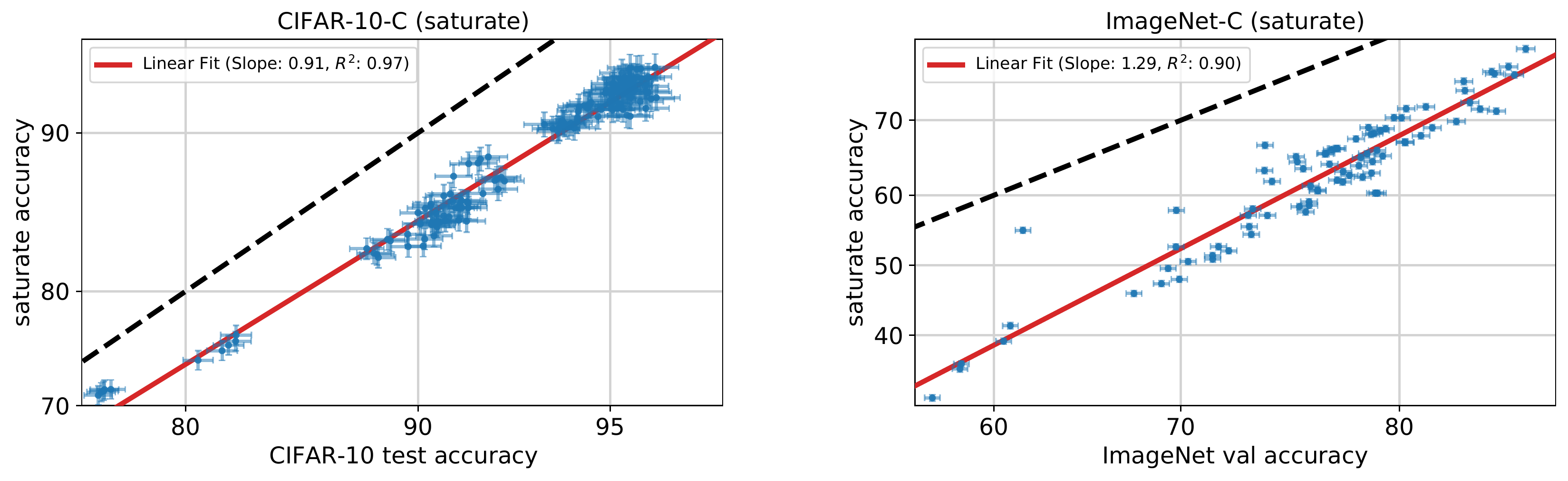}
	\includegraphics[width=0.95\linewidth]{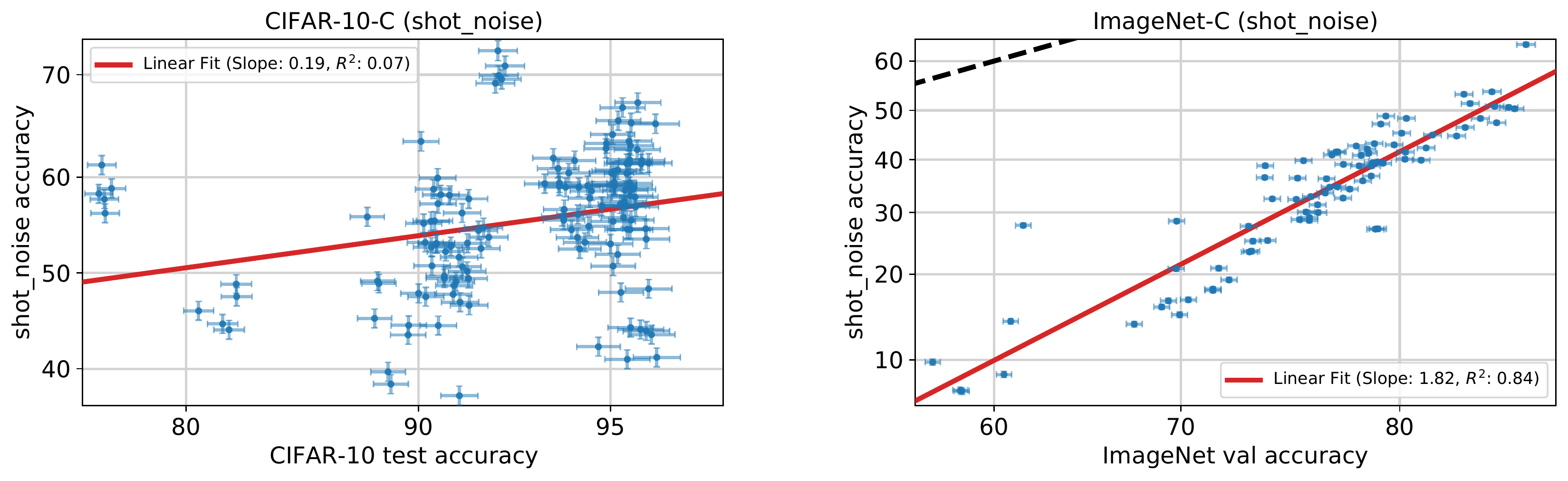}
	\includegraphics[width=0.95\linewidth]{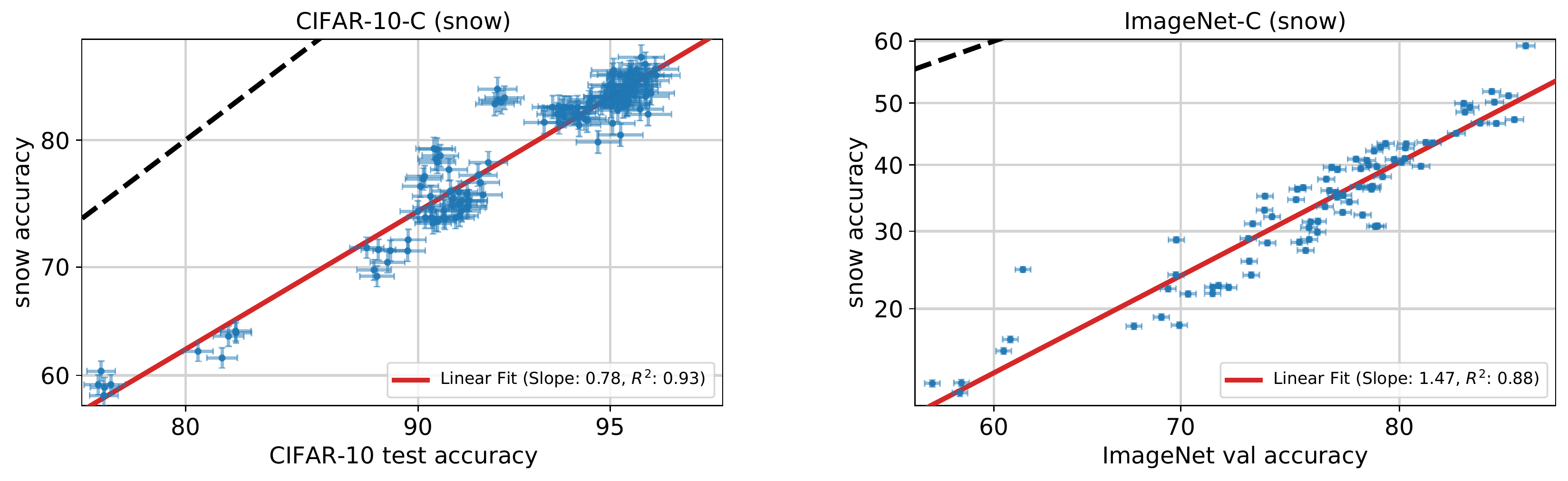}
	\includegraphics[width=0.3\linewidth]{figures/corruptions/corruption_legend.pdf}
    \vspace{-0.2cm}
    \caption{
		Continuation of the corruption plots.
    }
    \label{fig:corruptions4}
\end{figure*}

\clearpage

\begin{figure*}[ht!]
    \centering
	\includegraphics[width=0.95\linewidth]{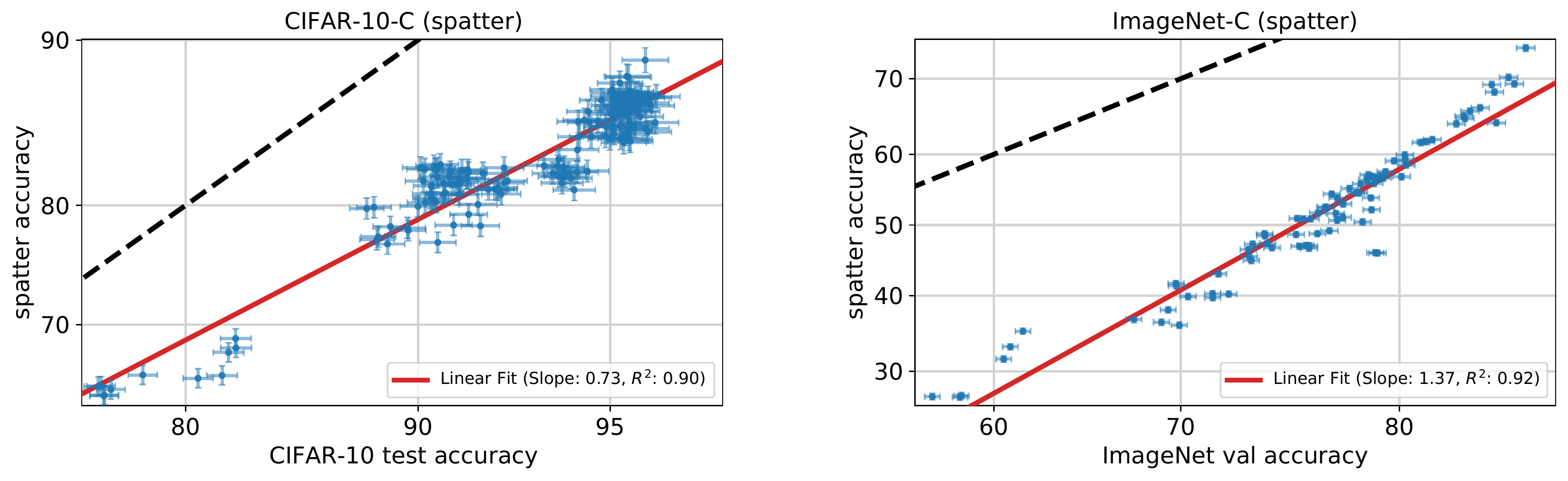}
	\includegraphics[width=0.95\linewidth]{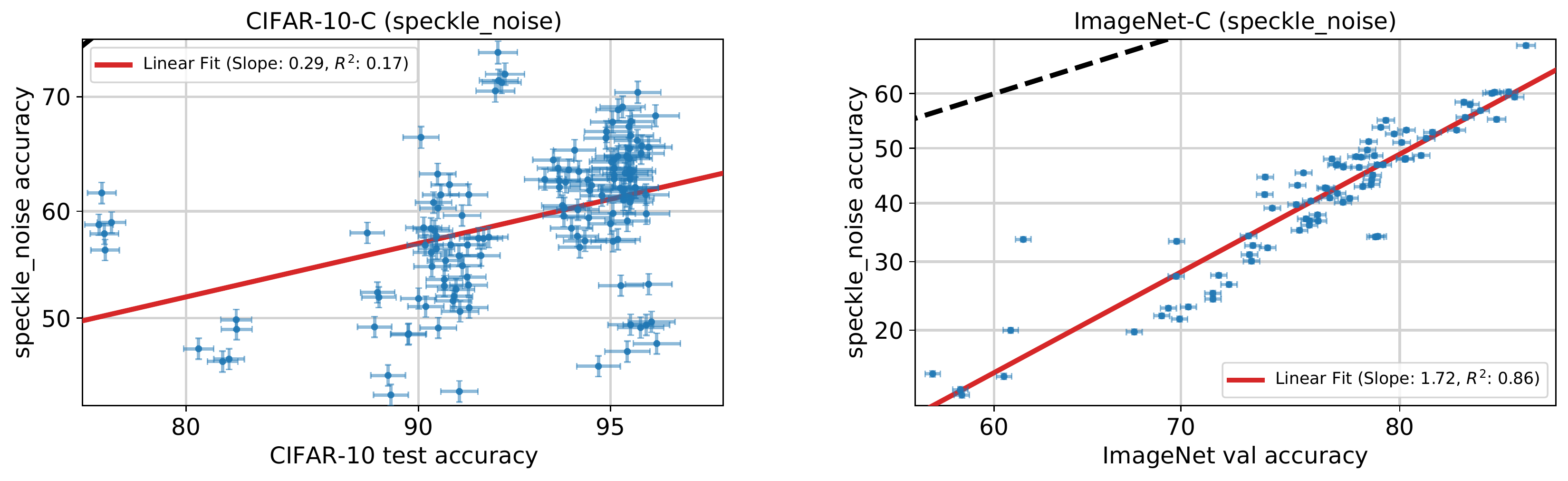}
	\includegraphics[width=0.95\linewidth]{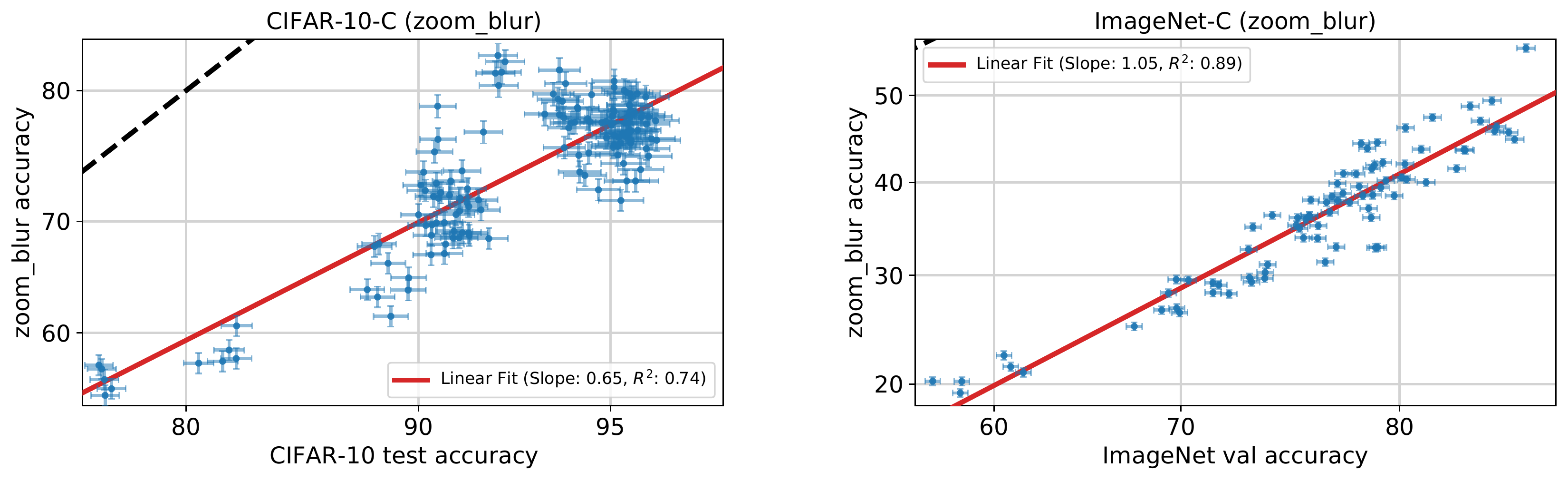}
	\includegraphics[width=0.3\linewidth]{figures/corruptions/corruption_legend.pdf}
    \vspace{-0.2cm}
    \caption{
		Continuation of the corruption plots.
    }
    \label{fig:corruptions5}
\end{figure*}

\iftoggle{isarxiv}{}{
\subsection{CIFAR10-C Gaussian covariance}
\label{app:cifar10cgaussian}
In this section, we investigate the relationship betweeen the in-distribution and out-of-distribution
data covariances, in line with our theoretical model from Section~\ref{sec:theory}.
The theoretical model predicts that linear fits occur if the data covariances between ID and OOD are the same
up to a constant scaling factor.
Thus, in Figure~\ref{fig:cifar10_gaussian_noise_comparison},
we compare adding \emph{isotropic} Gaussian noise to the \cifarten test set versus adding Gaussian noise with the \emph{same covariance as data examples from \cifarten}.
We find that when the out-of-distribution covariance matches the in-distribution covariance, 
the linear fit is substantially better ($R^2 = 0.93$ vs.\ $R^2 = 0.44$). 
This finding is consistent with the
theoretical model we propose and discuss in Section~\ref{sec:theory}.
\begin{figure*}[ht!]
    \centering
    \includegraphics[height=5.3cm]{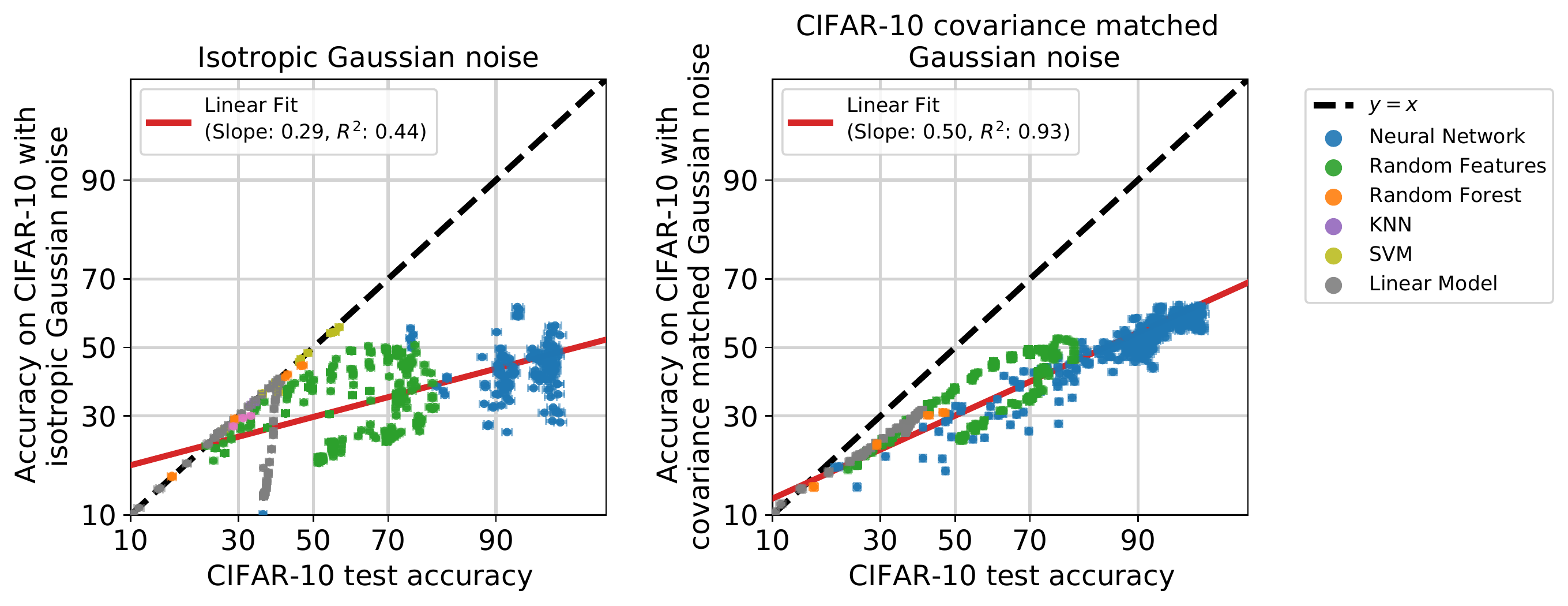}
    \caption{
        When the out-of-distribution data covariance matches the in-distribution data covariance, the linear fit is significantly better.
        \textbf{Left:} A collection of models trained on \cifarten and evaluated in-distribution on \cifarten and out-of-distribution on \cifarten images corrupted with \emph{isotropic} Gaussian noise from \cifartenc.
        \textbf{Right:} The same collection of models evaluated out-of-distribution on \cifarten images corrupted with Gaussian noise with the \emph{same covariance as \cifarten}.
    }
    \label{fig:cifar10_gaussian_noise_comparison}
\end{figure*}
}

\subsection{\iwildcam-v1.0} \label{app:linear-trends-iwc}
We now study version 1.0 of \iwildcam (from WILDS version 1.0), which has a different split between training and ID test sets, compared to the version of \iwildcam we have studied thus far (\iwildcam version 2.0 from WILDS version 1.1). In version 1.0, images from training cameras are assigned uniformly at random between the train and ID test sets, whereas images are randomly partitioned by date between train and ID test splits in version 2.0.  Since the images tend to be taken in bursts, the earlier version of the dataset contains some training and ID test examples that are taken within the same image sequence, and these images tend to be similar because they often capture the same animal at the same location. In other words, we are changing how we measure in-distribution performance, and in this way, our investigation on ID/OOD correlations study different distribution shifts between the two versions. Nevertheless, both versions of the dataset measure out-of-distribution performance in the same way, with train and OOD test splits containing images from disjoint cameras. 

While we use version 2.0 in all other sections, it is still interesting to understand how a different in-distribution train-test split affects the ID/OOD correlation. In Figure~\ref{fig:iwc-v1}, we repeat the experiment reported in Figure~\ref{fig:iwc-arch} on the v1.0 split.\footnote{ There are some differences in training hyper-parameters: Our v1.0 experiments used images with resolution 224x224, slightly different learning rates and number of epochs, and no label noise reduction via MegaDetector-based filtering as described in Appendix~\ref{app:datasets-iwc}. Nevertheless, we are confident that the primary cause for the difference between Figures~\ref{fig:iwc-v1} and~\ref{fig:iwc-arch} is the change in test/train split.} As the figure shows, the ID/OOD correlation is far less pronounced when using the v1.0 split. Moreover, the fine-tuned models show a near-vertical line, with models concentrated around high ID accuracy values but spread across many OOD values, and this could potentially be explained by the high image similarity between train and ID test sets. 

Finally, we remark that while the v2.0 split eliminates overlap in image sequence between train and ID test sets, some near-duplicates inevitably persist in that version as well, particularly for empty frames taken during similar times in the day by the same camera. Investigating the effect of this on the linear trend on \iwildcam v2.0, in which we observe higher variation in performance than in other datasets, is interesting future work.

\begin{figure*}
	\centering
	\includegraphics[width=0.9\linewidth]{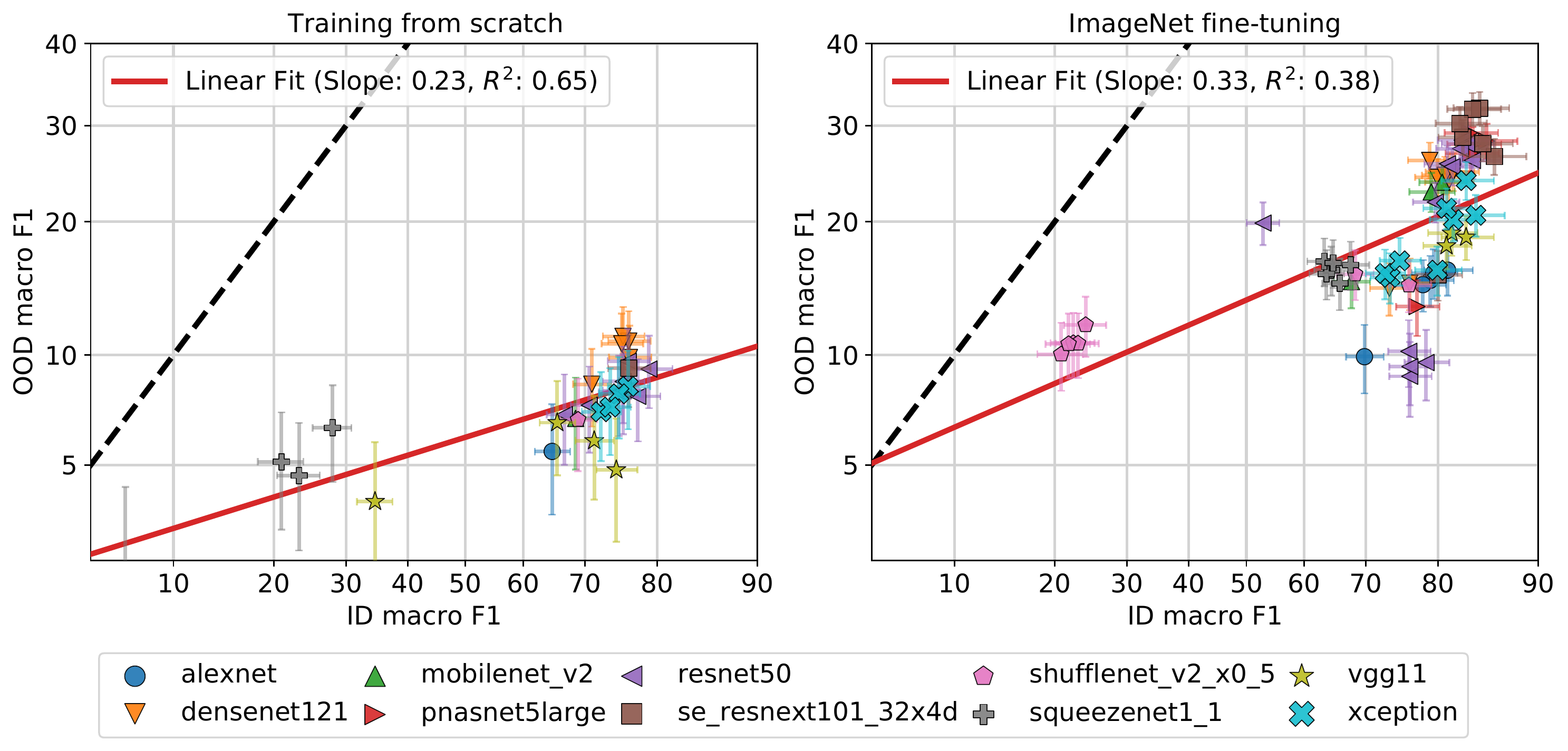}
	\caption{\label{fig:iwc-v1}
		OOD vs.\ ID macro F1 scores for \iwildcam-v1.0 models trained from scratch (left) or fine-tuned from pretrained ImageNet models (right), with varying model architecture and learning rate, but weight decay fixed to zero. Contrast with Figure~\ref{fig:iwc-arch} for results on the v2.0 ID test/train split.
	}
\end{figure*}

\section{The effect of pretrained models}\label{app:more-data}

\subsection{Detailed findings for \cifarten}\label{app:more-data-c10}

In Figure~\ref{fig:cifar10-more-data-app} (left) we reproduce the results shown in Figure~\ref{fig:more-data} and add to it a number of additional models; Figure~\ref{fig:cifar10-more-data-app} (right) graphs the performance of the same model when measuring their OOD performance on \cifartenone instead of \cifartentwo. Let us describe the additional models and their relationship to the linear trend.

First, as a middle ground between zero-shot use of \imagenet models (which is above the line) and fine-tuning (which is on the line), we consider neural network models trained only on the subset of \cinicten that originates from \imagenet (as opposed to \cifarten).
It is worth noting that in this case, the \cinicten subset includes images from \imagenet-21k, which is a superset of the more common \imagenet-1k dataset containing approximately 21,000 classes.
Similar to the zero-shot case, these models use only \imagenet data and so we expect their accuracy to not obey the same \cifarten/\cifartentwo relationship of models trained on \cifarten data (in fact, for these models both \cifarten and \cifartentwo are OOD).
Similar to fine-tuned models, these models are specialized to the task of classifying only the 10 \cifarten classes (as opposed to the 1000 \imagenet classes), and so we expect them to have better accuracy. Figure~\ref{fig:cifar10-more-data-app} lists these models as ``Training on \imagenet data,'' and confirms our expectations: these models are above the linear fit and have better accuracy than the ImageNet zero-shot models. However, in comparison to the zero-shot models, they appear to lie closer to the linear fit for \cifarten-trained models.

Second, we consider two publicly released CLIP models~\citep{radford2021learning}, based on ResNet 50 and Vision Transformer, respectively. Both zero-shot application of CLIP and the training of only its final layer (denoted ``linear probe'') produce performance that is above the line, particularly for the higher-performing Vision Transformer. See below for additional details on the use of CLIP in our experiments.

Finally, we consider models trained on auxiliary unlabeled data originating from the 80 Million Tiny Images~\citep{tinyimages}, abbreviated 80MTI below, which is a superset of both \cifarten and its reproductions \cifartenone and \cifartentwo. In particular, we consider a model trained via self-training using a subset of TinyImages~\citep{carmon2019unlabeled}, listed as 80MTI ST in Figure~\ref{fig:cifar10-more-data-app}), and two models trained via out-distribution aware self-training~\citep{augustin2020out}, listed as 80MTI ODST. As the figure shows, despite using auxiliary data,  the ``80MTI ST'' performance is precisely on the \cifarten-only linear trend. This might be due to the fact that~\citet{carmon2019unlabeled} filter 80MTI, using a model supervised with the \cifarten training set, thereby possibly losing the additional diversity of TinyImages.\footnote{We also note that the additional unlabeled data used to train this model potentially contains images from \cifartenone and \cifartentwo; if it does, they seem to do little to help its performance on that dataset.} The performance of the ODST models appear to deviate from the linear trend. However, the direction of the deviation is inconsistent, being below the line on \cifartentwo and above the line on \cifartenone.
Since these are the highest-accuracy models in our testbed it is not completely clear whether these deviations are due to use of extra data or a deviation of the overall ID-OOD trend from a perfect probit linear fit at high accuracies.

\paragraph{Experiment details.}
Below we provide some additional details on out \cifarten auxilliary data experiments.
\begin{itemize}
	\item \textbf{Zero-shot classification with \imagenet models.} To investigate models that are minimally affected by the CIFAR-10 training set, we utilized pre-trained \imagenet models directly for the CIFAR-10 classification task without any fine-tuning (``zero-shot'').
	A complication here is that the CIFAR-10 classes do not match the ImageNet classes.
	For instance, ImageNet contains more than 100 different dog classes corresponding to different breeds while CIFAR-10 contains only one dog class.
	To address this point, we manually constructed a mapping from \cifarten classes to \imagenet classes.
	Our mapping roughly followed the WordNet hierarchy with some refinements from the class structure used in the human annotation experiments conducted by \citep{humanaccuracy}.
	We then evaluated the ImageNet models using only the logits for classes appearing in this mapping and picked the CIFAR-10 class as prediction that corresponded to the ImageNet class with the largest logit.
	\item \textbf{Zero-shot classification with CLIP models.} For the models described as ``CLIP zero-shot", we use the publicly released CLIP package,
	which includes the ResNet and VisionTransformer models as well as the text tokenizer and encoder used to encode the zero-shot text prompts.
	We obtained the \cifarten prompts through private correspondence with the OpenAI team. For each \cifarten class, we ensembled the prompts by
	averaging the embeddings of the prompts together before using it for final classification.
	\item \textbf{Linear probes.} In the models described above as ``linear probes'' we train only the last layer of a pre-trained neural network by performing (exact) least-squares linear regression of 1-hot class representation using the activations of the network's penultimate layer.
\end{itemize}

\begin{figure*}
	\centering
	\includegraphics[width=\linewidth]{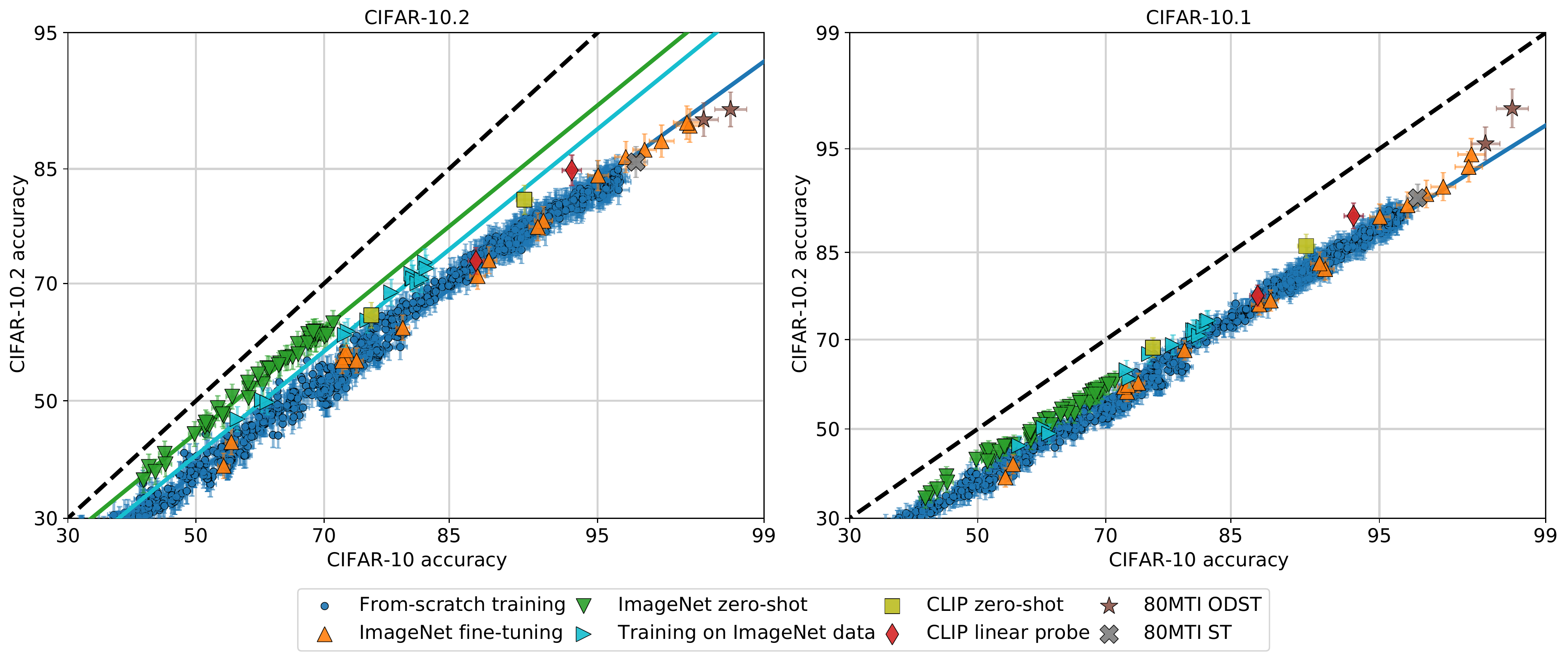}
	\caption{\label{fig:cifar10-more-data-app}
		The effect of additional training data on OOD accuracy on \cifartentwo (left) and \cifartenone (right).
	}
\end{figure*}

\subsection{Detailed findings for \fmow}\label{app:more-data-fmow}

Figure~\ref{fig:fmow-more-data-app} shows a reproductions of Figure~\ref{fig:more-data} (middle) when using different combination of worst-region and average-region accuracy metric for the ID and OOD data (recall that the ID/OOD split is based on time). As the figure shows, the effect of fine-tuning pre-trained models is consistent across the four combinations: fine-tuning improves performance without deviating from the line. We also consider a linear probe of CLIP (see description in the previous subsection). Unlike the result on \cifarten, here the CLIP models do not significantly deviate from the linear trend. A possible explanation for this difference is that the web images on which CLIP was trend contain far more images of objects relevant for the \cifarten classification task than they do for the \fmow satellite image classification task.

\begin{figure*}
	\centering
	\includegraphics[width=\linewidth]{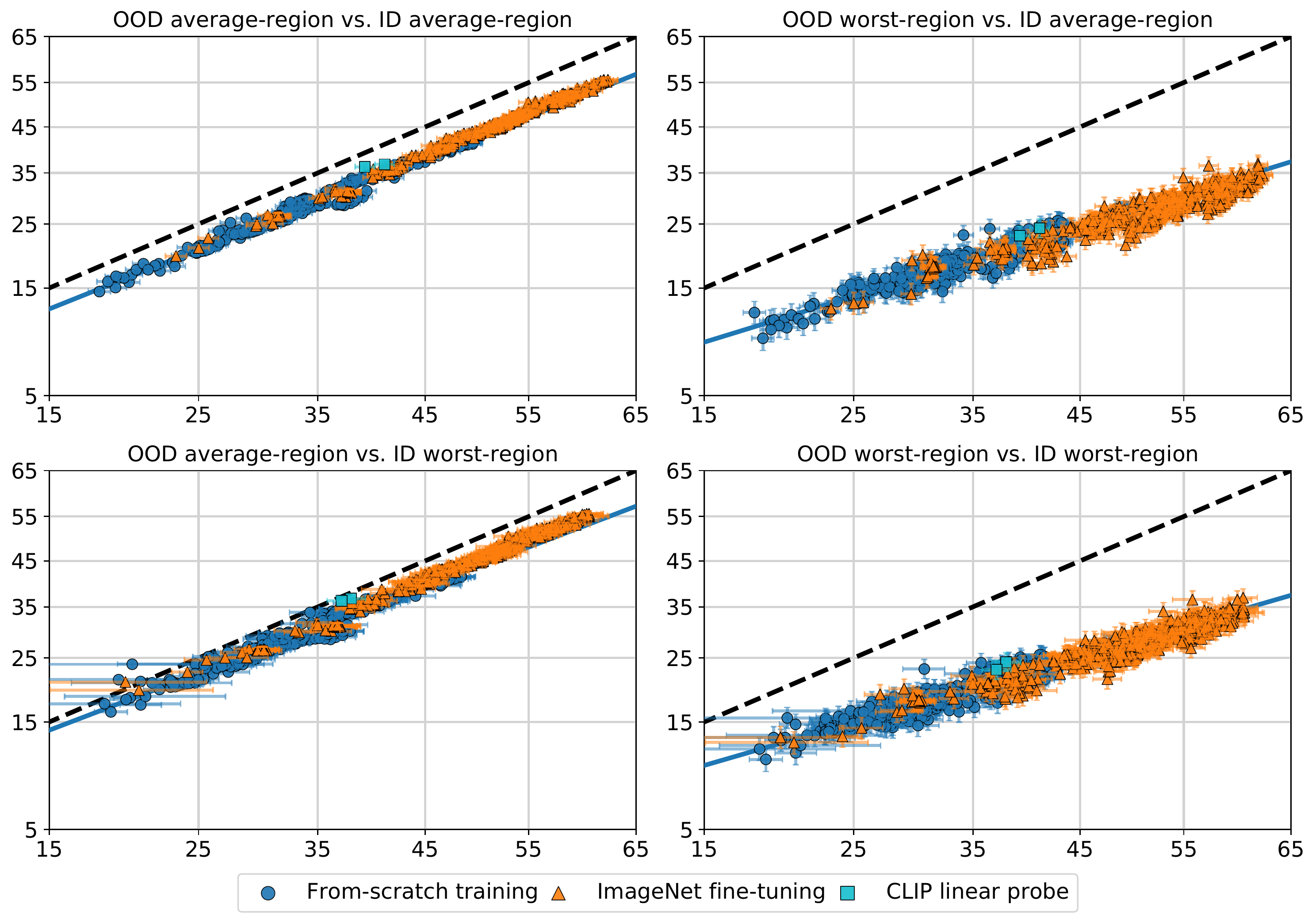}
	\caption{\label{fig:fmow-more-data-app}
		The effect of additional training and different accuracy metrics on \fmow ID/OOD performance.
	}
\end{figure*}

\subsection{Detailed findings for \iwildcam}\label{app:more-data-iwc}

Figure~\ref{fig:iwc-app-wd} shows the same models plotted in the \iwildcam panel of Figure~\ref{fig:main_figure}, but separating the models trained from scratched and the fine-tuned models, and coloring points by the weight decay parameters. (For each weight decay we vary model architecture and learning rate). For fine-tuned models, there is a clear difference in the ID/OOD linear between model using weight decay 0 and models using nonzero weight decay. In particular, points with nonzero weight decay seem to lie above the zero weight decay linear trend. For models trained from scratch the macro F1 measurement  error does not allow us to conclude with confidence whether weight decay affects the linear trend. Finally, it is worth noting that---even though increasing weight decay appears to move models above the zero weight decay line---the models with the best performance, both ID and OOD, do not use weight decay.

\begin{figure*}
	\centering
	\includegraphics[width=0.9\linewidth]{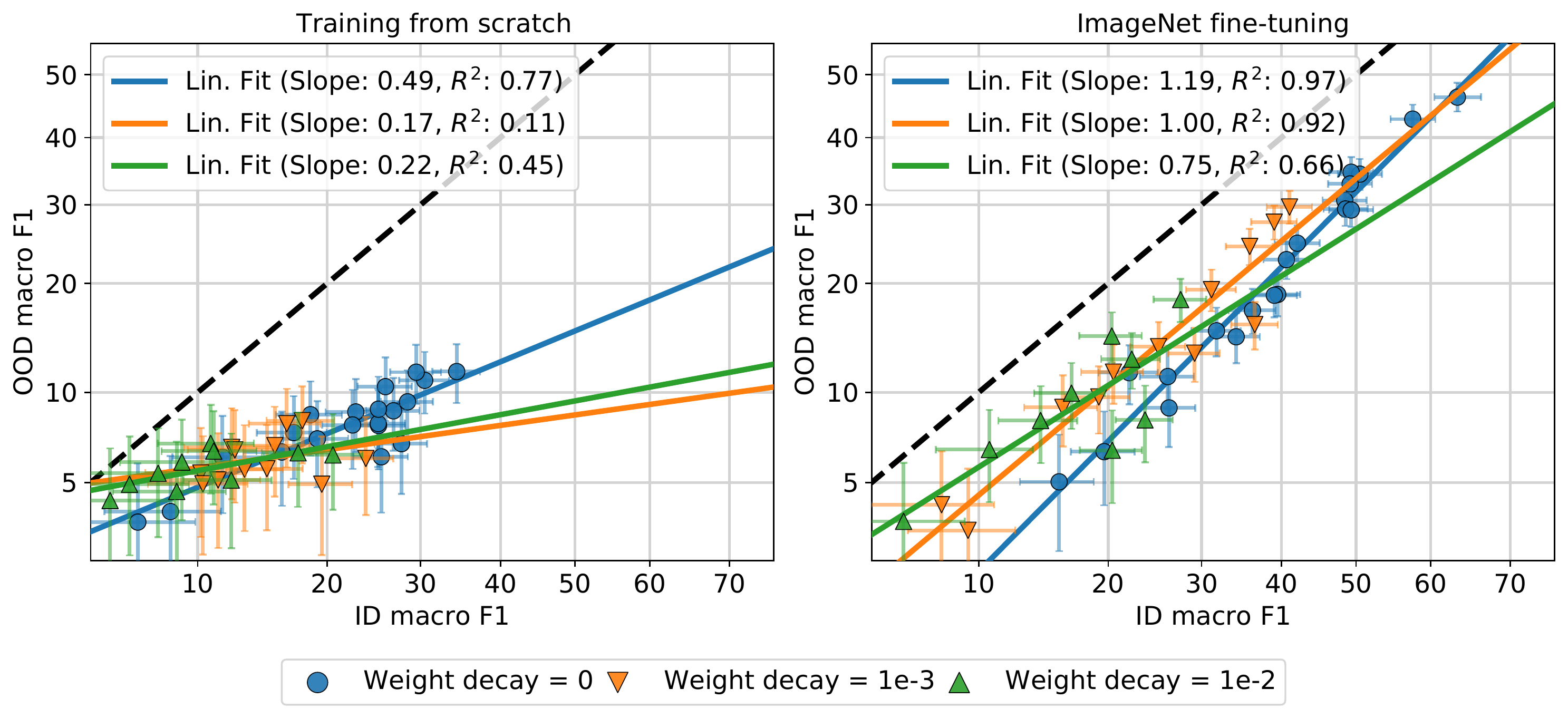}
	\caption{\label{fig:iwc-app-wd}
	OOD vs.\ ID macro F1 scores for \iwildcam models trained from scratch (left) or fine-tuned from pretrained ImageNet models (right), with varying model architecture, learning rates, and weight decay. We observe that fine-tuned models exhibit a different linear trend than models trained from scratch, and moreover that the weight decay parameters affects the ID/OOD correlation, at least for fine-tuned models.
	}
\end{figure*}

\section{Theoretical models for linear fits}

\subsection{Proof of Theorem~\ref{thm:theory}}\label{app:theory-proof}

\restateTheOneTheoremWeHave*

\begin{proof}
We begin by deriving expression for the accuracy of linear classifiers in our Gaussian distributional model. Under distribution $D$, the accuracy of linear classifier $\classifier$ is 
\begin{flalign*}
	\acc_{D}(\classifier) & = \Pr\prn*{ \sign(\classifier^\top \vx) =  y }
	= \Pr \prn*{ y \cdot \classifier^\top \vx  \ge 0 }
	= \Pr \prn*{ \mathcal{N}( \classifier^\top \meanvar; \norm{\classifier}^2 \sigma^2 ) \ge 0} \\ &
	= \Pr \prn*{ \norm{\classifier} \sigma \cdot  \mathcal{N}(0; 1) \ge - \classifier^{\top} \meanvar}
	= \Phi \prn*{ \frac{\classifier^{\top}\meanvar}{\norm{\classifier}\sigma}},
\end{flalign*}
where we recall that $\Phi(t) = \int_{-t}^{\infty} \frac{1}{\sqrt{2\pi}} e^{-s^2/2}\mathrm{d} s = \int_{-\infty}^{t} \frac{1}{\sqrt{2\pi}} e^{-s^2/2}\mathrm{d} s$ is the standard Normal cdf. Similarly, for the shifted distribution $D'$ we have
\begin{equation*}
	\acc_{D'}(\classifier) = \Phi \prn*{ \frac{\classifier^{\top}\meanvar'}{\norm{\classifier}\sigma'}}
	= \Phi \prn*{
		\frac{\alpha}{\gamma} \cdot \frac{ \classifier^{\top}\meanvar}{\norm{\classifier}\sigma}
	+ \frac{\beta}{\gamma \sigma} \cdot \frac{ \classifier^{\top}\shiftvar}{\norm{\classifier}}.
	}
\end{equation*}
Therefore,
\begin{equation}\label{eq:theoretical-line-departure}
\abs*{\Phi^{-1}\prn*{ \acc_{D'}(\classifier)} - \frac{\alpha}{\gamma}  \Phi^{-1}\prn*{ \acc_{D}(\classifier)}}
 = \frac{\beta}{\gamma\sigma} \abs*{ (\classifier/\norm{\classifier})^\top \shiftvar}.
\end{equation}
Since $\classifier$ is independent of $\shiftvar$ and $\classifier/\norm{\classifier}$ is a unit vector, the inner product $(\classifier/\norm{\classifier})^\top \shiftvar$ is distributed identically to the first coordinate of $\shiftvar$. A standard concentration bound on the sphere~\citep[see, e.g.,][Lemma 2.2]{ball1997elementary} states that
\begin{equation*}
	\Pr( \abs{\shiftvar_1} > z) \le 2 e^{- d z^2 / 2}
\end{equation*}
for all $z\ge 0$. Substituting $z=\sqrt{2d^{-1}\log\frac{2}{\delta}}$ completes the proof.
\end{proof}

\paragraph{Remarks.}
We conclude this subsection with two additional remarks on the application of Theorem~\ref{thm:theory}.
\begin{itemize}
	\item \textbf{Classifiers trained on samples from $D$.} We note that any mapping of samples from $D$ to a linear classifier results by definition in a classifier independent on $\shiftvar$, and consequently Theorem~\ref{thm:theory} applies to it. In particular, it applies to the linear classifier we train in the simulation described in Figure~\ref{fig:theory-main}.
	\item \textbf{A guarantee for multiple models.} Given $N$ linear classifiers $\classifier_1,\ldots, \classifier_K$ such that each one is independent of  $\shiftvar$, we may apply Theorem~\ref{thm:theory} with probability parameter $\delta/N$ in conjunction with a union bound to conclude that, with probability at least $1-\delta$ we have 
	$\abs*{\Phi^{-1}\prn*{ \acc_{D'}(\classifier_i)} - \frac{\alpha}{\gamma}  \Phi^{-1}\prn*{ \acc_{D}(\classifier_i)}}  \le \; \frac{\beta}{\gamma \sigma}\sqrt{\frac{2\log\sfrac{2N}{\delta}}{d}}$
	for all $i=1,\ldots,N$. This precisely implies a linear trend in scatter plots such as Figure~\ref{fig:theory-main}.
\end{itemize}


\subsection{Departures from the linear trend}\label{app:theory-departures}
\begin{figure*}[t]
	\centering
	\includegraphics[height=6.0cm]{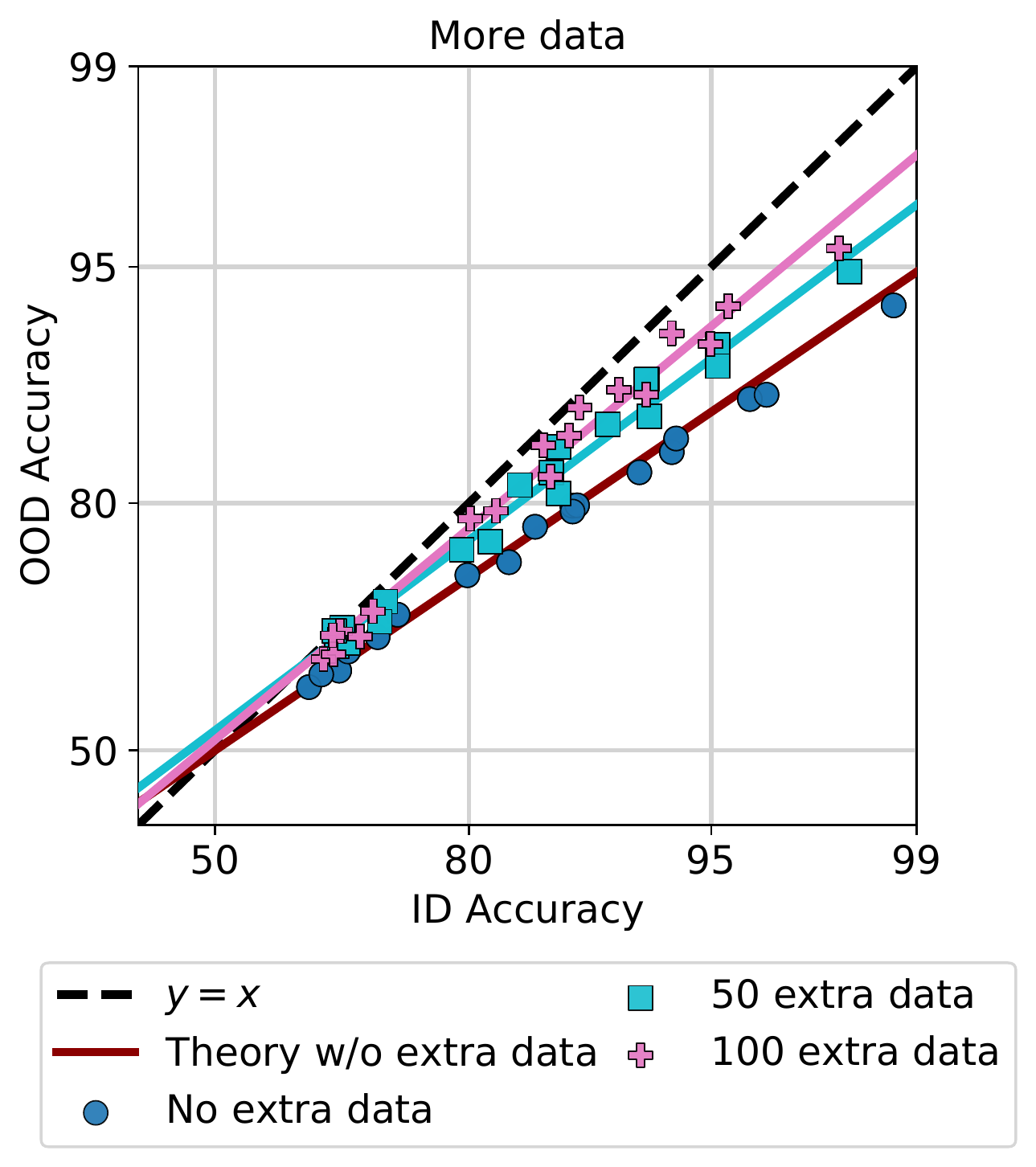}
	\includegraphics[height=6.0cm]{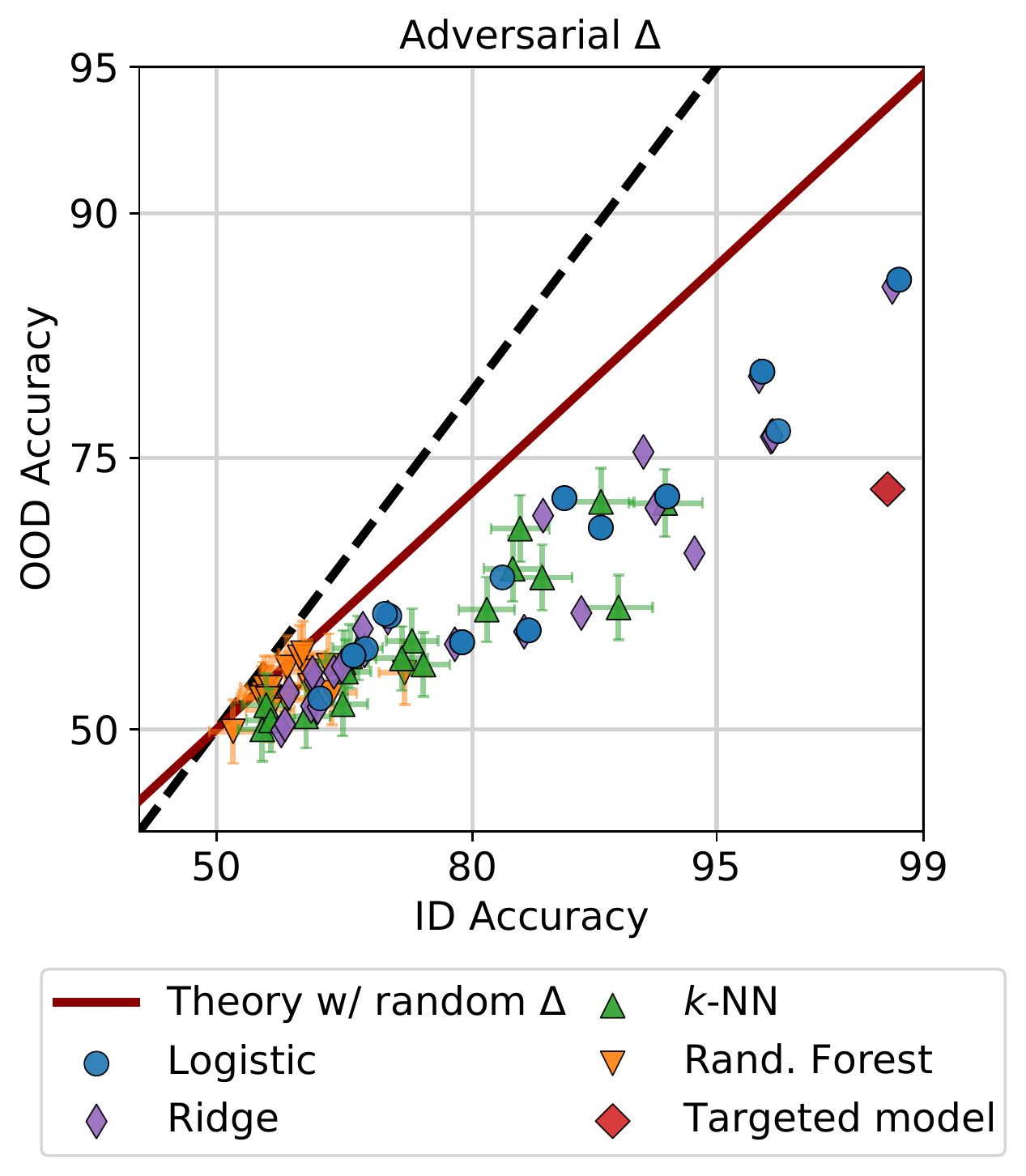}
	\includegraphics[height=6.0cm]{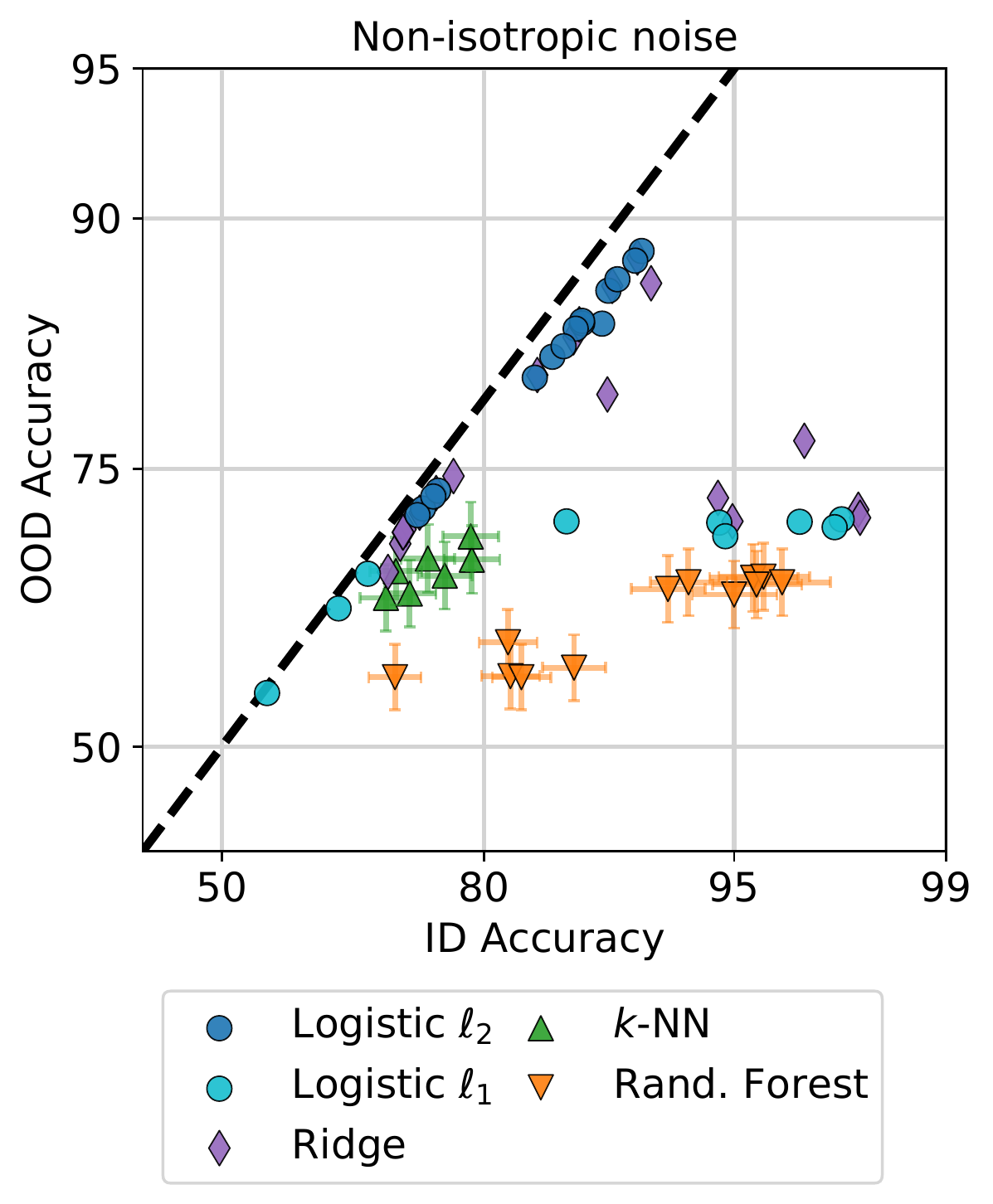}
	\caption{\label{fig:theory-departures}%
	Modifications to our theoretical model showcasing departures from the linear trend. \textbf{Left:} training on auxiliary data related to $D'$ (this plot only shows logistic regression models). \textbf{Middle:} choosing the parameter $\shiftvar$ adversarially to reduce the OOD performance of a particular targeted model. \textbf{Right:} changing the noise covariance to be non-isotropic.
	}
\end{figure*}

We now detail a number of modifications to our distribution model which break the linear trend predicted by Theorem~\ref{thm:theory} and shown in Figure~\ref{fig:theory-main}. In each case, we mathematically define the modified model, provide intuition for why the linear trend no longer holds, and demonstrate the departure from the linear trend via simulation where train prediction models using samples from $D$ and evaluate them on $D'$. We defer the full simulation details to the next subsection.
The modifications we describe are not the only possible way to depart from the linear fit, but we focus on them because we believe they potentially represent departures from the trend seen in practice,

\paragraph{More data.}
Section~\ref{sec:training_data}, as well as prior work, show that using additional training data from a broader distribution can cause departure form the linear trend. 
To simulate such a scenario, we consider a third distribution $D''$ defined by  $\vx | y \sim \mathcal{N} ( \meanvar'' \cdot y;  (\sigma'')^2 I )$, with $\meanvar'' = \meanvar' + \beta \tilde{\shiftvar}$ and $\tilde{\shiftvar}$ uniformly distributed on the unit sphere and independent of $\shiftvar$. (Recall that $\meanvar' = \alpha \meanvar + \beta \shiftvar$). Since $D''$ is more similar to $D'$, we expect that including $D''$ samples in the training will result in better OOD performance. However, this inclusion could harm ID performance. 

We demonstrate these effects in Figure~\ref{fig:theory-departures} (left), where we train logistic regression models using samples from $D$ and 0, 50 or 100 samples from $D'$, with $\sigma'' = \sqrt{2}\sigma$ and all other parameters identical to the experiment shown in Figure~\ref{fig:theory-main}. As expected, the extra data results in better OOD performance but worse ID performance. Moreover, the models trained on each amount of external training data appear to roughly follow linear trends (the plot shows empirical probit linear fits). However, we note that our theoretical analysis does not guarantee such linear fit, because the training data used to compute the classifier depends on $\shiftvar$ through the samples from $D''$.

\paragraph{Adversarial distribution shift.}
As previously mentioned, the randomization of the distribution shift was crucial for our assumption, because if we allow $\shiftvar$ to be a fix deterministic vector we cannot rule out that it depends adversarially on the trained classifier. Let us now spell out the implications of this possibility, by allowing $\shiftvar$ to be any arbitrary vector of norm at most 1. Given some target classifier given a target classifier $\classifier$, suppose  pick $\shiftvar = c\cdot \classifier / \norm{\classifier} = c \cdot  \classifier / \norm{\classifier}$ for some $c\in [-1,1]$. This makes the inner product $\hat{\classifier}^{\top} \shiftvar = c$. Recalling Eq.~\eqref{eq:theoretical-line-departure}, this clearly implies a large departure from the linear trend when $|c|$ is close to 1. In particular, by picking negative $c$ we may substantially reduce the performance of the model on $D'$. We note that this form of distribution shift is precisely the Gaussian model of adversarial examples proposed by~\citet{schmidt2018adversarially}.

In Figure~\ref{fig:theory-departures} (middle), we demonstrate this technique by selecting one of the linear classifiers shown in Figure~\ref{fig:theory-main}, call it $\classifier^\star$, and letting $\shiftvar = -0.03 \classifier^\star / \norm{\classifier^\star}$. As the figure shows, the linear trend breaks substantially, particularly for the targeted classifier. 

\paragraph{Non-isotropic covariance shifts.}
Finally, we consider the case where the noise covariance under $D$ is not isotropic. That is, we let $\vx | y \sim \mathcal{N} ( \meanvar \cdot y; \Sigma )$ for some $\Sigma$ that is not a multiple of the identity. Instead of considering shifts to the mean $\meanvar$, we consider random covariance shifts of the form
\begin{equation*}
	\Sigma' = \Sigma + (\sigma')^2 I_{d \times d},
\end{equation*}
i.e., simple additive white Gaussian noise with variance $\sigma'$.  Under this distribution shift model, the probit accuracies are
\begin{equation*}
\Phi^{-1}\prn*{ \acc_{D}(\classifier)} = \frac{\classifier^\top \meanvar}{\sqrt{\classifier^\top \Sigma \classifier}}
~~\mbox{and}~~
\Phi^{-1}\prn*{ \acc_{D'}(\classifier)} = \frac{\classifier^\top \meanvar}{\sqrt{\classifier^\top \Sigma' \classifier}}
\end{equation*}
For a the linear ID-OOD probit accuracy relationship to hold for all $\classifier$, we must have that $\Phi^{-1}\prn*{ \acc_{D}(\classifier)} / \Phi^{-1}\prn*{ \acc_{D'}(\classifier)}$ is a constant independent of $\classifier$, which happens if and only if
\begin{equation*}
	\frac{\classifier^\top \Sigma' \classifier}{\classifier^\top \Sigma\classifier}
	=
	 1 + \frac{\sigma'^2 \norm{\classifier}^2}{\classifier^\top \Sigma \classifier}
\end{equation*}
is a constant independent of $\classifier$. However, this only holds when $\Sigma$ is a multiple of the identity, contradictory to our assumptions. Indeed, whenever $\Sigma$ is not a multiple of the identity, 
there could be a  tradeoff between ID and OOD performance: the former favors $\classifier$ with small $\Sigma$-weighted norms, while the latter also depends on the standard Euclidean norm $\norm{\classifier}^2$. Consequently, we expect regularization that limits $\norm{\classifier}^2$ to provide better OOD performance.

Figure~\ref{fig:theory-departures} (right) demonstrates this phenomenon via simulation. In the figure, we set $\Sigma$ to be diagonal with a portion of the entries close to zero so that giving the corresponding coordinates of $\classifier$ larger weight results in better ID accuracy. For the distribution shift we let $(\sigma')^2 = \tr(\Sigma) / d$. As the figure shows, the linear trend no longer holds, despite the fact that the distribution shift is ``only'' adding isotropic Gaussian noise to the covariates. Moreover, as the above discussion predicts, the logistic and ridge regression models that attain strong OOD performance are those with stronger $\ell_2$ regularization. We also show logistic regression trained classifiers with $\ell_1$ regularization---these classifiers do not achieve good OOD performance.

\subsection{Simulation details}\label{app:theory-figure-details}
Below, we provide additional details about the simulations described in Figures~\ref{fig:theory-main} and~\ref{fig:theory-departures}. 

\paragraph{Training parameters.} We fit logistic regression, ridge regression, nearest neighbors and random forest models using their scikit-learn implementations~\cite{scikitlearn}. For logistic regression we use values of the inverse-regularization parameter $C$ ranging from $10^{-6}$ to $1$; we use $\ell_2$ penalty throughout except the covariance shift experiment where we also consider $\ell_1$ penalty. For ridge regression we use values of the regularization parameter $\alpha$ ranging from $10^{-3}$ to $10$. For both types of linear models we do not fit an intercept. For nearest neighbors we use $1$ or $3$ nearest neighbors, and for random forests we use $3$, $30$ or $100$ estimators. The remaining parameters are set to their scikit-learn defaults. 

\newcommand{\nsub}{n_{\mathrm{sub}}}
\newcommand{\dproj}{d_{\mathrm{proj}}}

In addition to varying the learning hyperparameters described above, to produce models with varying accuracy we also modulate the training set size and dimensionality reduction. To reduce the training set size to size $\nsub$, we pick the first $\nsub$ entries from a fixed training set generated once. To reduce dimensionality down to $\dproj$, we simply pick the first $\dproj$ coordinates of $x$. For the all simulations except covariance shift, we let $\nsub$ range between $30$ and $100$, and use $\dproj$ in the range $50$ to $3000$. In the covariance shift simulation we use $\nsub\in[100,2000]$ and fix $\dproj=d=500$. 

\paragraph{Accuracy measurement.}
For linear models we compute the accuracy exactly (see Subsection~\ref{app:theory-proof} for closed-form expressions). Consequently, we do not show error bars for these models. For the remaining models we estimate the accuracy on samples from the appropriate distributions and use error bars to show 95\% Clopper-Pearson confidence intervals, consistently with the rest of the paper.  

\paragraph{Distribution model parameter setting.}
Throughout, we pick $\mu$ to be random unit vector (i.e., with the same distribution as $\shiftvar$).  
For all simulations except covariance shift, we let $d=10^{5}$, $\sigma=10^{-1.5}$, $\alpha=0.7$, $\beta=0.5$ and $\gamma=1$. For the covariance shift simulation, we found that using a smaller dimension and more training points lead to more noticeable effects. Therefore, for this simulation we let $d=500$ (recall that $\alpha, \beta$ and $\gamma$ do not exist in the covariance shift model). We let the covariance matrix $\Sigma$ be diagonal, with $490$ diagonal entries of size $1/2$  and the remainder of size $1/200$; the locations of the small entries were chosen at random. The shifted covariance is $\Sigma' = \Sigma + \frac{1}{8} I_{d\times d}$.

\section{Additional related work}\label{app:related-work}
We now summarize some of the additional work related to the phenomena we study in our paper.
Our focus here is mostly on recent work.
For early work on distribution shift, we refer the reader to \citep{datasetshiftbook,torralba2011unbiased}.

\paragraph{PAC-Bayesian analysis of distribution shift.} Performance under distribution shift has also been characterized under the PAC-Bayesian setting where the learning algorithm outputs a posterior distribution over the h hypothesis class~\citep{li2007bayesian, germain2013pac, germain2016new}.~\citet{li2007bayesian} directly bound the error on the target distribution (OOD) in terms of the empirical error on a small number of labeled samples from the target and a ``divergence prior'' which measures some divergence between the source and target domains.~\citet{germain2013pac} relate the OOD performance to the ID performance via a disagreement measure induced by the hypothesis class. These bounds do not explain the linear trends we find in this paper---\citet{li2007bayesian} do not relate the source and target error directly, and the bounds in~\citet{germain2013pac} are functionally similar to those of~\citet{bendavid2006analysis} where the ID performance is highly predictive of the OOD performance only if they are equal (Figure 1).~\citet{germain2016new} present a different analysis where the domain divergence appears as a \emph{multiplicative} term rather than an additive one like in previous bounds. However, this bound expresses a linear relation between the OOD performance and some exponent of the ``expected joint error'' on the source domain which is different from the ID performance. Furthermore, the bound is an inequality which only provides an upper bound on the OOD performance, while our empirical results require a bound in the other direction as well.

\paragraph{Reliability of machine learning benchmarks.}
When assessing the reproducibility and reliability of statistical findings,
\citet{yu2013stability} advocates for considerations of \emph{stability} of
statistical results under perturbations of the model and data. Key to this
account is a notion of a stability target that is estimated under changes to the
data used to estimate the target.  Viewed in this light, our experimental
methodology of training models on one distribution and evaluating them on a
collection of out-of-distribution test sets corresponds to testing the stability
of an appropriately chosen target under perturbations to the test set. The
consistent accuracy drops we observe across out-of-distribution test sets
suggests that model accuracies are not stable; however, the precise linear
trends we find suggest that \emph{model rankings} are in fact stable. As a
consequence, benchmarks like \cifarten or \imagenet where we observe precise
linear trends may provide more reliable knowledge about relative model
performance than absolute performance on new out-of-distribution test sets.

\paragraph{Theoretical models for linear trends in earlier work on dataset reproduction.} Both \citet{recht2018cifar10} and \citet{recht2019imagenet} contain simple models for the linear fits observed in their reproductions of \cifarten and \imagenet.
\citet{recht2018cifar10} propose a mixture model with an ``easy'' and ``hard'' component and model the distribution shift as a change in the weigts of these two components.
Their model does indeed give a linear fit, but only with linear axis scaling.
As we have seen several times throughout this paper, the scatter plots show cleaner linear trends with logit or probit scaling on the axes.
It is also not clear what the ``easy'' and ``hard'' components correspond to in distribution shifts such as \cifartenone.

\citet{recht2019imagenet} developed their model further.
Instead of discrete mixture components, each distribution is now parametrized by a Gaussian distribution over the ``hardness'' of each image.
In addition, every model has a scalar ``skill'' parameter that determines the probability of a model classifying an image with a given hardness correctly.
This model now produces linear fits in the probit domain, which yields a closer fit to empirical results.
While a continuous hardness parametrization also is more plausible, it is again unclear what this hardness corresponds to.

Neither the models of  \citet{recht2018cifar10,recht2019imagenet} nor our  model of Section~\ref{sec:theory} allow us to predict where linear trends occur in actual data; such predictive power is important because---as we demonstrate---some distributions do not yield linear trends. However, our theoretical analysis is based on a concrete generative  is based on a concrete generative, rather than postulated abstract properties of data and classifiers. One advantage of this fact is that it allows us to consider modifications of our generative models which show departures from the linear trend, as we do in Appendix~\ref{app:theory-departures}.

\paragraph{Linear trends in image classification with natural language supervision.}  Among other results, \citet{radford2021learning} show two important phenomena that are closely related to this paper.
First, their training approach (contrastive language image pre-training, ``CLIP''), which combines a large training set and natural language supervision, produces image classifiers substantially above the linear trend given by a wide range of \imagenet model in the distribution shift testbed of \citet{taori2020measuring}.
This result provides further evidence for the hypothesis that training data plays an important role in the linear trends we describe in this paper.
Second, \citet{radford2021learning} find that once their training set is fixed and they vary model architecture (ResNet variants and Vision Transformers~\citep{dosovitskiy2021image}) and compute available for training, the resulting models again follow a clear linear trend.
This demonstrates that linear trends between in-distribution and out-of-distribution accuracy occur in a diverse range of settings.

\paragraph{Linear trends under sub-population shift.} One specific type of distribution shift is \emph{sub-population} shift.
In sub-population shift, each class is composed of a set of sub-populations, e.g., the ``dog'' class in an image classification task may be composed of images from a specific set of dog breeds.
A natural goal then is that a trained classifier should generalize to previously unseen dog breeds and still correctly labels them as ``dog''.
\citet{hendrycks2018benchmarking} found that a set of eight convolutional neural networks follow a linear trend on a sub-population shift derived from \imagenet-22K.
\citet{santurkar2021breeds} construct a range of sub-population shifts from \imagenet and find approximately linear trends for several of the shifts they consider.
Their testbed contained 13 convolutional neural networks, some of them with interventions such as adversarial training \citep{madry2017towards}.
Some of the plots in \citep{santurkar2021breeds} are not directly comparable to ours since they display a relative accuracy measure on the y-axes, not the absolute accuracy (i.e., average 0-1 loss).

\paragraph{Underspecification as defined in \citet{damor2020underspecification}.} \citet{damor2020underspecification} conduct a broad empirical study and show that out-of-distribution performance can vary widely even for models with the same in-distribution performance.
Since this result may at first glance disagree with our results here, we now discuss their empirical results most relevant to our paper in detail.
In particular, we focus on their results in computer vision domains.
\begin{itemize}
    \item \citet{damor2020underspecification} point out ImageNet-C as an example of underspecification in image classification.
    Similar to \citep{taori2020measuring}, we also find in Figure \ref{fig:corruptions1} that some of the perturbations in ImageNet-C show substantial variation as a function of ImageNet accuracy.
    In addition, we find that this variation occurs in \cifartenc.
    As mentioned before, not all shifts in ImageNet-C and \cifartenc are affected by underspecification, with some shifts exhibiting comparatively clean linear trends.
    \item The second example for underspecification in image classification is ObjectNet \cite{barbu2019objectnet}.
    While it is indeed true that the accuracy variation on ObjectNet may increase compared to ImageNet, overall ObjectNet still shows predictable behavior as a function of ImageNet accuracy.
    See Figure 2 in \citep{taori2020measuring}.
    \item In addition to standard computer vision benchmarks, \citet{damor2020underspecification} also investigate two medical imaging datasets, which give an important complementary perspective.
    In the first dataset (ophthalmological imaging), they find evidence of underspecification.
    In the second dataset (dermatological imaging), the evidence is less clear since the tests for statistically significant variation in the four domains give p-values of 0.54, 0.42, 0.29, and 0.03.
    While the fourth p-value is below 0.05, the authors did not correct for multiple hypothesis testing and remark that this is an exploratory data analysis.
\end{itemize}
Overall, we find that the empirical evidence for underspecification in computer vision tasks is nuanced.
As in our work, some distribution shifts studied by \citet{damor2020underspecification} exhibit stronger correlation between in-distribution and out-of-distribution than others.
Hence there is no clear contradiction between our results and those of \citet{damor2020underspecification}.
Understanding when precise linear trends occur and when underspecification is dominant is an important direction for future work.

\paragraph{Further distribution shifts without universal linear trends.} While we have seen several distribution shifts with clean linear trends between in-distribution and out-of-distribution generalization in this paper, there are also obvious counterexamples.
One prominent counterexample are adversarial distribution shifts, e.g., $\ell_p$ adversarial examples \citep{biggio2013,szegedy2013intriguing,biggio2017wild}.
For models trained without a robustness intervention, it is usually easy to construct adversarial examples that cause the model to misclassify most inputs despite high accuracy on unperturbed examples.
While adversarial robustness is far from solved, it is now possible to train CIFAR-10 networks with about 65\% accuracy against the common $\ell_\infty$ adversary with $\varepsilon = 8 / 255$ and standard (unperturbed) accuracy of 91\% \cite{gowal2020uncovering}.
Since CIFAR-10 classifiers without a robustness intervention have only 0--10\% robust accuracy in this setting, it is clear that there cannot be a precise linear trend between in-distribution and out-of-distribution accuracy. Adversarial distribution shifts can bring about departures from the linear trend in our theoretical setup as well, as we discuss in Appendix~\ref{app:theory-departures}.  
We refer the reader to \citet{taori2020measuring} and \citet{hendrycks2020faces} for additional examples of models not following a linear trend in ImageNet variants, e.g., on some of the ImageNet-C corruptions and ImageNet-R.

\paragraph{Benchmarks for distribution shift.} Recently several groups conducted broad empirical surveys of distribution shift, comparing a wide range of available methods.
Most closely related to our paper is \citet{taori2020measuring}, where the authors also find clean linear trends on multiple distribution shifts related to \imagenet.
\citet{djolonga2020robustness} also observed high correlations on the same distribution shifts for a smaller number of models.
Both experiments were limited to \imagenet as in-distribution test set and convolutional neural networks.
Here we study multiple different in-distribution datasets for image classification, an additional task (pose estimation), and several models beyond convolutional neural networks.

\citet{lostgeneralization} conduct a broad survey of algorithms for the closely related problem of domain generalization.
In domain generalization, the training set is drawn from multiple distinct domains, and the learning algorithm has access to the domain labels.
At test time, the trained models is evaluated on samples from a new domain.
\citet{lostgeneralization} found that on a range of datasets, current domain generalization algorithms perform only as well or worse as an empirical risk minimization baseline that ignores the domain structure.
At a high level, this result is similar to the aforementioned distribution shift benchmarks that also found small or no gains from current robustness interventions on most distribution shifts.

Our results raise similar questions as these benchmarks for distribution shift and domain generalization: when and how is it possible to improve over empirical risk minimization as a baseline for robustness to distribution shift, i.e., to ``go above the line'' in our scatter plots?

\paragraph{Training methods to improve robustness.} 
Researchers have proposed a large number of robustness interventions over the past few years.
Due to the volume of papers, we only refer to recent surveys here. 
Methods for improving robustness divide into two categories: those which use
samples from the target distribution (which we refer to as the OOD data), and
those that do not. The former methods are usually called transfer learning and
domain adaptation methods~\citep{transfer_survey,wang2018deep}. These methods
typically assume that the target distribution data is more constrained than the
in-distribution data, either lacking labels or having smaller quantity, and
algorithms focus on mitigating these issues. The linear trends observed by
\citet{kornblith2019better} in the context of transfer learning suggest that
there may be important similarities. See \citet{robey2021model} for one example
applying such techniques to the WILDS OOD shifts considered in this work.

While domain adaptation and transfer learning techniques are helpful in many settings, they are not always applicable. 
For instance, when we want an autonomous vehicle to drive safely in a new town it has not visited before, we have no additional training data available to adapt the car's perception system.
Such scenarios motivate our study of the correlation between in-distribution and out-of-distribution generalization in this paper.
The second category of training methods---sometimes referred to as domain robustness or domain generalization---attempt to learn models that are reliable in the presence of distribution shifts for which there is no direct training data. Instead, these methods often leverage data from multiple other, related domains. \citet{lostgeneralization} provide an overview of current methods for domain generalization.

\end{document}